\documentclass[acmtog]{acmart}
\acmSubmissionID{330}

\usepackage{booktabs} 

\usepackage{multirow}
\usepackage{caption}
\usepackage{subcaption}
\usepackage{soul}
\usepackage{pgf,tikz,pgfplots}
\usepackage[ruled,linesnumbered]{algorithm2e} 
\usepackage{comment}

\SetAlFnt{\small}
\SetAlCapFnt{\small}
\SetAlCapNameFnt{\small}
\SetAlCapHSkip{0pt}

\acmJournal{TOG}

\setcopyright{acmlicensed}\acmJournal{TOG}
\acmYear{2021}\acmVolume{40}\acmNumber{4}\acmArticle{91}\acmMonth{8} \acmDOI{10.1145/3450626.3459817}

\begin{document}
\title{Discovering Diverse Athletic Jumping Strategies}

\author{Zhiqi Yin}
\affiliation{
  \institution{Simon Fraser University}
     \city{Burnaby}
   \country{Canada}
}
\email{zhiqi_yin@sfu.ca}
\author{Zeshi Yang}
\affiliation{
  \institution{Simon Fraser University}
   \city{Burnaby}
   \country{Canada}
}
\email{zeshi_yang@sfu.ca}
\author{Michiel van de Panne}
\affiliation{
  \institution{University of British Columbia}
   \city{Vancouver}
   \country{Canada}
}
\email{van@cs.ubc.ca}
\author{KangKang Yin}
\affiliation{
  \institution{Simon Fraser University}
   \city{Burnaby}
   \country{Canada}
}
\email{kkyin@sfu.ca}

\newcommand*{\clip}{\mathop{\mathrm{Clip}}\nolimits}
\newcommand*{\diag}{\mathop{\mathrm{diag}}\nolimits}

\begin{teaserfigure}
 \centering
 \begin{subfigure}[b]{0.46\linewidth}
    \includegraphics[width=0.99\linewidth]{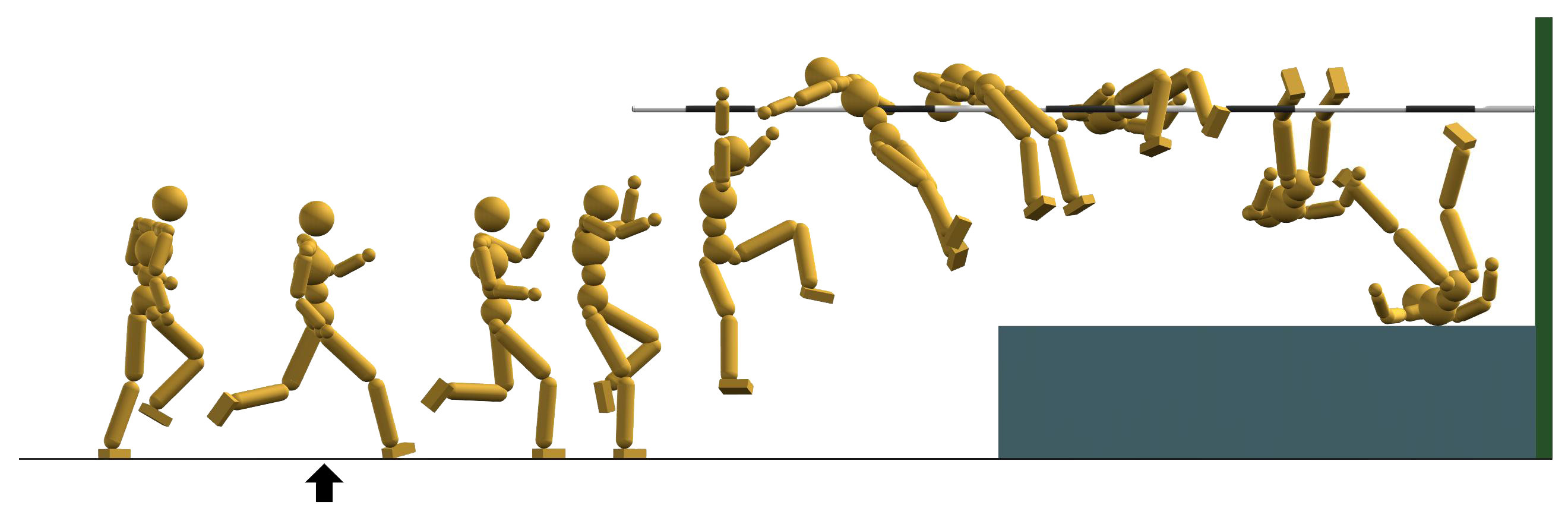}
    \caption{Fosbury Flop -- max height=$200cm$}
    \label{fig:highjump-flop}
 \end{subfigure}
 \begin{subfigure}[b]{0.53\linewidth}
    \includegraphics[width=0.99\linewidth]{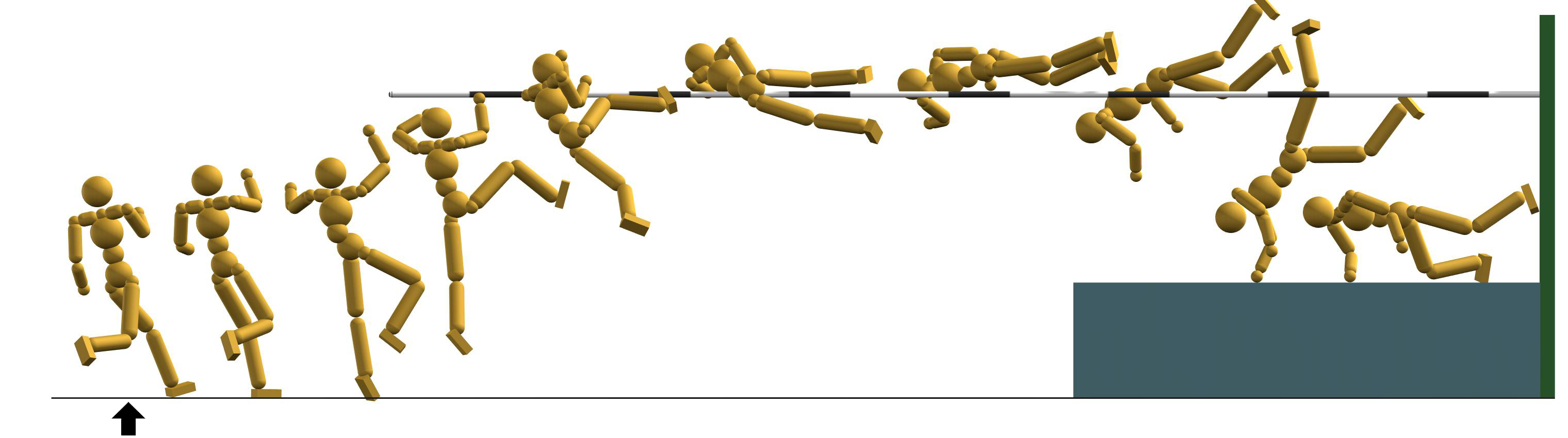}
     \caption{Western Roll (facing sideways) -- max height=$195cm$}
    \label{fig:highjump-rollside}
 \end{subfigure}
 \caption{Two of the eight high jump strategies discovered by our two-stage learning framework, as achieved by physics-based control policies. The first stage is a sample-efficient Bayesian diversity search algorithm that explores the space of take-off states, as indicated by the black arrows. In the second stage we explicitly encourage novel policies given a fixed initial state discovered in the first stage.} 
\label{fig:teaser}
\end{teaserfigure}

\begin{abstract}
We present a framework that enables the discovery of diverse and natural-looking motion strategies for athletic skills such as the high jump. The strategies are realized as control policies for physics-based characters. Given a task objective and an initial character configuration, the combination of physics simulation and deep reinforcement learning (DRL) provides a suitable starting point for automatic control policy training. To facilitate the learning of realistic human motions, we propose a Pose Variational Autoencoder (P-VAE) to constrain the actions to a subspace of natural poses. In contrast to motion imitation methods, a rich variety of novel strategies can naturally emerge by exploring initial character states through a sample-efficient Bayesian diversity search (BDS) algorithm. A second stage of optimization that encourages novel policies can further enrich the unique strategies discovered. Our method allows for the discovery of diverse and novel strategies for athletic jumping motions such as high jumps and obstacle jumps with no motion examples and less reward engineering than prior work.
\end{abstract}

%
%


%
%


\maketitle

\section{Introduction}

Athletic endeavors are a function of strength, skill, and strategy. For the high-jump, the choice of strategy has been of particular historic importance. Innovations in techniques or strategies have repeatedly redefined world records over the past 150 years, culminating in the now well-established Fosbury flop (Brill bend) technique. In this paper, we demonstrate how to discover diverse strategies, as realized through physics-based controllers which are trained using reinforcement learning. We show that natural high-jump strategies can be learned without recourse to motion capture data, with the exception of a single generic run-up motion capture clip. We further demonstrate diverse solutions to a box-jumping task.

Several challenges stand in the way of being able to discover iconic athletic strategies such as those used for the high jump. The motions involve high-dimensional states and actions. The task is defined by a sparse reward, i.e., successfully making it over the bar or not. It is not obvious how to ensure that the resulting motions are natural in addition to being physically plausible. Lastly, the optimization landscape is multimodal in nature.

We take several steps to address these challenges. First, we identify the take-off state as a strong determinant of the resulting jump strategy, which is characterized by low-dimensional features such as the net angular velocity and approach angle in preparation for take-off. To efficiently explore the take-off states, we employ Bayesian diversity optimization. Given a desired take-off state, we first train a run-up controller that imitates a single generic run-up motion capture clip while also targeting the desired take-off state. The subsequent jump control policy is trained with the help of a curriculum, without any recourse to motion capture data. We make use of a pose variational autoencoder to define an action space that helps yield more natural poses and motions. We further enrich unique strategy variations by a second optimization stage which reuses the best discovered take-off states and encourages novel control policies.

\newpage
In summary, our contributions include:
\begin{itemize}
    \item A system which can discover common athletic high jump strategies,
        and execute them using learned controllers and physics-based simulation. The discovered strategies include the Fosbury flop (Brill bend), Western Roll, and a number of other styles. We further evaluate the system on box jumps and on a number of high-jump variations and ablations.
    \item The use of Bayesian diversity search for sample-efficient exploration of take-off states, which are strong determinants of resulting strategies.
    \item Pose variational autoencoders used in support of learning natural athletic motions.
\end{itemize}

\section{RELATED WORK}

We build on prior work from several areas,
including character animation, diversity optimization, human pose modeling, and high-jump analysis from biomechanics and kinesiology.

\subsection{Character Animation}
Synthesizing natural human motion is a long-standing challenge in computer animation. We first briefly review kinematic methods, and then provide a more detailed review of physics-based methods. To the best of our knowledge, there are no previous attempts to synthesize athletic high jumps or obstacle jumps using either kinematic or physics-based approaches. Both tasks require precise coordination and exhibit multiple strategies.

\paragraph{Kinematic Methods} 
Data-driven kinematic methods have demonstrated their effectiveness for synthesizing high-quality human motions based on captured examples. Such kinematic models have evolved from graph structures~\cite{Kovar:2002:MotionGraphs,Safonova-Interpolated-Motion-Graphs}, to Gaussian Processes~\cite{levine2012continuous, ye2010synthesis-responsive}, and recently deep neural networks \cite{Holden:2017:PFNN, Zhang:2018:MANN, starke2019neural-state-machine, starke2020local-motion-phases, lee2018interactive-RNN, Ling20}. Non-parametric models that store all example frames have limited capability of generalizing to new motions due to their inherent nature of data interpolation \cite{Clavet16}. Compact parametric models learn an underlying low-dimensional motion manifold. Therefore they tend to generalize better as new motions not in the training dataset can be synthesized by sampling in the learned latent space~\cite{Holden16}. Completely novel motions and strategies, however, are still beyond their reach. Most fundamentally, kinematic models do not take into account physical realism, which is important for athletic motions. We thus cannot directly apply kinematic methods to our problem of discovering unseen strategies for highly dynamic motions. However, we do adopt a variational autoencoder (VAE) similar to the one in~\cite{Ling20} as a means to
improve the naturalness of our learned motion strategies.

\paragraph{Physics-based Methods}
Physics-based control and simulation methods generate motions with physical realism and environmental interactions. The key challenge is the design or learning of robust controllers. Conventional manually designed controllers have achieved significant success for locomotion, e.g.,~\cite{Yin07,wang2009optimizing,Coros10,wang2012optimizing,geijtenbeek2013flexible,lee2014locomotion,felis2016synthesis,jain2009optimization}. The seminal work from Hodgins \textit{et al.} demonstrated impressive controllers for athletic skills such as a handspring vault, a standing broad jump, a vertical leap, somersaults to different directions, and platform dives~\cite{Hodgins95,Wooten98}. Such handcrafted controllers are mostly designed with finite state machines (FSM) and heuristic feedback rules, which require deep human insight and domain knowledge, and tedious manual trial and error. Zhao and van de Panne~\shortcite{Zhao05} thus proposed an interface to ease such a design process, and demonstrated controllers for diving, skiing and snowboarding. Controls can also be designed using objectives and constraints adapted to each motion phase, e.g.,~\cite{jain2009optimization,deLasa2010},
or developed using a methodology that mimics human coaching~\cite{ha2014-human-coach}. In general, manually designed controllers remain hard to generalize to different strategies or tasks. 

With the wide availability of motion capture data, many research endeavors have been focused on tracking-based controllers, which are capable of reproducing high-quality motions by imitating motion examples. Controllers for a wide range of skills have been demonstrated through trajectory optimization \cite{sok2007simulating,da2008interactive,muico2009contact,lee2010data-driven-biped,ye2010optimal,lee2014locomotion}, sampling-based algorithms \cite{Liu:2010:Samcon,Liu:2015:Samcon2,Liu16}, and deep reinforcement learning ~\cite{Peng2017deepLoco,Peng:2018:DeepMimic,Peng:2018:SFV,ma2021spacetime,Liu18,Lee19}.
Tracking controllers have also been combined with kinematic motion generators to support interactive control of simulated characters~\cite{Bergamin:2019:DReCon,Park:2019:LPS,won2020scalable}. Even though tracking-based methods have demonstrated their effectiveness on achieving task-related goals~\cite{Peng:2018:DeepMimic}, the imitation objective inherently restricts them from generalizing to novel motion strategies fundamentally different from the reference. Most recently, style exploration has also been demonstrated within a physics-based DRL framework using spacetime bounds~\cite{ma2021spacetime}. However, these remain style variations rather than strategy variations. Moreover, high jumping motion capture examples are difficult to find. We obtained captures of three high jump strategies, which we use to compare our synthetic results to.

Our goal is to discover as many strategies as possible, so example-free methods are most suitable in our case. Various tracking-free methods have been proposed via trajectory optimization or deep reinforcement learning. Heess \textit{et al.} \shortcite{Heess:2017:Emergence} demonstrate a rich set of locomotion behaviors emerging from just complex environment interactions. However, the resulting motions show limited realism in the absence of effective motion quality regularization. Better motion quality is achievable with sophisticated reward functions and domain knowledge, such as sagittal symmetry, which do not directly generalize beyond locomotion \cite{Yu:2018:LSLL,coumans2016pybullet,Xie2020allsteps,mordatch2013-cio-locomotion,mordatch2015interactive}. Synthesizing diverse physics-based skills without example motions generally requires optimization with detailed cost functions that are engineered specifically for each skill~\cite{AlBorno13}, and often only works for simplified physical models~\cite{Mordatch12}.

\subsection{Diversity Optimization}
Diversity Optimization is a problem of great interest in artificial intelligence \cite{hebrard2005finding-diverse,ursem2002diversity,srivastava2007diverse-plan,coman2011generating-diverse,pugh2016quality-diversity,lehman2011novelty-search}. It is formulated as searching for a set of configurations such that the corresponding outcomes have a large diversity while satisfying a given objective. Diversity optimization has also been utilized in computer graphics applications~\cite{merrell2011interactive,Agrawal:2013:Diverse}. For example, a variety of 2D and simple 3D skills have been achieved through jointly optimizing task objectives and a diversity metric within 
a trajectory optimization framework~\cite{Agrawal:2013:Diverse}. Such methods are computationally prohibitive for our case as learning the athletic tasks involve expensive DRL training through non-differentiable simulations, e.g., a single strategy takes six hours to learn even on a high-end desktop. 
We propose a diversity optimization algorithm based on the successful Bayesian Optimization (BO) philosophy for sample efficient black-box function optimization.

In Bayesian Optimization, objective functions are optimized purely through function evaluations as no derivative information is available. A Bayesian statistical \textit{surrogate model}, usually a Gaussian Process (GP) \cite{rasmussen2003gaussian-process}, is maintained to estimate the value of the objective function along with the uncertainty of the estimation. An \textit{acquisition function} is then repeatedly maximized for fast decisions on where to sample next for the actual expensive function evaluation. The next sample needs to be promising in terms of maximizing the objective function predicted by the surrogate model, and also informative in terms of reducing the uncertainty in less explored regions of the surrogate model~\cite{jones1998-bo-expected-improvement, frazier2009-bo-knowledge-gradient, GP-bandit}. BO has been widely adopted in machine learning for parameter and hyperparameter optimizations \cite{snoek2015scalable,klein2017fast,kandasamy2018neural,kandasamy2020tuning,korovina2020chembo,snoek2012practical}. Recently BO has also seen applications in computer graphics~\cite{koyama2017sequential,koyama2020sequential}, such as parameter tuning for fluid animation systems \cite{brochu2007preference}. 

We propose a novel acquisition function to encourage discovery of diverse motion strategies. We also decouple the exploration from the maximization for more robust and efficient strategy discovery. We name this algorithm Bayesian Diversity Search (BDS). The BDS algorithm searches for diverse strategies by exploring a low-dimensional initial state space defined at the take-off moment. Initial states exploration has been applied to find appropriate initial conditions for desired landing controllers \cite{ha2012falling}. In the context of DRL learning, initial states are usually treated as hyperparameters rather than being explored.

Recently a variety of DRL-based learning methods have been proposed to discover diverse control policies in machine learning, e.g.,  \cite{Eysenbach19,zhang2019novel-policies,Sun2020novel-policies,achiam2018variational,sharma2019dynamics,haarnoja2018soft,conti2018improving,houthooft2016vime,hester2017intrinsically,schmidhuber1991curious}. These methods mainly encourage exploration of unseen states or actions by jointly optimizing the task and novelty objectives~\cite{zhang2019novel-policies}, or optimizing intrinsic rewards such as heuristically defined curiosity terms~\cite{Eysenbach19,sharma2019dynamics}. We adopt a similar idea for novelty seeking in Stage~2 of our framework after BDS, but with a novelty metric and reward structure more suitable for our goal. Coupled with the Stage~1 BDS, we are able to learn a rich set of strategies for challenging tasks such as athletic high jumping.

\subsection{Natural Pose Space}

In biomechanics and neuroscience, it is well known that muscle synergies, or muscle co-activations, serve as motor primitives for the central nervous system to simplify movement control of the underlying complex neuromusculoskeletal systems~\cite{Overduin15,Zhao19}. In character animation, human-like character models are much simplified, but are still parameterized by 30+ DoFs. Yet the natural human pose manifold learned from motion capture databases is of much lower dimension \cite{Holden16}. The movement of joints are highly correlated as typically they are strategically coordinated and co-activated. Such correlations have been modelled through traditional dimensionality reduction techniques such as PCA \cite{chai2005performance-lowd}, or more recently, Variational AutoEncoders (VAE) \cite{habibie2017,Ling20}.

We rely on a VAE learned from mocap databases to produce natural target poses for our DRL-based policy network. Searching behaviors in low dimensional spaces has been employed in physics-based character animation to both accelerate the nonlinear optimization and improve the motion quality \cite{Safonova:2004:OptPCA}. Throwing motions based on muscle synergies extracted from human experiments have been synthesized on a musculoskeletal model \cite{cruz2017synergy}. Recent DRL methods either directly imitate mocap examples \cite{Peng:2018:DeepMimic, won2020scalable}, which makes strategy discovery hard if possible; or adopt a \textit{de novo} approach with no example at all \cite{Heess2015learning}, which often results in extremely unnatural motions for human like characters. Close in spirit to our work is \cite{ranganath2019lowd-joint-coactivation}, where a low-dimensional PCA space learned from a single mocap trajectory is used as the action space of DeepMimic for tracking-based control. We aim to discover new strategies without tracking, and we use a large set of generic motions to deduce a task-and-strategy-independent natural pose space. We also add action offsets to the P-VAE output poses so that large joint activation can be achieved for powerful take-offs.

{Reduced or latent parameter spaces based on statistical analysis of poses have been used for grasping control \cite{Ciocarlie10,Andrews13,Osa18}. A Trajectory Generator (TG) can provide a compact parameterization that can enable learning of reactive policies for complex behaviors~\cite{Iscen18}.
Motion primitives can also be learned from mocap and then be composed to learn new behaviors~\cite{Peng19}.}

\subsection{History and Science of High Jump}

The high jump is one of the most technically complex, strategically nuanced, and physiologically demanding sports among all track and field events \cite{Donnelly14}. Over the past 100 years, high jump has evolved dramatically in the Olympics. Here we summarize the well-known variations \cite{SevenStyles,Donnelly14}, and we refer readers to our supplemental video for more visual illustrations.
\begin{itemize}
    \item {The Hurdle}: the jumper runs straight-on to the bar, raises one leg up to the bar, and quickly raises the other one over the bar once the first has cleared. The body clears the bar upright.
    \item {Scissor Kick}: the jumper approaches the bar diagonally, throws first the inside leg and then the other over the bar in a scissoring motion. The body clears the bar upright.
    \item {Eastern Cutoff}: the jumper takes off like the scissor kick, but extends his back and flattens out over the bar.  
    \item {Western Roll}:the jumper also approaches the bar diagonally, but the inner leg is used for the take-off, while the outer leg is thrust up to lead the body sideways over the bar.
    \item {The Straddle}: similar to Western Roll, but the jumper clears the bar face-down.
    \item {Fosbury Flop}: The jumper approaches the bar on a curved path and leans away from the bar at the take-off point to convert horizontal velocity into vertical velocity and angular momentum. In flight, the jumper progressively arches their shoulders, back, and legs in a rolling motion, and lands on their neck and back. The jumper's Center of Mass (CoM) can pass under the bar while the body arches and slide above the bar. It has been the favored high jump technique in Olympic competitions since used by Dick Fosbury in the 1968 Summer Olympics. It was concurrently developed by Debbie Brill.
\end{itemize}

In biomechanics, kinesiology, and physical education, high jumps have been analyzed to a limited extent. We adopt the force limits reported in \cite{Okuyama03} in our simulations. Dapena simulated a higher jump by making small changes to a recorded jump \cite{Dapena02}. Mathematical models of the Center of Mass (CoM) movement have been developed to offer recommendations to increase the effectiveness of high jumps~\cite{Adashevskiy13}.

\section{SYSTEM OVERVIEW}
\begin{figure}[t]
    \centering
    \includegraphics[width=\linewidth]{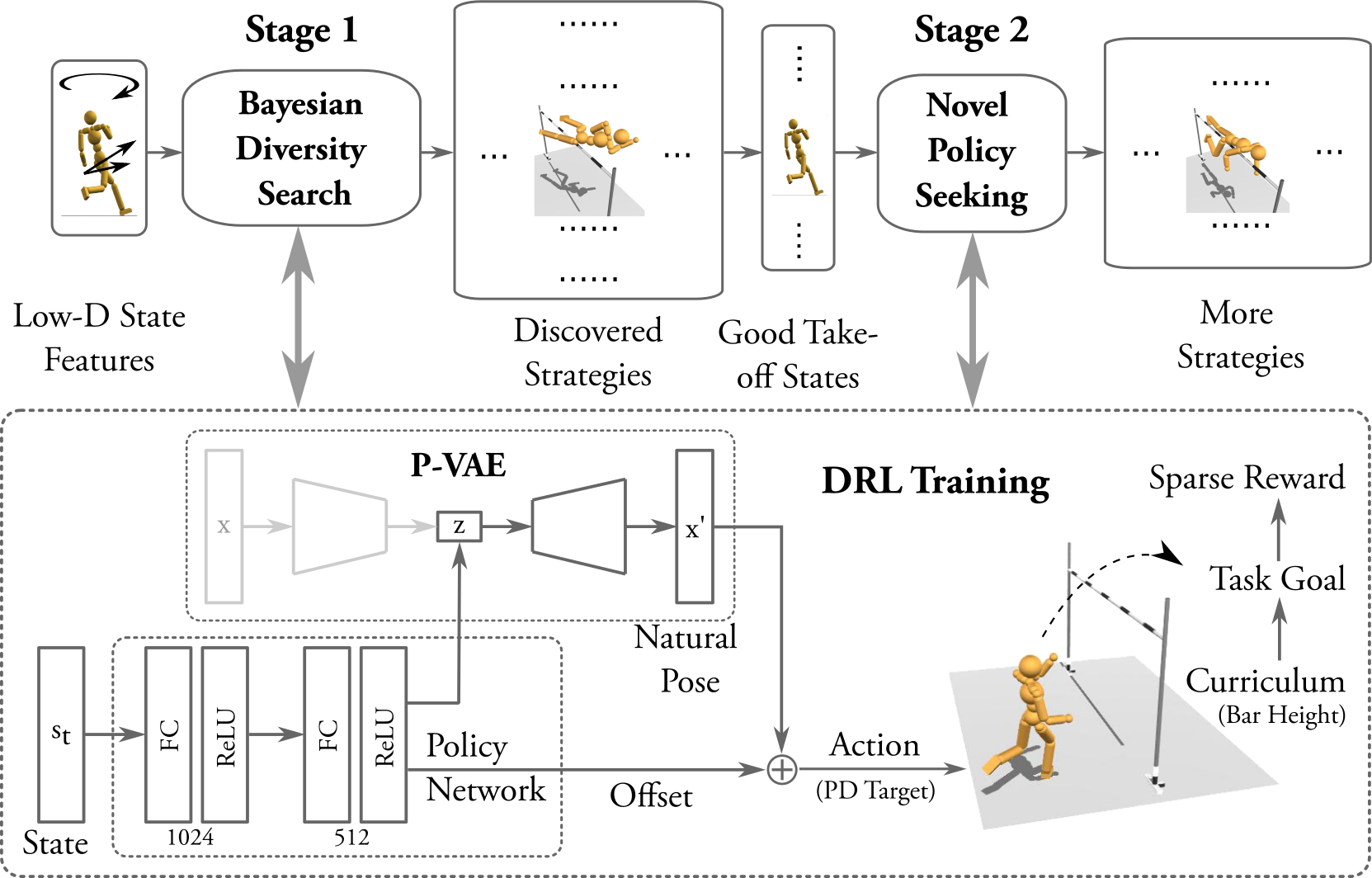}
    \caption{{Overview of our strategy discovery framework. The Stage 1 Bayesian Diversity Search algorithm explores a low-dimensional feature vector of the take-off states to discover multiple jumping strategies. The output strategies from Stage 1 and their corresponding take-off states are taken as input to warm-start further training in Stage 2, which encourages novel policies that lead to additional visually distinct strategies. Both stages share the same DRL training component, which utilizes a P-VAE to improve the motion quality and a task curriculum to gradually increase the learning difficulty.}}
    \label{fig:overview}
\end{figure}

We now give an overview of our learning framework as illustrated in Figure~\ref{fig:overview}. Our framework splits athletic jumps into two phases: a run-up phase and a jump phase. The {\em take-off state} marks the transition between these two phases, and consists of a time instant midway through the last support phase before becoming airborne. The take-off state is key to our exploration strategy, as it is a strong determinant of the resulting jump strategy. We characterize the take-off state by a feature vector that captures key aspects of the state, such as the net angular velocity and body orientation. This defines a low-dimensional take-off feature space that we can sample in order to explore and evaluate a variety of motion strategies. While random sampling of take-off state features is straightforward, it is computationally impractical as evaluating one sample involves an expensive DRL learning process that takes hours even on modern machines. Therefore, we introduce a sample-efficient Bayesian Diversity Search (BDS) algorithm as a key part of our Stage~1 optimization process.

Given a specific sampled take-off state, we then need to produce an optimized run-up controller and a jump controller that result in the best possible corresponding jumps. This process has several steps. We first train a {\em }run-up controller, using deep reinforcement learning, that imitates a single generic run-up motion capture clip while also targeting the desired take-off state. For simplicity, the run-up controller and its training are not shown in Figure~\ref{fig:overview}. These are discussed in Section~\ref{sec:Experiments-Runup}. The main challenge lies with the synthesis of the actual jump controller which governs the remainder of the motion, and for which we wish to discover strategies without any recourse to known solutions.

The jump controller begins from the take-off state and needs to control the body during take-off, over the bar, and to prepare for landing. This poses a challenging learning problem because of the demanding nature of the task, the sparse fail/success rewards, and the difficulty of also achieving natural human-like movement. We apply two key insights to make this task learnable using deep reinforcement learning. First, we employ an action space defined by a subspace of natural human poses as modeled with a Pose Variational Autoencoder (P-VAE). Given an action parameterized as a target body pose, individual joint torques are then realized using PD-controllers. We additionally allow for regularized {\em offset} PD-targets that are added to the P-VAE targets to enable strong takeoff forces. Second, we employ a curriculum that progressively increases the task difficulty, i.e., the height of the bar, based on current performance.

A diverse set of strategies can already emerge after the Stage 1 BDS optimization. To achieve further strategy variations, we reuse the take-off states of the existing discovered strategies for another stage of optimization. The diversity is explicitly incentivized during this Stage 2 optimization via a novelty reward, which is focused specifically on features of the body pose at the peak height of the jump. As shown in Figure~\ref{fig:overview}, Stage~2 makes use of the same overall DRL learning procedure as in Stage~1, albeit with a slightly different reward structure.

\section{LEARNING NATURAL STRATEGIES}
\label{sec:discovery}
Given a character model, an environment, and a task objective, we aim to learn feasible natural-looking motion strategies using deep reinforcement learning. We first describe our DRL formulation in Section~\ref{sec:methods-DRL-formulation}. To improve the learned motion quality, we propose a Pose Variational Autoencoder (P-VAE) to constrain the policy actions in Section~\ref{sec:methods-PVAE}.

\subsection{DRL Formulation}
\label{sec:methods-DRL-formulation}
Our strategy learning task is formulated as a standard reinforcement learning problem, where the character interacts with the environment to learn a control policy which maximizes a long-term reward. The control policy $\pi_\theta(a|s)$ parameterized by $\theta$ models the conditional distribution over action $a \in \mathcal{A}$ given the character state $s \in \mathcal{S}$. At each time step $t$, the character interacts with the environment with action $a_t$ sampled from $\pi(a|s)$ based on the current state $s_t$. The environment then responds with a new state $s_{t+1}$ according to the transition dynamics $p(s_{t+1}|s_t,a_t)$, along with a reward signal $r_t$. The goal of reinforcement learning is to learn the optimal policy parameters $\theta^*$ which maximizes the expected return defined as
\begin{equation}
    J(\theta) = \mathbb{E}_{\tau\sim p_\theta(\tau)}\left[
        \sum_{t=0}^T{\gamma^t r_t}
    \right],
\end{equation}
where $T$ is the episode length, $\gamma \leq 1$ is a discount factor, and $p_\theta(\tau)$ is the probability of observing trajectory $\tau = \{s_0, a_0, s_1, ..., a_{T-1}, s_T\}$ given the current policy $\pi_\theta(a|s)$.

\paragraph{States}
The state $s$ describes the character configuration. We use a similar set of pose and velocity features as those proposed in DeepMimic \cite{Peng:2018:DeepMimic}, including relative positions of each link with respect to the root, their rotations parameterized in quaternions, along with their linear and angular velocities. Different from DeepMimic, our features are computed directly in the global frame without direction-invariant transformations for the studied jump tasks. The justification is that input features should distinguish states with different relative transformations between the character and the environment obstacle such as the crossbar. In principle, we could also use direction-invariant features as in DeepMimic, and include the relative transformation to the obstacle into the feature set. However, as proved in \cite{Ma19}, there are no direction-invariant features that are always singularity free. Direction-invariant features change wildly whenever the character's facing direction approaches the chosen motion direction, which is usually the global up-direction or the $Y$-axis. For high jump techniques such as the Fosbury flop, singularities are frequently encountered as the athlete clears the bar facing upward. Therefore, we opt to use global features for simplicity and robustness. Another difference from DeepMimic is that time-dependent phase variables are not included in our feature set. Actions are chosen purely based on the dynamic state of the character.

\paragraph{Initial States}
\label{sec:methods-initial-states}
The initial state $s_0$ is the state in which an agent begins each episode in DRL training. We explore a chosen low-dimensional feature space ($3\sim4D$) of the take-off states for learning diverse jumping strategies. As shown by previous work \cite{ma2021spacetime}, the take-off moment is a critical point of jumping motions, where the volume of the feasible region of the dynamic skill is the smallest. In another word, bad initial states will fail fast, which in a way help our exploration framework to find good ones quicker. Alternatively, we could place the agent in a fixed initial pose to start with, such as a static pose before the run-up. This is problematic for several reasons. First, different jumping strategies need different length for the run-up. The planar position and facing direction of the root is still a three dimensional space to be explored. Second, the run-up strategies and the jumping strategies do not correlate in a one-to-one fashion. Visually, the run-up strategies do not look as diverse as the jumping strategies. Lastly, starting the jumps from a static pose lengthens the learning horizon, and makes our learning framework based on DRL training even more costly. Therefore we choose to focus on just the jumping part of the jumps in this work, and leave the run-up control learning to DeepMimic, which is one of the state-of-the-art imitation-based DRL learning methods. More details are given in Section~\ref{sec:Experiments-Runup}.    

\paragraph{Actions}
The action $a$ is a target pose described by internal joint rotations. We parameterize 1D revolute joint rotations by scalar angles, and 3D spherical joint rotations by exponential maps \cite{Grassia:1998:ExpMap}. Given a target pose and the current character state, joint torques are computed through the Stable Proportional Derivative (SPD) controllers \cite{Tan:2011:SPD}. Our control frequency $f_{\text{control}}$ ranges from 10 $Hz$ to 30 $Hz$ depending on both the task and the curriculum. For challenging tasks like high jumps, it helps to quickly improve initial policies through stochastic evaluations at early training stages. A low-frequency policy enables faster learning by reducing the needed control steps, or in another word, the dimensionality and complexity of the actions $(a_0,...,a_T)$. This is in spirit similar to the 10 $Hz$ control fragments used in SAMCON-type controllers \cite{Liu16}. Successful low-frequency policies can then be gradually transferred to high-frequency ones according to a curriculum to achieve finer controls and thus smoother motions. We discuss the choice of control frequency in more detail in Section~\ref{sec:Experiments-Curriculum}.
\paragraph{Reward}
We use a reward function consisting of the product of two terms for all our strategy discovery tasks as follows:
\begin{equation}
\label{eq:stage1-reward}
    r = r_{\text{task}} \cdot r_{\text{naturalness}} 
\end{equation}    
where $r_{\text{task}}$ is the task objective and $r_{\text{naturalness}}$ is a naturalness reward term computed from the P-VAE to be described in Section \ref{sec:methods-PVAE}. For diverse strategy discovery, a simple $r_{\text{task}}$ which precisely captures the task objective is preferred. For example in high jumping, the agent receives a sparse reward signal at the end of the jump after it successfully clears the bar. In principle, we could transform the sparse reward into a dense reward to reduce the learning difficulty, such as to reward CoM positions higher than a parabolic trajectory estimated from the bar height. However in practice, such dense guidance reward can mislead the training to a bad local optimum, where the character learns to jump high in place rather than clears the bar in a coordinated fashion. Moreover, the CoM height and the bar height may not correlate in a simple way. For example, the CoM passes underneath the crossbar in Fosbury flops. As a result, a shaped dense reward function could harm the diversity of the learned strategies. We will discuss reward function settings for each task in more details in Section~\ref{sec:Experiments-Reward}.

\paragraph{Policy Representation}
We use a fully-connected neural network parameterized by weights $\theta$ to represent the control policy $\pi_\theta(a|s)$. Similar to the settings in \cite{Peng:2018:DeepMimic}, the network has two hidden layers with $1024$ and $512$ units respectively. ReLU activations are applied for all hidden units. Our policy maps a given state $s$ to a Gaussian distribution over actions $a=\mathcal{N}(\mu(s), \Sigma)$. The mean $\mu(s)$ is determined by the network output. The covariance matrix $\Sigma=\sigma I$ is diagonal, where $I$ is the identity matrix and $\sigma$ is a scalar variable measuring the action noise. We apply an annealing strategy to linearly decrease $\sigma$ from $0.5$ to $0.1$ in the first $1.0\times 10^7$ simulation steps, to encourage more exploration in early training and more exploitation in late training.

\paragraph{Training}
We train our policies with the Proximal Policy Optimization (PPO) method \cite{Schulman:2017:PPO}. PPO involves training both a policy network and a value function network. The value network architecture is similar to the policy network, except that there is only one single linear unit in the output layer. We train the value network with TD($\lambda$) multi-step returns. We estimate the advantage of the PPO policy gradient by the Generalized Advantage Estimator GAE($\lambda$) \cite{Generalized-Advantage-Estimation}.

\subsection{Pose Variational Autoencoder}\label{sec:methods-PVAE}
The dimension of natural human poses is usually much lower than the true degrees of freedom of the character model. We propose a generative model to produce natural PD target poses at each control step. More specifically, we train a Pose Variational Autoencoder (P-VAE) from captured natural human poses, and then sample its latent space to produce desired PD target poses for control. Here a pose only encodes internal joint rotations without the global root transformations. We use publicly available human motion capture databases to train our P-VAE. Note that none of these databases consist of high jumps or obstacle jumps specifically, but they already provide enough poses for us to learn the natural human pose manifold successfully.

\paragraph{P-VAE Architecture and Training}
Our P-VAE adopts the standard Beta Variational Autoencoder ($\beta$-VAE) architecture \cite{beta-VAE}. The encoder maps an input feature $x$ to a low-dimensional latent space, parameterized by a Gaussian distribution with a mean $\mu_x$ and a diagonal covariance $\Sigma_x$. The decoder maps a latent vector sampled from the Gaussian distribution back to the original feature space as $x'$. The training objective is to minimize the following loss function:
\begin{equation}
    \mathcal{L} = \mathcal{L}_\text{MSE}(x,x') + \beta \cdot \text{KL}(\mathcal{N}(\mu_x, \Sigma_x), \mathcal{N}(0, I)),
\end{equation}
where the first term is the MSE (Mean Squared Error) reconstruction loss, and the second term shapes the latent variable distribution to a standard Gaussian by measuring their Kulback-Leibler divergence. We set $\beta = 1.0\times10^{-5}$ in our experiments, so that the two terms in the loss function are within the same order of magnitude numerically.

We train the P-VAE on a dataset consisting of roughly $20,000$ poses obtained from the CMU and SFU motion capture databases. We include a large variety of motion skills, including walking, running, jumping, breakdancing, cartwheels, flips, kicks, martial arts, etc. The input features consist of all link and joint positions relative to the root in the local root frames, and all joint rotations with respect to their parents. We parameterize joint rotations by a $6$D representation for better continuity, as described in \cite{zhou2019continuity, Ling20}.

We model both the encoder and the decoder as fully connected neural networks with two hidden layers, each having 256 units with $tanh$ activation. We perform PCA (Principal Component Analysis) on the training data and choose $d_{\text{latent}} = 13$ to cover $85\%$ of the training data variance, where $d_{\text{latent}}$ is the dimension of the latent variable. We use the Adam optimizer to update network weights \cite{kingma2014adam}, with the learning rate set to $1.0\times10^{-4}$. Using a mini-batch size of $128$, the training takes $80$ epochs within $2$ minutes on an NVIDIA GeForce GTX 1080 GPU and an Intel i7-8700k CPU. We use this single pre-trained P-VAE for all our strategy discovery tasks to be described.

\paragraph{Composite PD Targets}
PD controllers provide actuation based on positional errors. So in order to reach the desired pose, the actual target pose needs to be offset by a certain amount. Such offsets are usually small to just counter-act the gravity for free limbs. However, for joints that interact with the environment, such as the lower body joints for weight support and ground takeoff, large offsets are needed to generate powerful ground reaction forces to propel the body forward or into the air. Such complementary offsets combined with the default P-VAE targets help realize natural poses during physics-based simulations. Our action space $\mathcal{A}$ is therefore $d_{\text{latent}} + d_{\text{offset}}$ dimensional, where $d_{\text{latent}}$ is the dimension of the P-VAE latent space, and $d_{\text{offset}}$ is the dimension of the DoFs that we wish to apply offsets for. We simply apply offsets to all internal joints in this work. Given $a=(a_{\text{latent}},a_{\text{offset}}) \in \mathcal{A}$ sampled from the policy $\pi_\theta(a|s)$, where $a_{\text{latent}}$ and $a_{\text{offset}}$ correspond to the latent and offset part of $a$ respectively, the final PD target is computed by $D_{\text{pose}}(a_{\text{latent}}) + a_{\text{offset}}$. Here $D_{\text{pose}}(\cdot)$ is a function that decodes the latent vector $a_{\text{latent}}$ to full-body joint rotations. We minimize the usage of rotation offsets by a penalty term as follows:
\begin{equation}
\label{eq:r-pvae}
    r_{\text{naturalness}} = 1 - \clip\left(\left(\frac{||a_{\text{offset}}||_1}{c_{\text{offset}}}\right)^2, 0, 1\right) ,
\end{equation}
where $c_{\text{offset}}$ is the maximum offset allowed. For tasks with only a sparse reward signal at the end, $||a_{\text{offset}}||_1$ in Equation~\ref{eq:r-pvae} is replaced by the average offset norm $\frac{1}{T}\sum_{t=0}^T{||a^{(t)}_{\text{offset}}||_1}$ across the entire episode. We use $L1$-norm rather than the commonly adopted $L2$-norm to encourage sparse solutions with fewer non-zero components \cite{tibshirani1996regression, chen2001atomic}, as our goal is to only apply offsets to essential joints to complete the task while staying close to the natural pose manifold prescribed by the P-VAE.
\section{LEARNING DIVERSE STRATEGIES}
\label{sec:diverse}
Given a virtual environment and a task objective, we would like to discover as many strategies as possible to complete the task at hand. Without human insights and demonstrations, this is a challenging task. To this end, we propose a two-stage framework to enable stochastic DRL to discover solution modes such as the Fosbury flop.

The first stage focuses on strategy discovery by exploring the space of initial states. For example in high jump, the Fosbury flop technique and the straddle technique require completely different initial states at take-off, in terms of the approaching angle with respect to the bar, the take-off velocities, and the choice of inner or outer leg as the take-off leg. A fixed initial state may lead to success of one particular strategy, but can miss other drastically different ones. We systematically explore the initial state space through a novel sample-efficient Bayesian Diversity Search (BDS) algorithm to be described in Section~\ref{sec:methods-bayesian-diversity-search}.

The output of Stage 1 is a set of diverse motion strategies and their corresponding initial states. Taken such a successful initial state as input, we then apply another pass of DRL learning to further explore more motion variations permitted by the same initial state. The intuition is to explore different local optima while maximizing the novelty of the current policy, compared to previously found ones. We describe our detailed settings for the Stage 2 novel policy seeking algorithm in Section~\ref{sec:methods-phase2}.

\subsection{Stage 1: Initial States Exploration with Bayesian Diversity Search}\label{sec:methods-bayesian-diversity-search}

In Stage 1, we perform diverse strategy discovery by exploring initial state variations, such as pose and velocity variations, at the take-off moment. We first extract a feature vector $f$ from a motion trajectory to characterize and differentiate between different strategies. A straightforward way is to compute the Euclidean distance between time-aligned motion trajectories, but we hand pick a low-dimensional visually-salient feature set as detailed in Section~\ref{sec:Experiments-Strategy-Features}. We also define a low-dimensional exploration space $\mathcal{X}$ for initial states, as exploring the full state space is computationally prohibitive. Our goal is to search for a set of representatives $X_n = \{x_1, x_2, ..., x_n | x_i \in \mathcal{X}\}$, such that the corresponding feature set $F_n= \{f_1, f_2, ..., f_n | f_i \in \mathcal{F}\}$ has a large diversity. Note that as DRL training and physics-based simulation are involved in producing the motion trajectories from an initial state, the computation of $f_i=g(x_i)$ is a stochastic and expensive black-box function. We therefore design a sample-efficient Bayesian Optimization (BO) algorithm to optimize for motion diversity in a guided fashion.

Our BDS (Bayesian Diversity Search) algorithm iteratively selects the next sample to evaluate from $\mathcal{X}$, given the current set of observations $X_t = \{x_1, x_2, ..., x_t\}$ and $F_t = \{f_1, f_2, ..., f_t\}$. More specifically, the next point $x_{t+1}$ is selected based on an acquisition function $a(x_{t+1})$ to maximize the diversity in $F_{t+1}=F_t \cup \{f_{t+1}\}$. We choose to maximize the minimum distance between $f_{t+1}$ and all $f_i \in F_t$:
\begin{equation}\label{eq:diversity-objective}
    a(x_{t+1}) = \min_{f_i \in F_t}{||f_{t+1} - f_i||} .
\end{equation}
Since evaluating $f_{t+1}$ through $g(\cdot)$ is expensive, we employ a surrogate model to quickly estimate $f_{t+1}$, so that the most promising sample to evaluate next can be efficiently found through Equation~\ref{eq:diversity-objective}.

We maintain the surrogate statistical model of $g(\cdot)$ using a Gaussian Process (GP) \cite{rasmussen2003gaussian-process}, similar to standard BO methods. A GP contains a prior mean $m(x)$ encoding the prior belief of the function value, and a kernel function $k(x,x')$ measuring the correlation between $g(x)$ and $g(x')$. More details of our specific $m(x)$ and $k(x,x')$ are given in Section~\ref{sec:Experiments-GP-Priors-Kernels}. Hereafter we assume a one-dimensional feature space $\mathcal{F}$. Generalization to a multi-dimensional feature space is straightforward as multi-output Gaussian Process implementations are readily available, such as \cite{GPflow2020multioutput-gp}. Given $m(\cdot)$, $k(\cdot, \cdot)$, and current observations $\{X_t, F_t\}$, posterior estimation of $g(x)$ for an arbitrary $x$ is given by a Gaussian distribution with mean $\mu_{t}$ and variance $\sigma^2_{t}$ computed in closed forms:
\begin{equation}\label{eq:gp-mu-sigma}
    \begin{gathered}
        \mu_{t}(x) = k(X_t, x)^T (K + \eta^{2} I)^{-1} \left(F_t - m(x)\right) + m(x), \\
        \sigma_{t}^{2}(x) = k(x, x) + \eta^{2} - k(X_t, x)^T(K + \eta^{2}I)^{-1} k(X_t,x) ,
    \end{gathered}
\end{equation}
where $X_t \in \mathbb{R}^{t \times \text{dim}(\mathcal{X})}$, $F_t \in \mathbb{R}^{t}$, $K \in \mathbb{R}^{t \times t}, K_{i,j} = k(x_{i}, x_{j})$, and $k(X_t, x) = [k(x,x_{1}),k(x,x_{2}),...k(x,x_{t})]^T$. $I$ is the identity matrix, and $\eta$ is the standard deviation of the observation noise. Equation~\ref{eq:diversity-objective} can then be approximated by
\begin{equation}\label{eq:diversity-approximation}
    \hat{a}(x_{t+1}) = \mathbb{E}_{\hat{f}_{t+1} \sim \mathcal{N}(\mu_t(x_{t+1}), \sigma^2_t(x_{t+1}))}\left[\min_{f_i \in F_t}{||\hat{f}_{t+1} - f_i||}\right] .
\end{equation}
Equation~\ref{eq:diversity-approximation} can be computed analytically for one-dimensional features, but gets more and more complicated to compute analytically as the feature dimension grows, or when the feature space is non-Euclidean as in our case with rotational features. Therefore, we compute Equation~\ref{eq:diversity-approximation} numerically with Monte-Carlo integration for simplicity.

The surrogate model is just an approximation to the true function, and has large uncertainty where observations are lacking. Rather than only maximizing the function value when picking the next sample, BO methods usually also take into consideration the estimated uncertainty to avoid being overly greedy. For example, GP-UCB (Gaussian Process Upper Confidence Bound), one of the most popular BO algorithms, adds a variance term into its acquisition function. Similarly, we could adopt a composite acquisition function as follows: 
\begin{equation}\label{eq:diversity-ucb}
    a'(x_{t+1}) = \hat{a}(x_{t+1}) + \beta\sigma_t(x_{t+1}),
\end{equation}
where $\sigma_t(x_{t+1})$ is the heuristic term favoring candidates with large uncertainty, and $\beta$ is a hyperparameter trading off exploration and exploitation (diversity optimization in our case). Theoretically well justified choice of $\beta$ exists for GP-UCB, which guarantees optimization convergence with high probability \cite{GP-bandit}. However in our context, no such guarantees hold as we are not optimizing $f$ but rather the diversity of $f$, the tuning of the hyperparameter $\beta$ is thus not trivial, especially when the strategy evaluation function $g(\cdot)$ is extremely costly. To mitigate this problem, we decouple the two terms and alternate between exploration and exploitation following a similar idea proposed in \cite{alternative-explore-exploit}. During exploration, our acquisition function becomes:
\begin{equation}\label{eq:explore-acquisition}
    a_\text{exp}(x_{t+1}) = \sigma_t(x_{t+1}).
\end{equation}
The sample with the largest posterior standard deviation is chosen as $x_{t+1}$ to be evaluated next:
\begin{equation}\label{eq:explore-formula-2}
    x_{t+1} = \mathop{\arg\max}\limits_{x}\sigma_t(x).
\end{equation}
Under the condition that $g(\cdot)$ is a sample from GP function distribution $\mathcal{GP}(m(\cdot), k(\cdot,\cdot))$, Equation~\ref{eq:explore-formula-2} can be shown to maximize the Information Gain $I$ on function $g(\cdot)$:
\begin{equation}\label{eq:explore-formula-1}
    x_{t+1} = \mathop{\arg\max}\limits_{x}I\left(X_t \cup \{x\}, F_t \cup \{g(x)\} ; g\right),
\end{equation}
where $I(A;B)=H(A)-H(A|B)$, and $H(\cdot)=\mathbb{E}\left[-\log p(\cdot)\right]$ is the Shannon entropy \cite{information-theory}. 

We summarize our BDS algorithm in Algorithm~\ref{algorithm::BDS}. The alternation of exploration and diversity optimization involves two extra hyperparameters $N_\text{exp}$ and $N_\text{opt}$, corresponding to the number of samples allocated for exploration and diversity optimization in each round. Compared to $\beta$ in Equation~\ref{eq:diversity-ucb}, $N_\text{exp}$ and $N_\text{opt}$ are much more intuitive to tune. We also found that empirically the algorithm performance is insensitive to the specific values of $N_\text{exp}$ and $N_\text{opt}$. The exploitation stage directly maximizes the diversity of motion strategies. We optimize $\hat{a}(\cdot)$ with a sampling-based method DIRECT (Dividing Rectangle) \cite{Jones2001-DIRECT}, since derivative information may not be accurate in the presence of function noise due to the Monte-Carlo integration. This optimization does not have to be perfectly accurate, since the surrogate model is an approximation in the first place. The exploration stage facilitates the discovery of diverse strategies by avoiding repeated queries on well-sampled regions. We optimize $a_\text{exp}(\cdot)$ using a simple gradient-based method L-BFGS \cite{LBFGS}.

\begin{algorithm}[t]
\caption{Bayesian Diversity Search}\label{algorithm::BDS}
\KwIn{Strategy evaluation function $g(\cdot)$, exploration count $N_\text{exp}$ and diversity optimization count $N_\text{opt}$, total sample count $n$.}
\KwOut{Initial states $X_n = \{x_1, x_2, ..., x_n\}$ for diverse strategies.}
$t=0$; $X_0 \leftarrow \varnothing$; $F_0 \leftarrow \varnothing$\;
Initialize GP surrogate model with random samples\;
\While{$t < n$}
{
\eIf{$t \% (N_\text{exp} + N_\text{opt}) < N_\text{exp}$}
{
$x_{t+1} \leftarrow \mathop{\arg\max}a_\text{exp}(\cdot)$ by L-BFGS; 
\tcp*[f]{Equation~\ref{eq:explore-acquisition}}
}
{
$x_{t+1} \leftarrow \mathop{\arg\max}\hat{a}(\cdot)$ by DIRECT; \tcp*[f]{Equation~\ref{eq:diversity-approximation}}
}
$f_{t+1} \leftarrow g(x_{t+1})$\; 
$X_{t+1} \leftarrow X_t \cup \{x_{t+1}\}$; $F_{t+1} \leftarrow F_t \cup \{f_{t+1}\}$\;
Update GP surrogate model with $X_{t+1}$, $F_{t+1}$;
\hspace{8pt}\tcp{Equation~\ref{eq:gp-mu-sigma}}
$t \leftarrow t+1$;
}
\Return $X_n$
\end{algorithm}

\subsection{Stage 2: Novel Policy Seeking}
\label{sec:methods-phase2}

In Stage 2 of our diverse strategy discovery framework, we explore potential strategy variations given a fixed initial state discovered in Stage 1. Formally, given an initial state $x$ and a set of discovered policies $\Pi = \{\pi_1, \pi_2, ..., \pi_n\}$, we aim to learn a new policy $\pi_{n+1}$ which is different from all existing $\pi_i \in \Pi$. This can be achieved with an additional policy novelty reward to be jointly optimized with the task reward during DRL training. We measure the novelty of policy $\pi_i$ with respect to $\pi_j$ by their corresponding motion feature distance $||f_i - f_j||$. The novelty reward function is then given by
\begin{equation}
    r_\text{novelty}(f) = \text{Clip}\left(\frac{\min_{\pi_i \in \Pi}{||f_i - f||}}{d_\text{threshold}}, 0.01, 1\right) ,
\end{equation}
which rewards simulation rollouts showing different strategies to the ones presented in the existing policy set. $d_\text{threshold}$ is a hyperparameter measuring the desired policy novelty to be learned next. Note that the feature representation $f$ here in Stage 2 can be the same as or different from the one used in Stage 1 for initial states exploration.

Our novel policy search is in principle similar to the idea of \cite{zhang2019novel-policies,Sun2020novel-policies}. However, there are two key differences. First, in machine learning, policy novelty metrics have been designed and validated only on low-dimensional control tasks. For example in \cite{zhang2019novel-policies}, the policy novelty is measured by the reconstruction error between states from the current rollout and previous rollouts encapsulated as a deep autoencoder. In our case of high-dimensional 3D character control tasks, however, novel state sequences do not necessarily correspond to novel motion strategies. We therefore opt to design discriminative strategy features whose distances are incorporated into the DRL training reward.

Second, we multiply the novelty reward with the task reward as the training reward, and adopt a standard gradient-based method PPO to train the policy. Additional optimization techniques are not required for learning novel strategies, such as the Task-Novelty Bisector method proposed in \cite{zhang2019novel-policies} that modifies the policy gradients to encourage novelty learning. Our novel policy seeking procedure always discovers novel policies since the character is forced to perform a different strategy. However, the novel policies may exhibit unnatural and awkward movements, when the given initial state is not capable of multiple natural strategies.
\section{Task Setup and Implementation}
\label{sec:Experiments}

We demonstrate diverse strategy discovery for two challenging motor tasks: high jumping and obstacle jumping. We also tackle several variations of these tasks. We describe task specific settings in Section~\ref{sec:Experiments-Task-Setup}, and implementation details in Section~\ref{sec:Experiments-Implementation}. 

\subsection{Task Setup}
\label{sec:Experiments-Task-Setup}
The high jump task follows the Olympics rules, where the simulated athlete takes off with one leg, clears the crossbar, and lands on a crash mat. We model the landing area as a rigid block for simplicity. The crossbar is modeled as a rigid wall vertically extending from the ground to the target height to prevent the character from cheating during early training, i.e., passing through beneath the bar. A rollout is considered successful and terminated when the character lands on the rigid box with all body parts at least $20$ $cm$ away from the wall. A rollout is considered as a failure and terminated immediately, if any body part touches the wall, or any body part other than the take-off foot touches the ground, or if the jump does not succeed within two seconds after the take-off.

The obstacle jump shares most of the settings of the high jump. The character takes off with one leg, clears a box-shaped obstacle of $50$ $cm$ in height with variable widths, then lands on a crash mat. The character is required to complete the task within two seconds as well, and not allowed to touch the obstacle with any body part.

\subsubsection{Run-up Learning}
\label{sec:Experiments-Runup}
A full high jump or obstacle jump consists of three phases: run-up, take-off and landing. Our framework described so far can discover good initial states at take-off that lead to diverse jumping strategies. What is lacking is the matching run-up control policies that can prepare the character to reach these good take-off states at the end of the run. We train the run-up controllers with DeepMimic \cite{Peng:2018:DeepMimic}, where the DRL learning reward consists of a task reward and an imitation reward. The task reward encourages the end state of the run-up to match the desired take-off state of the jump. The imitation reward guides the simulation to match the style of the reference run. We use a curved sprint as the reference run-up for high jump, and a straight sprint for the obstacle jump run-up. For high jump, the explored initial state space is four-dimensional: the desired approach angle $\alpha$, the $X$ and $Z$ components of the root angular velocity $\omega$, and the magnitude of the $Z$ component of the root linear velocity $v_z$ in a facing-direction invariant frame. We fix the desired root $Y$ angular velocity to $3 \text{rad}/\text{s}$, which is taken from the reference curved sprint. In summary, the task reward $r_\text{G}$ for the run-up control of a high jump is defined as
\begin{equation}
    r_\text{G} = \text{exp}\left(-\frac{1}{3}\cdot||\omega - \bar{\omega}||_1 - 0.7\cdot(v_z - \bar{v}_z)^2\right),
\end{equation}
where $\bar{\omega}$ and $\bar{v}_z$ are the corresponding targets for $\omega$ and $v_z$. $\alpha$ does not appear in the reward function as we simply rotate the high jump suite in the environment to realize different approach angles. For the obstacle jump, we explore a three-dimensional take-off state space consisting of the root angular velocities along all axes. Therefore the run-up control task reward $r_\text{G}$ is given by
\begin{equation}
    r_\text{G} = \text{exp}(-\frac{1}{3}\cdot||\omega - \bar{\omega}||_1).
\end{equation}

\begin{table}[t]
\centering
\caption{Curriculum parameters for learning jumping tasks. $z$ parameterizes the task difficulty, i.e., the crossbar height in high jumps and the obstacle width in obstacle jumps. $z_\text{min}$ and $z_\text{max}$ specify the range of $z$, and $\Delta z$ is the increment when moving to a higher difficulty level. $R_T$ is the accumulated reward threshold to move on to the next curriculum difficulty.}
\begin{tabular}{|c|c|c|c|c|c|} 
\hline
Task & $z_\text{min}$(cm) & $z_\text{max}$(cm) & $\Delta z$(cm) & $R_T$ \\ 
\hline
High jump     & 50   & 200 & 1 & 30     \\
\hline
Obstacle jump & 5  & 250 & 5 & 50   \\
\hline
\end{tabular}
\label{tb:curriculum}
\end{table}

\subsubsection{Reward Function}
\label{sec:Experiments-Reward}
We use the same reward function structure for both high jumps and obstacle jumps, where the character gets a sparse reward only when it successfully completes the task. The full reward function is defined as in Equation~\ref{eq:stage1-reward} for Stage 1. For Stage 2, the novelty bonus $r_\text{novelty}$ as discussed in Section~\ref{sec:methods-phase2} is added:
\begin{equation}
    r = r_\text{task}\cdot r_\text{naturalness}\cdot r_\text{novelty}.
    \label{eq:stage2-reward}
\end{equation}
$r_\text{naturalness}$ is the motion naturalness term discussed in Section~\ref{sec:methods-PVAE}. For both stages, the task reward consists of three terms:
\begin{equation}
    r_\text{task} = r_{\text{complete}} \cdot r_{\omega} \cdot r_\text{safety}.
\end{equation}
$r_\text{complete}$ is a binary reward precisely corresponding to task completion. $r_{\omega} = \text{exp}(-0.02 ||\omega||)$ penalizes excessive root rotations where $||\omega||$ is the average magnitude of the root angular velocities across the episode. $r_\text{safety}$ is a term to penalize unsafe head-first landings. We set it to $0.7$ for unsafe landings and $1.0$ otherwise. $r_\text{safety}$ can also be further engineered to generate more landing styles, such as a landing on feet as shown in Figure~\ref{fig:variations}.

\subsubsection{Curriculum and Scheduling}
\label{sec:Experiments-Curriculum}
The high jump is a challenging motor skill that requires years of training even for professional athletes. We therefore adopt curriculum-based learning to gradually increase the task difficulty $z$, defined as the crossbar height in high jumps or the obstacle width in obstacle jumps. Detailed curriculum settings are given in Table~\ref{tb:curriculum}, where $z_\text{min}$ and $z_\text{max}$ specify the range of $z$, and $\Delta z$ is the increment when moving to a higher difficulty level.

We adaptively schedule the curriculum to increase the task difficulty according to the DRL training performance. At each training iteration, the average sample reward is added to a reward accumulator. We increase $z$ by $\Delta z$ whenever the accumulated reward exceeds a threshold $R_T$, and then reset the reward accumulator. Detailed settings for $\Delta z$ and $R_T$ are listed in Table~\ref{tb:curriculum}. The curriculum could also be scheduled following a simpler scheme adopted in \cite{Xie2020allsteps}, where task difficulty is increased when the average sample reward in each iteration exceeds a threshold. We found that for athletic motions, such average sample reward threshold is hard to define uniformly for different strategies in different training stages.

Throughout training, the control frequency $f_\text{control}$ and the P-VAE offset penalty coefficient $c_\text{offset}$ in Equation \ref{eq:r-pvae} are also scheduled according to the task difficulty, in order to encourage exploration and accelerate training in early stages. We set $f_\text{control} = 10 + 20 \cdot \text{Clip}(\rho, 0, 1)$ and $c_\text{offset} = 48 - 33 \cdot \text{Clip}(\rho, 0, 1)$, where $\rho = 2z-1$ for high jumps and $\rho = z$ for obstacle jumps. We find that in practice the training performance does not depend sensitively on these hyperparameters.

\begin{table}[t]
\centering
\caption{Model parameters of our virtual athlete and the mocap athlete.}
\begin{tabular}{|c|c|c|} 
\hline
Parameter & Simulated Athlete & Mocap Athlete\\ 
\hline
Weight (kg) & 60  & 70     \\
\hline
Height (cm) & 170  & 191     \\
\hline
hip height (cm) & 95  & 107     \\
\hline
knee height (cm) & 46  & 54     \\
\hline
\end{tabular}
\label{tb:modelParams}
\end{table}

\subsubsection{Strategy Features}
\label{sec:Experiments-Strategy-Features}
We choose low-dimensional and visually discriminate features $f$ of learned strategies for effective diversity measurement of discovered strategies. In the sports literature, high jump techniques are usually characterized by the body orientation when the athlete clears the bar at his peak position. The rest of the body limbs are then coordinated in the optimal way to clear the bar as high as possible. Therefore we use the root orientation when the character's CoM lies in the vertical crossbar plane as $f$. This three-dimensional root orientation serves well as a Stage 2 feature for high jumps. For Stage 1, this feature can be further reduced to one dimension, as we will show in Section \ref{sec:Experiments-Diverse-Strategies}. More specifically, we only measure the angle between the character's root direction and the global up vector, which corresponds to whether the character clears the bar facing upward or downward. Such a feature does not require a non-Euclidean GP output space that we need to handle in Stage 1. We use the same set of features for obstacle jumps, except that root orientations are measured when the character's CoM lies in the center vertical plane of the obstacle.

Note that it is not necessary to train to completion, i.e., the maximum task difficulty, to evaluate the feature diversity, since the overall jumping strategy usually remains unchanged after a given level of difficulty, which we denote by $z_\text{freeze}$. Based on empirical observations, we terminate the training after reaching $z_\text{freeze}=100cm$ for both high jump and obstacle jump tasks for strategy discovery. 

\subsubsection{GP Priors and Kernels}
\label{sec:Experiments-GP-Priors-Kernels}
We set GP prior $m(\cdot)$ and kernel $k(\cdot,\cdot)$ for BDS based on common practices in the Bayesian optimization literature. Without any knowledge on the strategy feature distribution, we set the prior mean $m(\cdot)$ to be the mean of the value range of a feature. Among the many common choices for kernel functions, we adopt the Mat\'ern$^5/_2$ kernel \cite{matern1960spatial,klein2017fast}, defined as:
\begin{equation}
    k_{^5/_2}(x, x') = \theta(1 + \sqrt{5}d_{\lambda}(x,x') + \frac{5}{3}d^{2}_{\lambda}(x, x'))e^{-\sqrt{5}d_{\lambda}(x, x')} 
\end{equation}
where $\theta$ and $\lambda$ are learnable parameters of the GP. 
$d_{\lambda}(x,x') = (x-x')^{T}\diag(\lambda)(x-x')$ is the Mahalanobis distance.

\subsection{Implementation}
\label{sec:Experiments-Implementation}

We implemented our system in PyTorch \cite{PyTorch} and PyBullet \cite{coumans2016pybullet}. The simulated athlete has 28 internal DoFs and 34 DoFs in total. We run the simulation at 600~$Hz$. {Torque limits for the hips, knees and ankles are taken from Biomechanics estimations for a human athlete performing a Fosbury flop \cite{Okuyama03}. Torque limits for other joints are kept the same as \cite{Peng:2018:DeepMimic}. Joint angle limits are implemented by penalty forces.} We captured three standard high jumps from a university athlete, whose body measurements are given in Table~\ref{tb:modelParams}. For comparison, we also list these measurements for our virtual athlete. 

For DRL training, we set $\lambda=0.95$ for both TD($\lambda$) and GAE($\lambda$). We set the discounter factor $\gamma=1.0$ since our tasks have short horizon and sparse rewards. The PPO clip threshold is set to $0.02$. The learning rate is $2.5\times 10^{-5}$ for the policy network and $1.0\times 10^{-2}$ for the value network. In each training iteration, we sample 4096 state-action tuples in parallel and perform five policy updates with a mini-batch size of 256. For Stage 1 diverse strategy discovery, we implement BDS using GPFlow \cite{GPflow2020multioutput-gp} with both $N_\text{exp}$ and $N_\text{opt}$ set to three. $d_\text{threshold}$ in Stage 2 novel policy seeking is set to $\pi/2$. Our experiments are performed on a Dell Precision 7920 Tower workstation, with dual Intel Xeon Gold 6248R CPUs (3.0 GHz, 48 cores) and an Nvidia Quadro RTX 6000 GPU. Simulations are run on the CPUs. One strategy evaluation for a single initial state, i.e. Line 9 in Algorithm~\ref{algorithm::BDS}, typically takes about six hours. Network updates are performed on the GPU.
\begin{figure}
    \centering
    \begin{subfigure}[b]{0.99\linewidth}
        \includegraphics[width=0.99\linewidth]{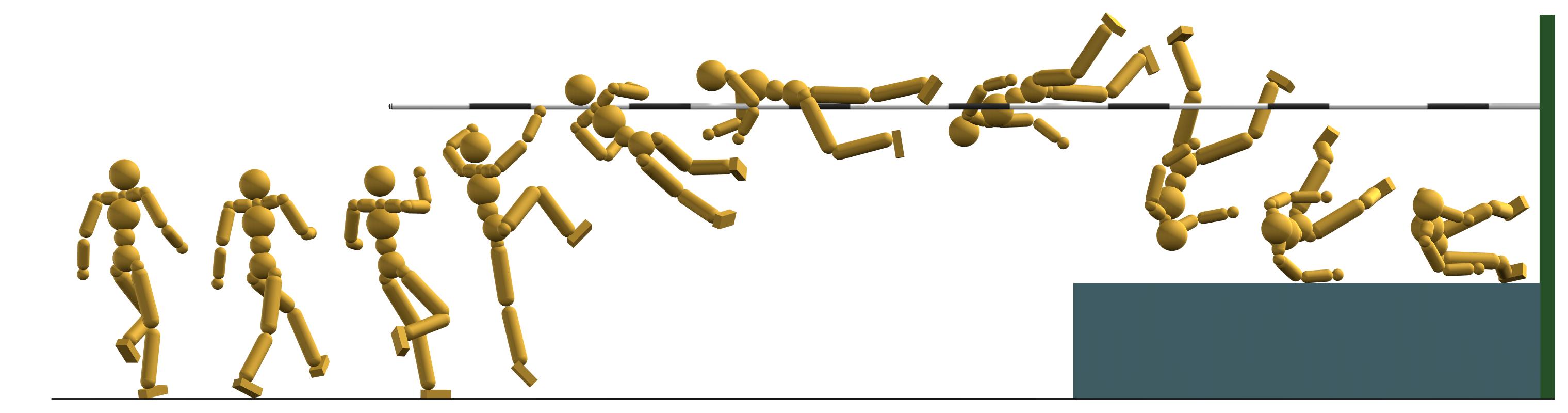}
        \caption{Straddle -- max height=$190cm$}
        \label{fig:highjump-straddle}
    \end{subfigure}
    \begin{subfigure}[b]{0.99\linewidth}
        \includegraphics[width=0.99\linewidth]{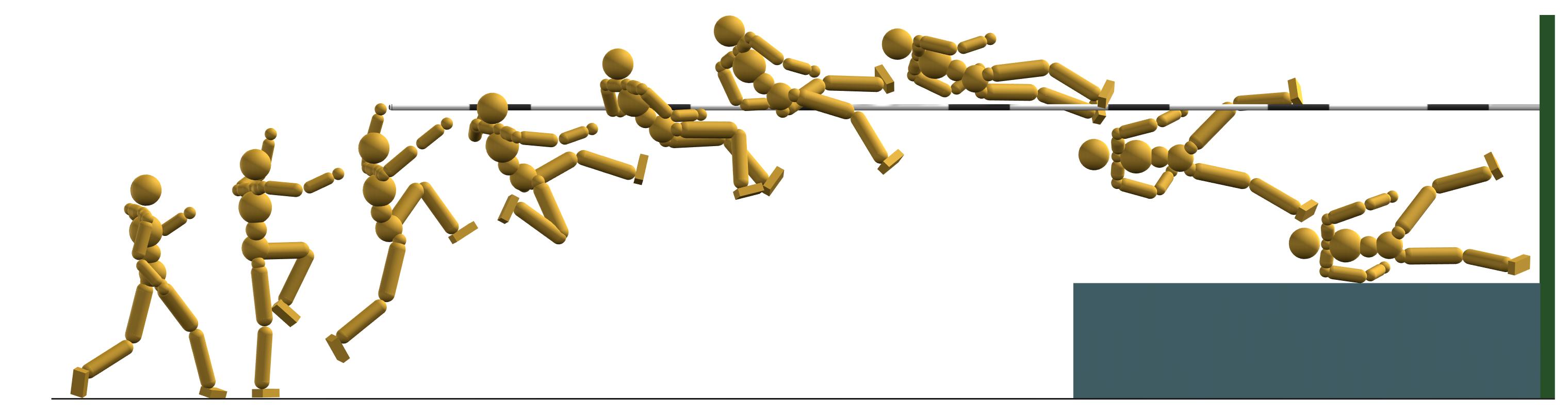}
         \caption{Front Kick -- max height=$180cm$}
         \label{fig:highjump-frontkick}
    \end{subfigure}
    \begin{subfigure}[b]{0.99\linewidth}
        \includegraphics[width=0.99\linewidth]{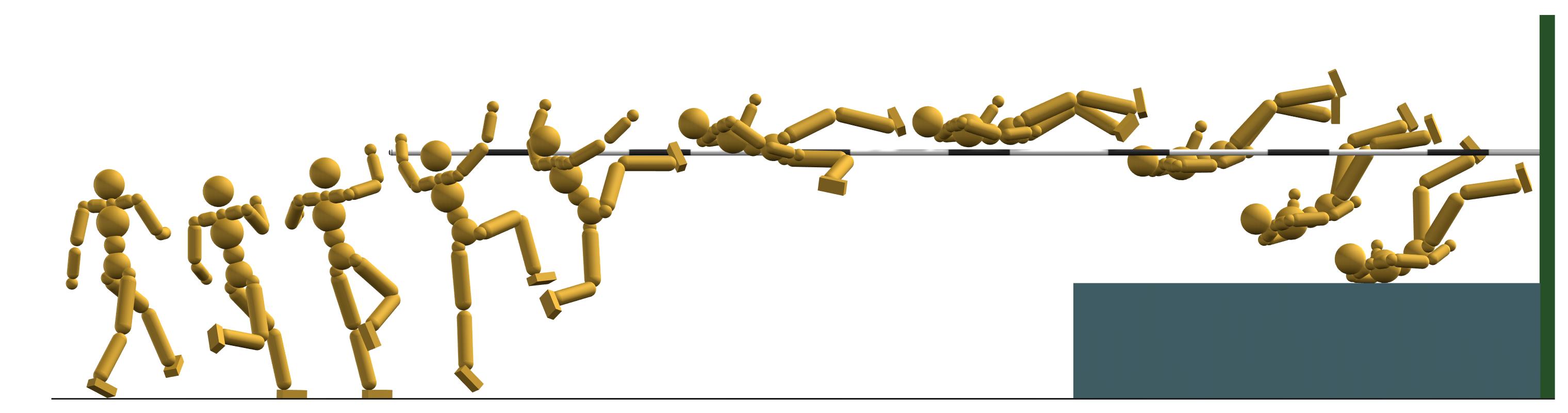}
         \caption{Western Roll (facing up) -- max height=$160cm$}
         \label{fig:highjump-rollup}
    \end{subfigure}
    \begin{subfigure}[b]{0.99\linewidth}
        \includegraphics[width=0.99\linewidth]{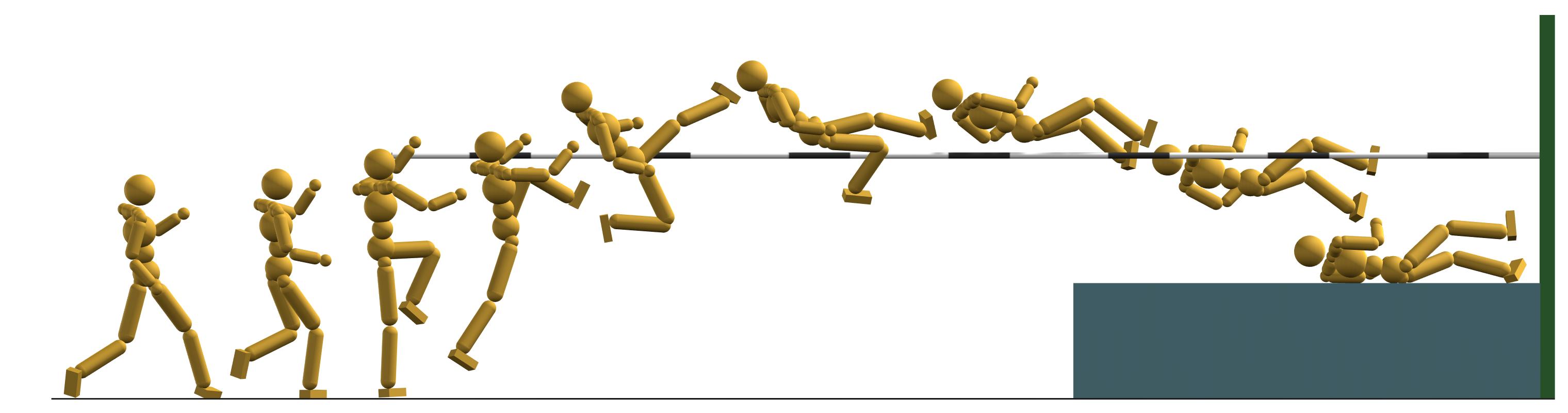}
         \caption{Scissor Kick -- max height=$150cm$}
         \label{fig:highjump-scissor}
    \end{subfigure}
    \begin{subfigure}[b]{0.99\linewidth}
        \includegraphics[width=0.99\linewidth]{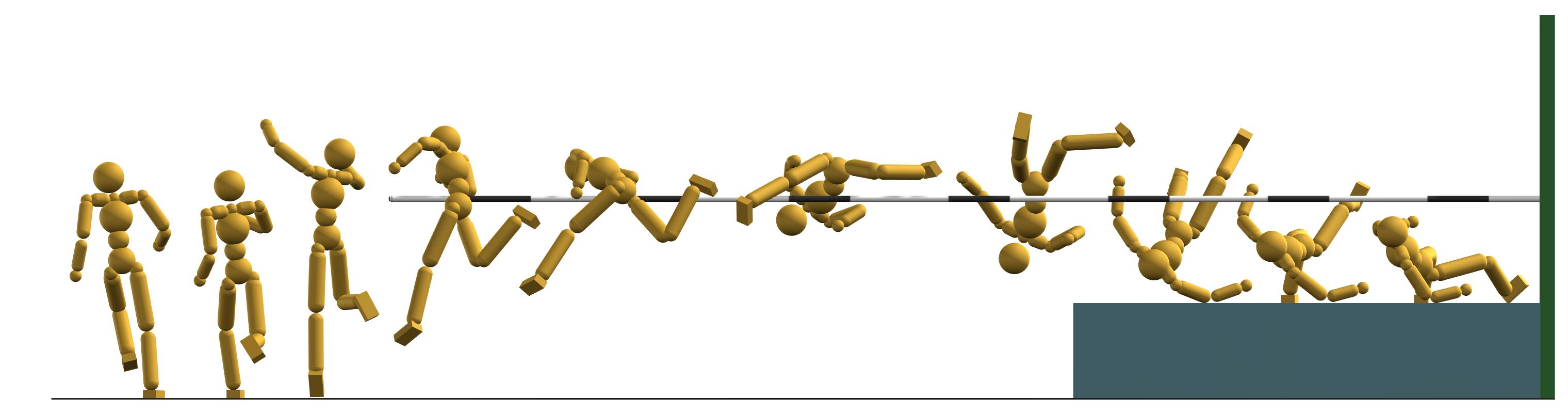}
         \caption{Side Dive -- max height=$130cm$}
         \label{fig:highjump-sidedive}
    \end{subfigure}
    \begin{subfigure}[b]{0.99\linewidth}
        \includegraphics[width=0.99\linewidth]{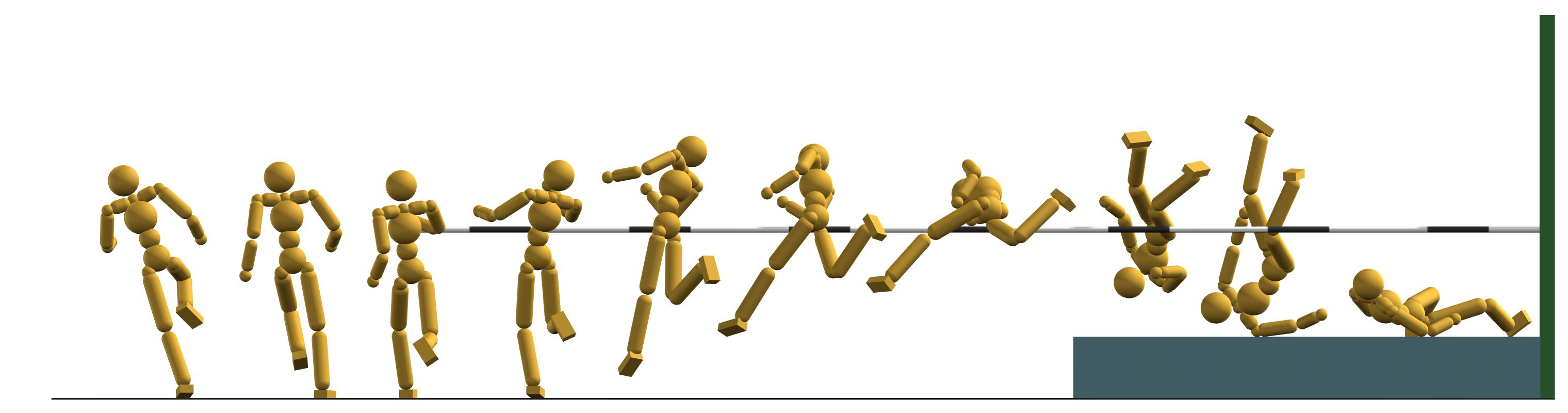}
         \caption{Side Jump -- max height=$110cm$}
         \label{fig:highjump-sidejump}
    \end{subfigure}
    \caption{Six of the eight high jump strategies discovered by our learning framework, ordered by their maximum cleared height. Fosbury Flop and Western Roll (facing sideways) are shown in Figure~\ref{fig:teaser}. Note that Western Roll (facing up) and Scissor Kick differ in the choice of inner or outer leg as the take-off leg. {The Western Roll (facing sideways) and the Scissor Kick are learned in Stage 2. All other strategies are discovered in Stage 1.}}
    \label{fig:highJumps}
\end{figure}

\begin{figure}
    \centering
    \begin{subfigure}[b]{0.99\linewidth}
        \includegraphics[width=0.99\linewidth]{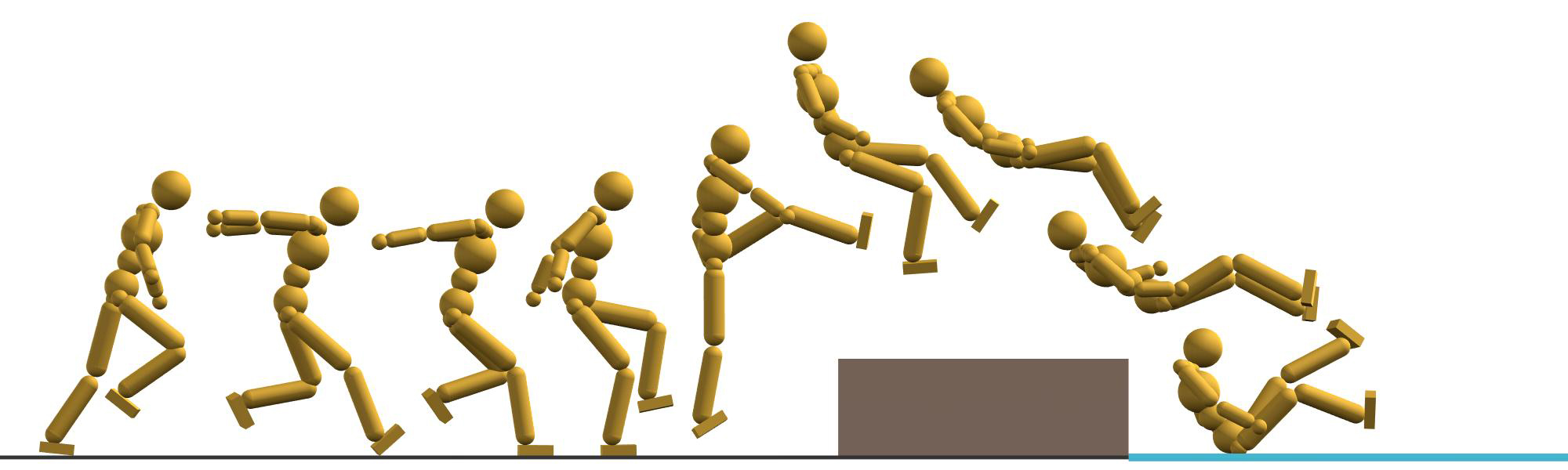}
        \caption{Front Kick -- max width=$150cm$}
        \label{fig:obstacle-frontKick}
    \end{subfigure}
    \begin{subfigure}[b]{0.99\linewidth}
        \includegraphics[width=0.99\linewidth]{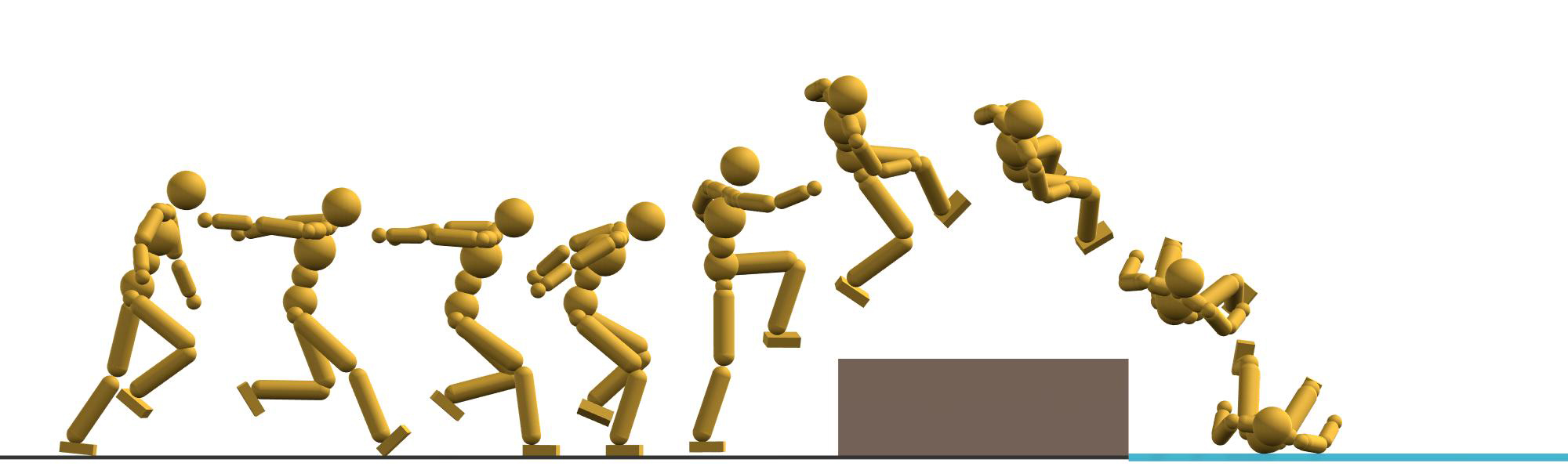}
        \caption{Side Kick -- max width=$150cm$ }
        \label{fig:obstacle-sideKick}
    \end{subfigure}
    \begin{subfigure}[b]{0.99\linewidth}
        \includegraphics[width=0.99\linewidth]{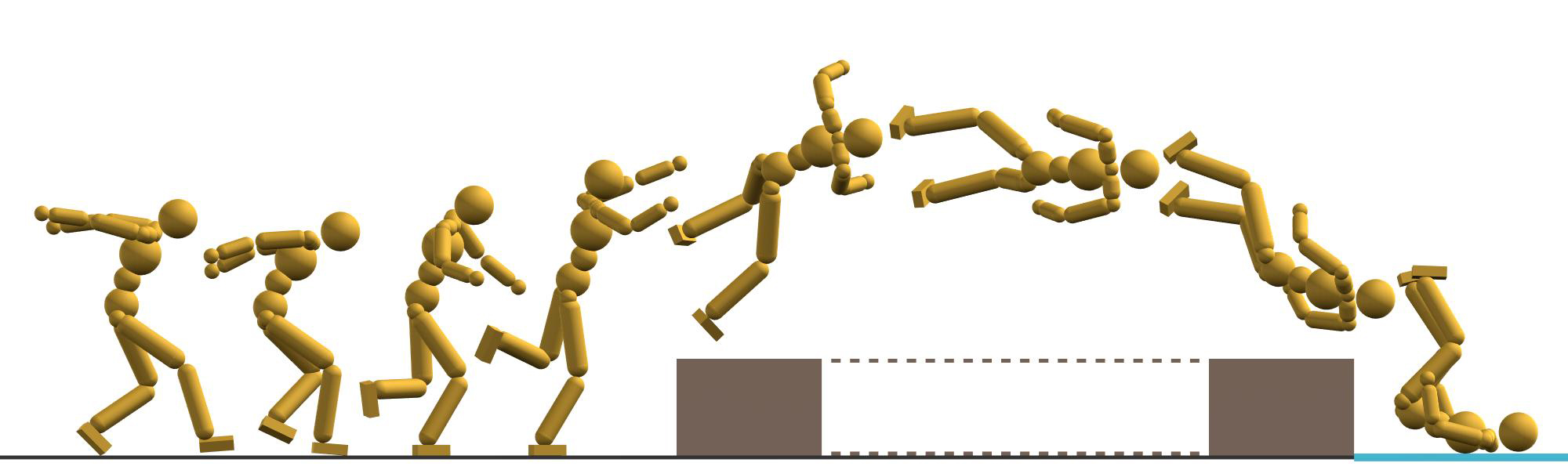}
        \caption{Twist Jump (clockwise) -- max width=$150cm$}
        \label{fig:obstacle-twistJumpC}
    \end{subfigure}
    \begin{subfigure}[b]{0.99\linewidth}
        \includegraphics[width=0.99\linewidth]{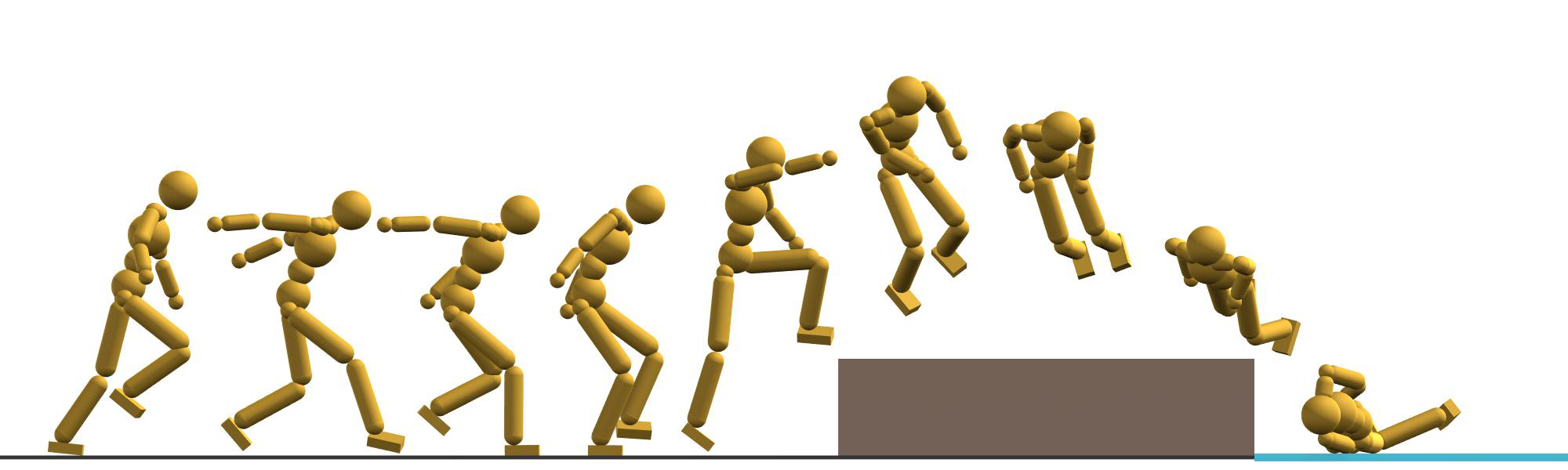}
        \caption{ Straddle -- max width=$215cm$}
        \label{fig:obstacle-straddle}
    \end{subfigure}
    \begin{subfigure}[b]{0.99\linewidth}
        \includegraphics[width=0.99\linewidth]{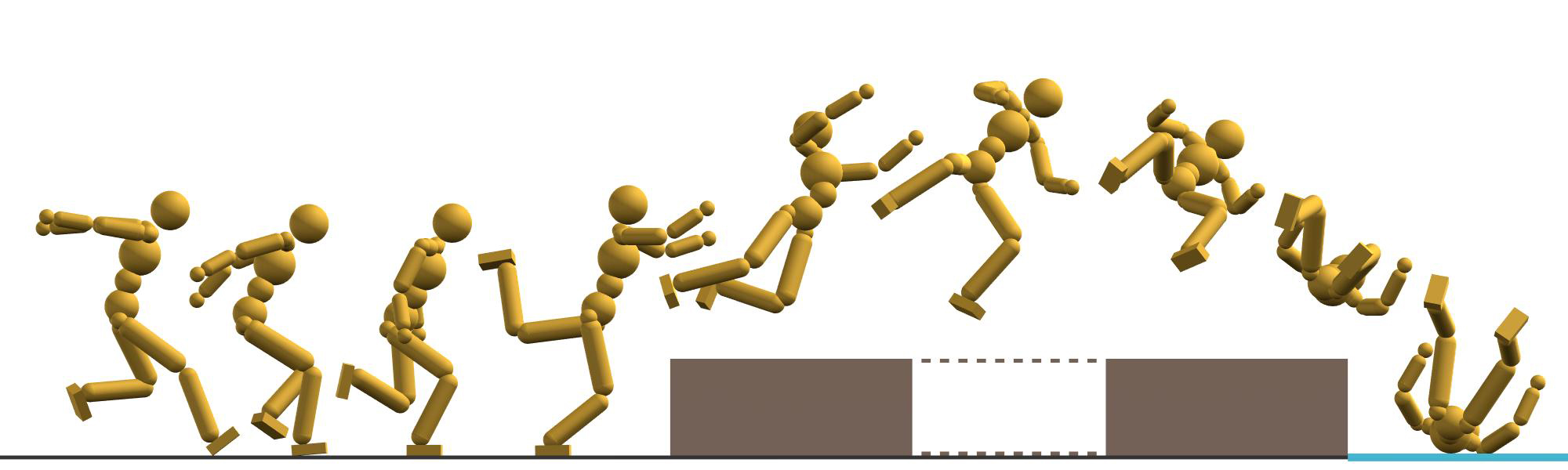}
        \caption{Twist Jump (counterclockwise) -- max width=$250cm$}
        \label{fig:obstacle-twistJumpCC}
    \end{subfigure}
    \begin{subfigure}[b]{0.99\linewidth}
        \includegraphics[width=0.99\linewidth]{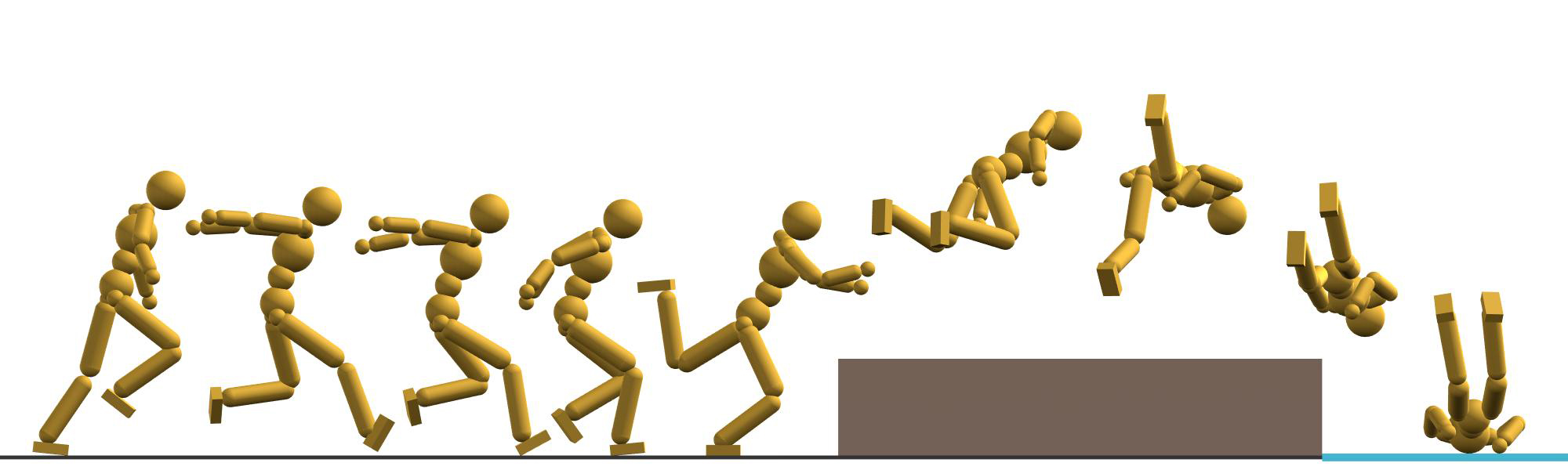}
        \caption{Dive Turn -- max width=$250cm$ }
        \label{fig:obstacle-diveTurn}
    \end{subfigure}
    \caption{Six obstacle jump strategies discovered by our learning framework in Stage 1, ordered by their maximum cleared obstacle width. For some of the strategies, the obstacle is split into two parts connected with dashed lines to enable better visualization of the poses over the obstacle.}
    \label{fig:obstacleJumps1}
\end{figure}

\begin{figure}
    \centering
    \begin{subfigure}[b]{0.99\linewidth}
        \includegraphics[width=0.99\linewidth]{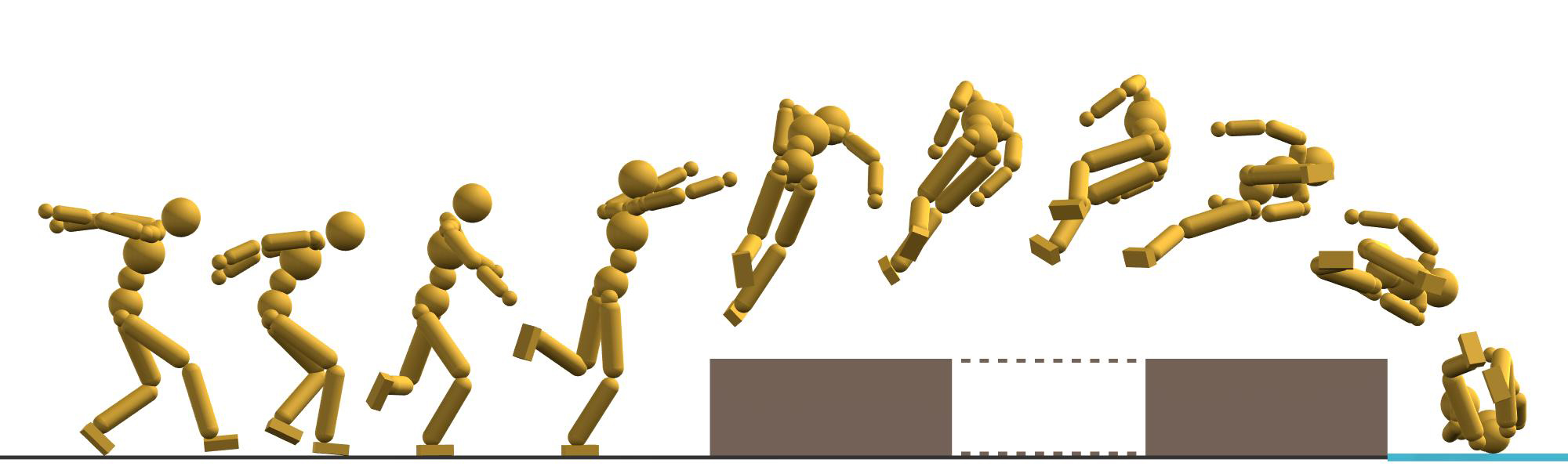}
        \caption{Twist Turn -- max width=$250cm$}
        \label{fig:obstacle-twistTurn}
    \end{subfigure}
    \begin{subfigure}[b]{0.99\linewidth}
        \includegraphics[width=0.99\linewidth]{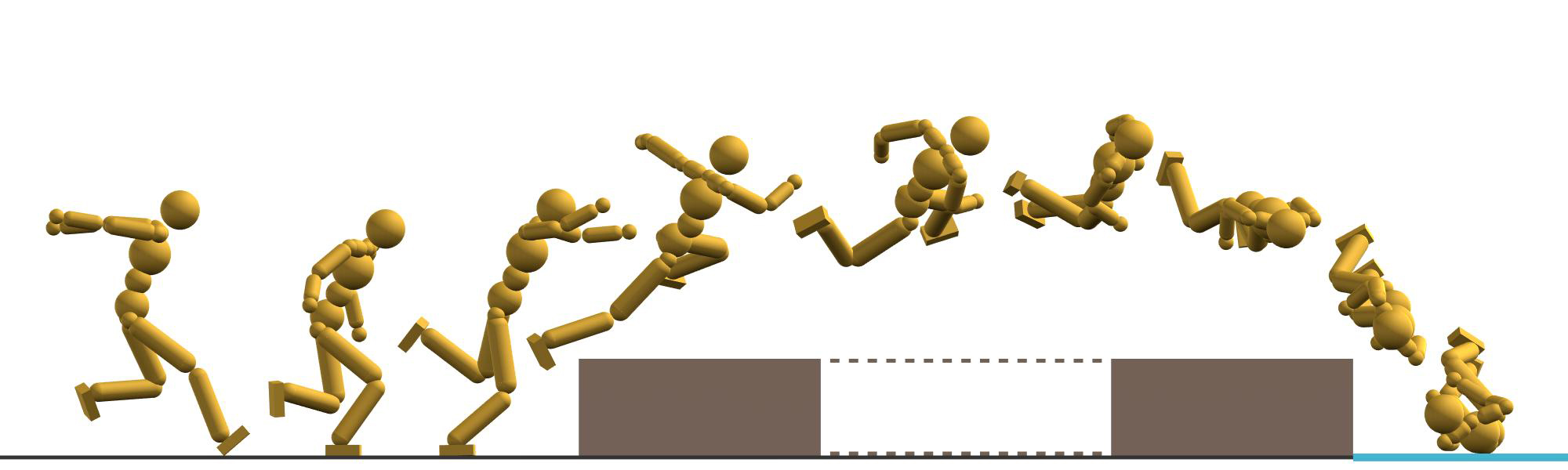}
        \caption{Western Roll -- max width=$250cm$}
        \label{fig:obstacle-roll}
    \end{subfigure}
    \caption{Two obstacle jump strategies discovered in Stage 2 of our learning framework.}
    \label{fig:obstacleJumps2}
\end{figure}
\begin{figure}
    \centering
    \includegraphics[width=1.0\linewidth]{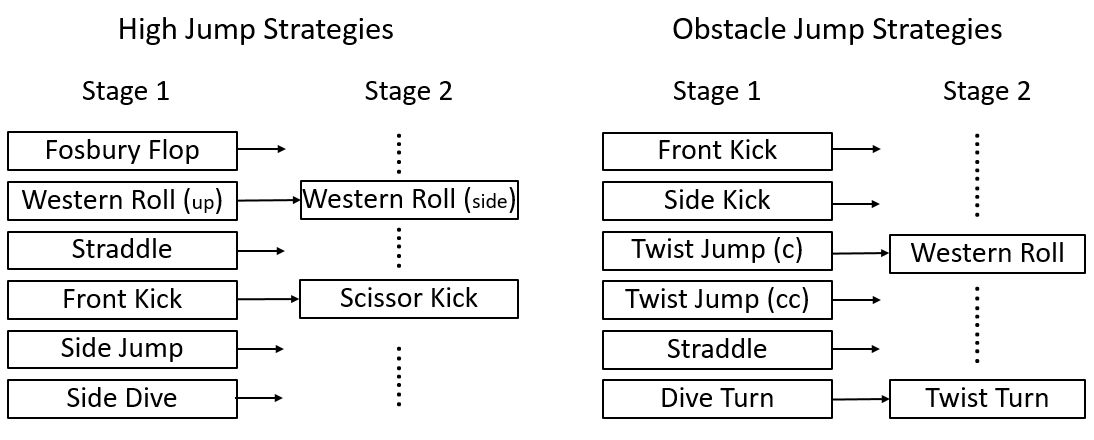}
    \caption{Diverse strategies discovered in each stage of our framework.}
\label{fig:strategies}
\end{figure}

\begin{figure*}
    \centering
    \begin{subfigure}[b]{0.12\textwidth}
    \includegraphics[width=\linewidth]{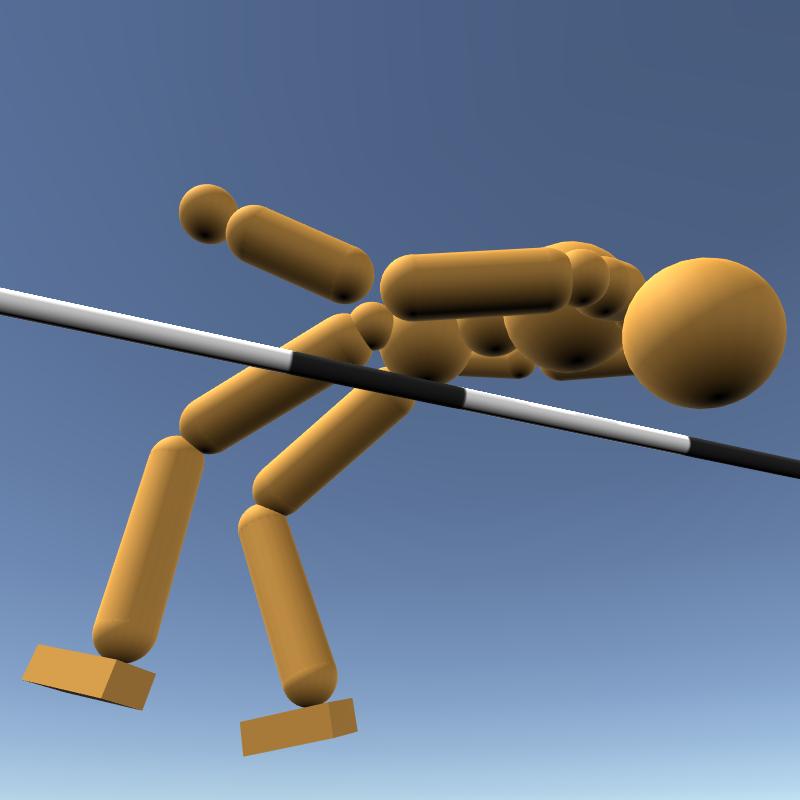}
    \includegraphics[width=\linewidth]{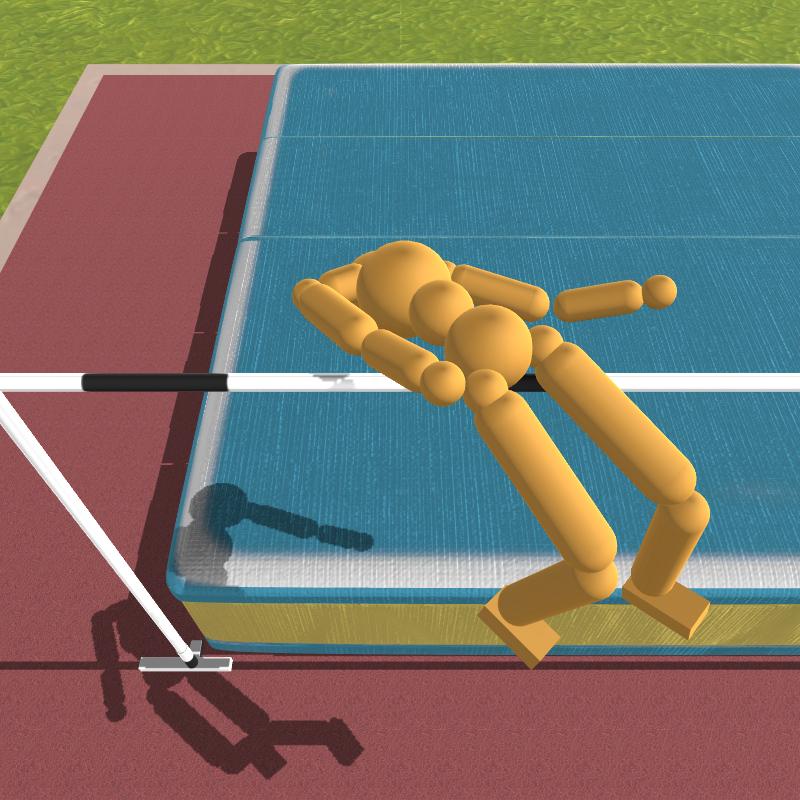}
    \caption{Fosbury Flop}
    \end{subfigure}
    \begin{subfigure}[b]{0.12\textwidth}
    \includegraphics[width=\linewidth]{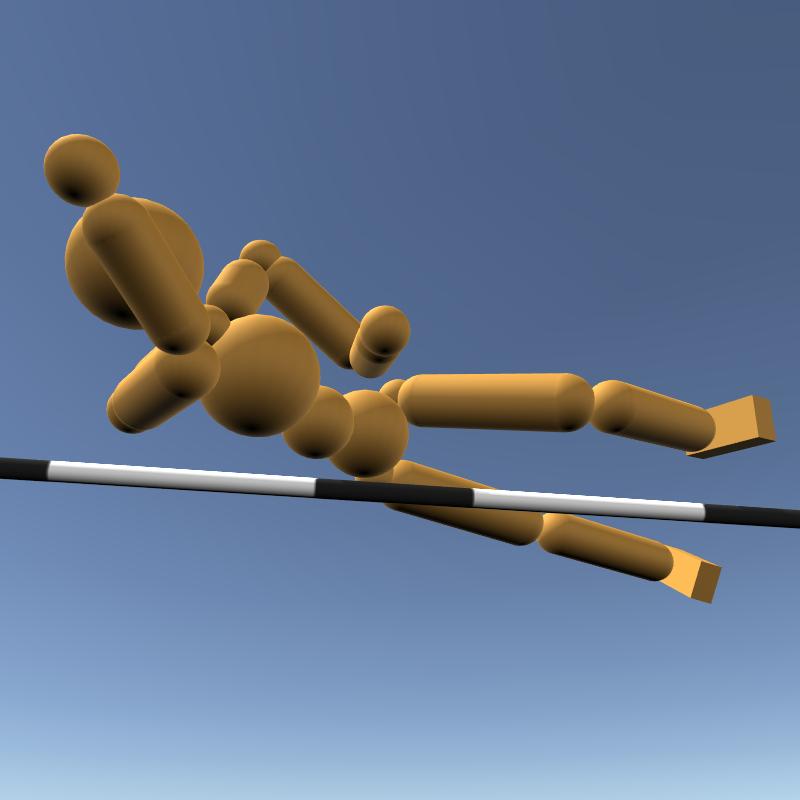}
    \includegraphics[width=\linewidth]{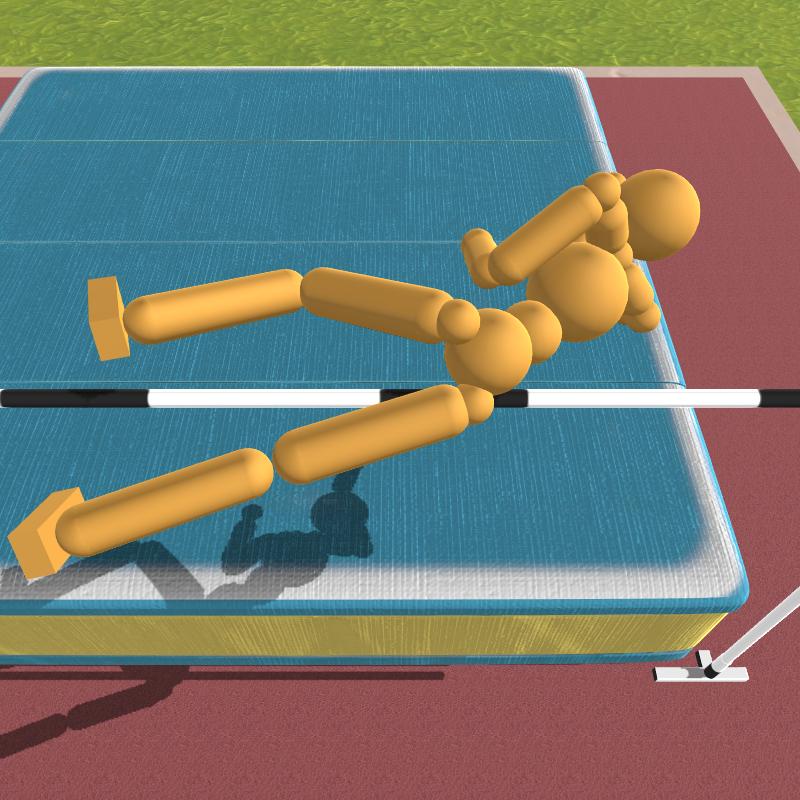}
    \caption{Western Roll}
    \end{subfigure}
    \begin{subfigure}[b]{0.12\textwidth}
    \includegraphics[width=\linewidth]{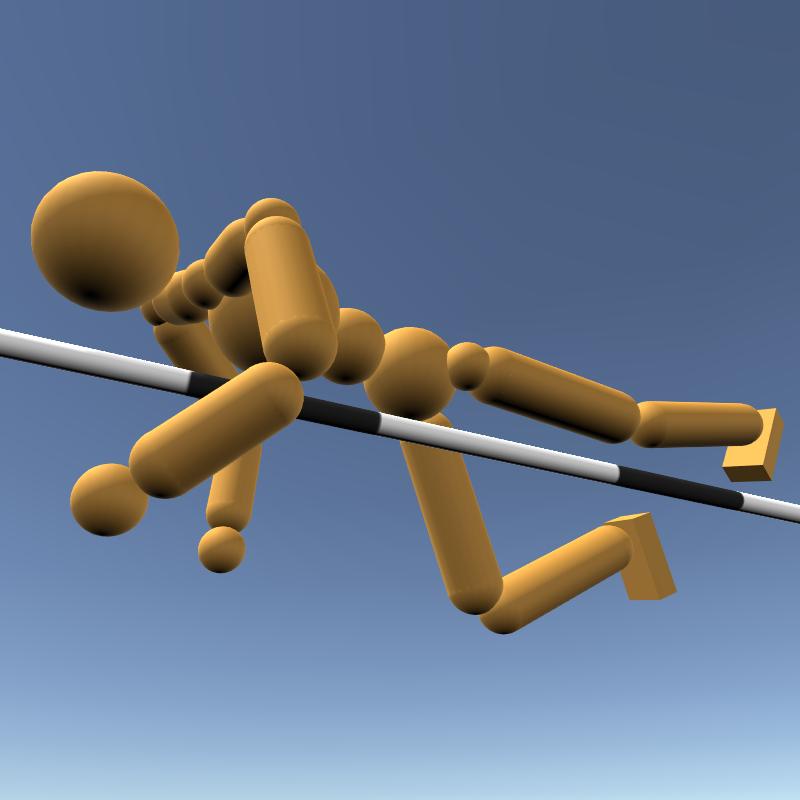}
    \includegraphics[width=\linewidth]{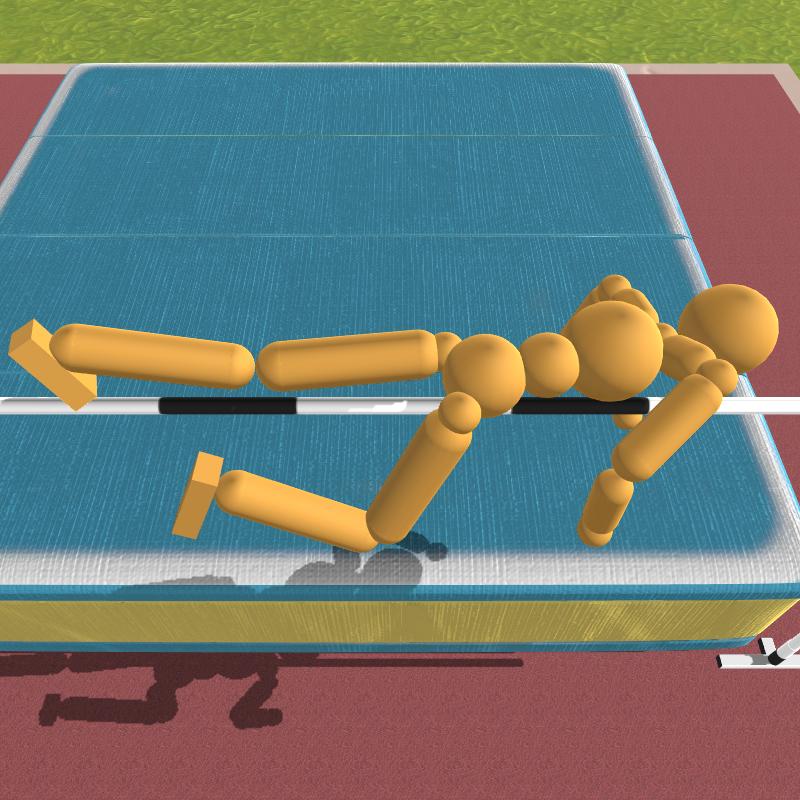}
    \caption{Straddle}
    \end{subfigure}
    \begin{subfigure}[b]{0.12\textwidth}
    \includegraphics[width=\linewidth]{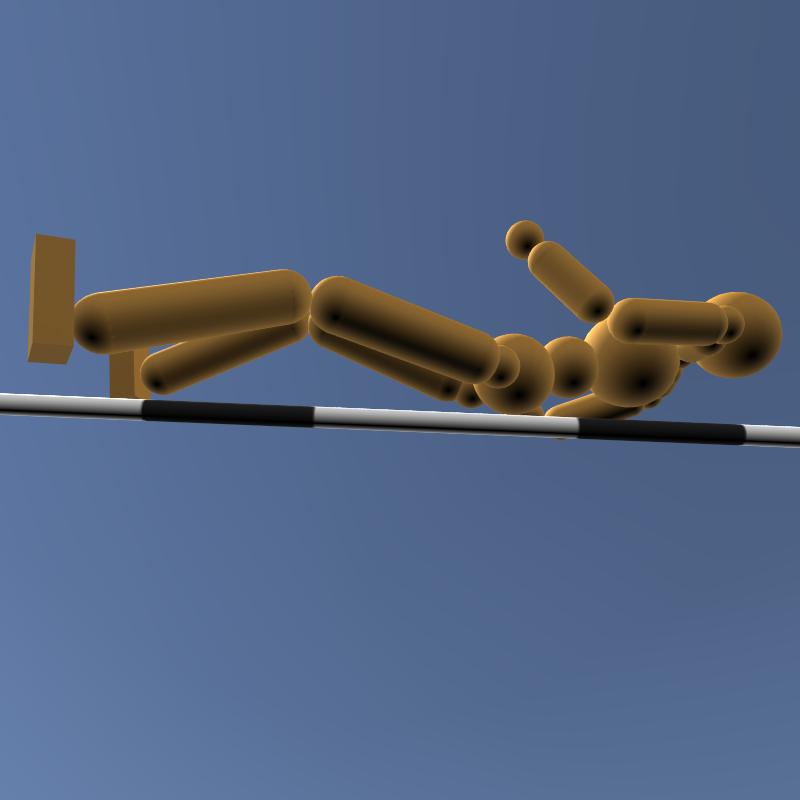}
    \includegraphics[width=\linewidth]{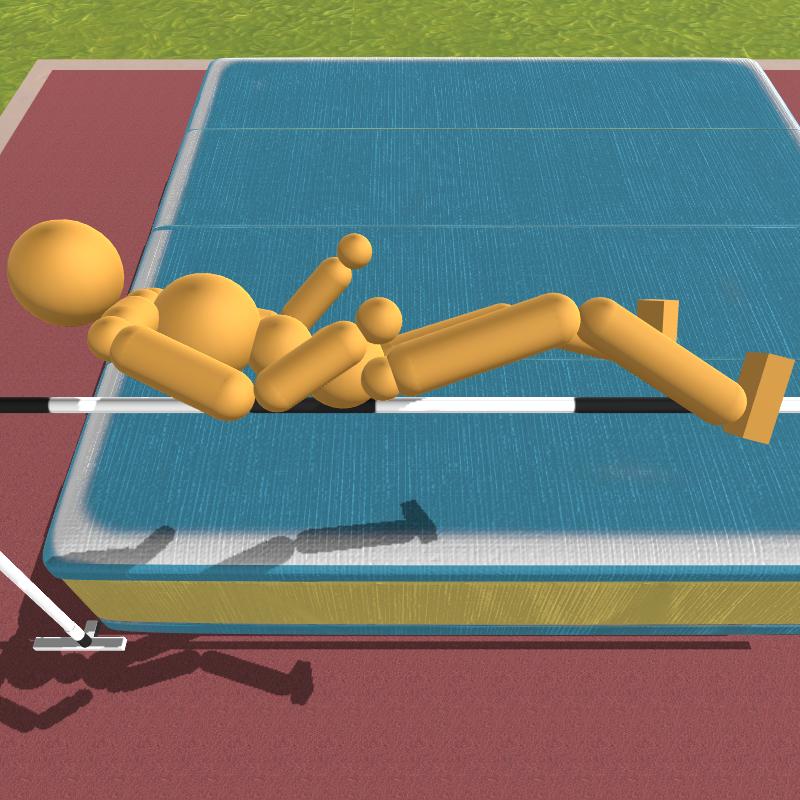}
    \caption{Front Kick}
    \end{subfigure}
    \begin{subfigure}[b]{0.12\textwidth}
    \includegraphics[width=\linewidth]{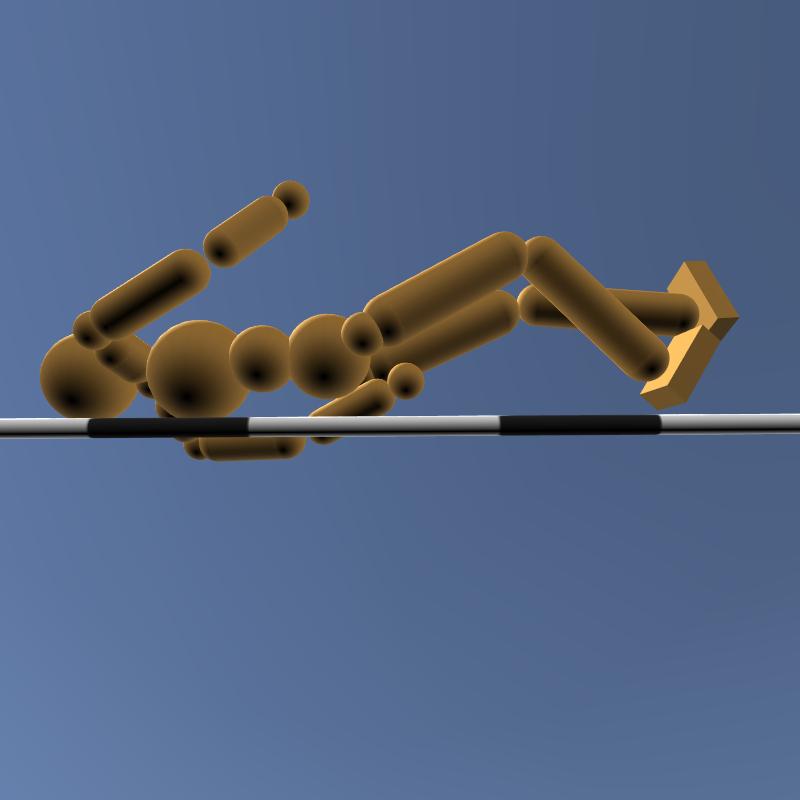}
    \includegraphics[width=\linewidth]{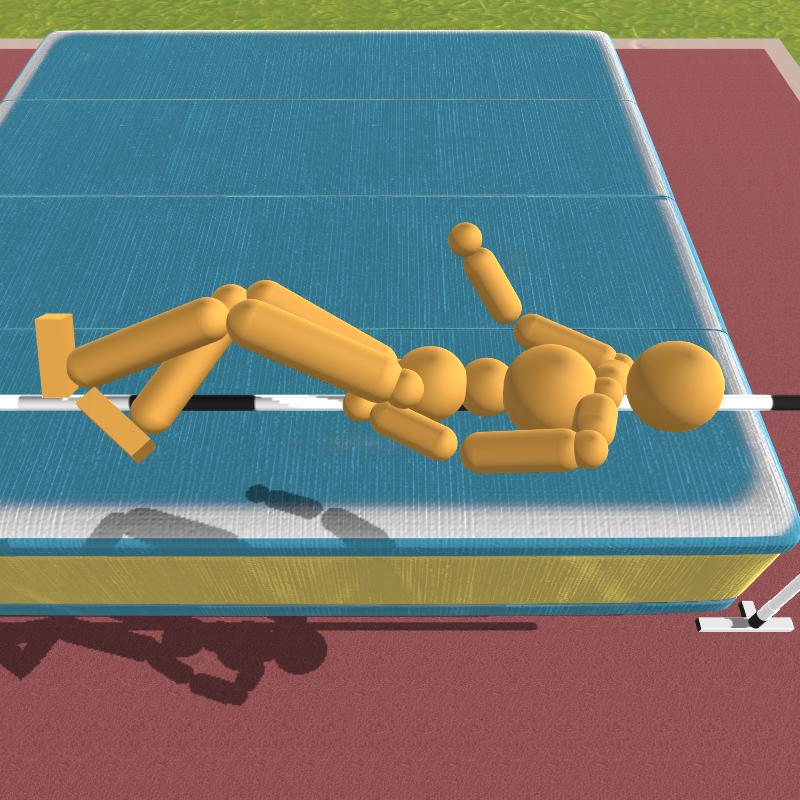}
    \caption{Western Roll*}
    \end{subfigure}
    \begin{subfigure}[b]{0.12\textwidth}
    \includegraphics[width=\linewidth]{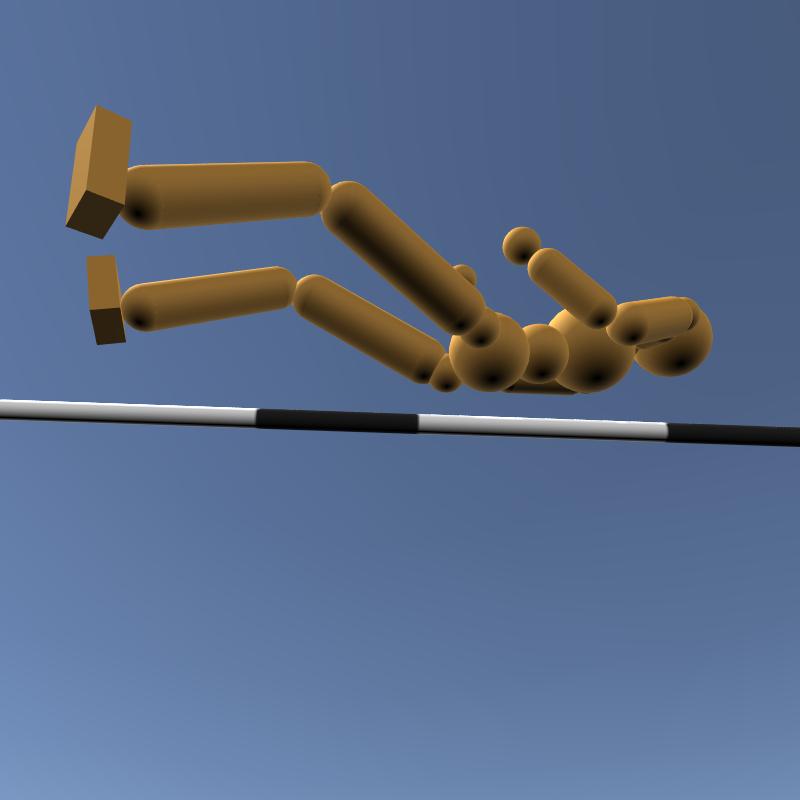}
    \includegraphics[width=\linewidth]{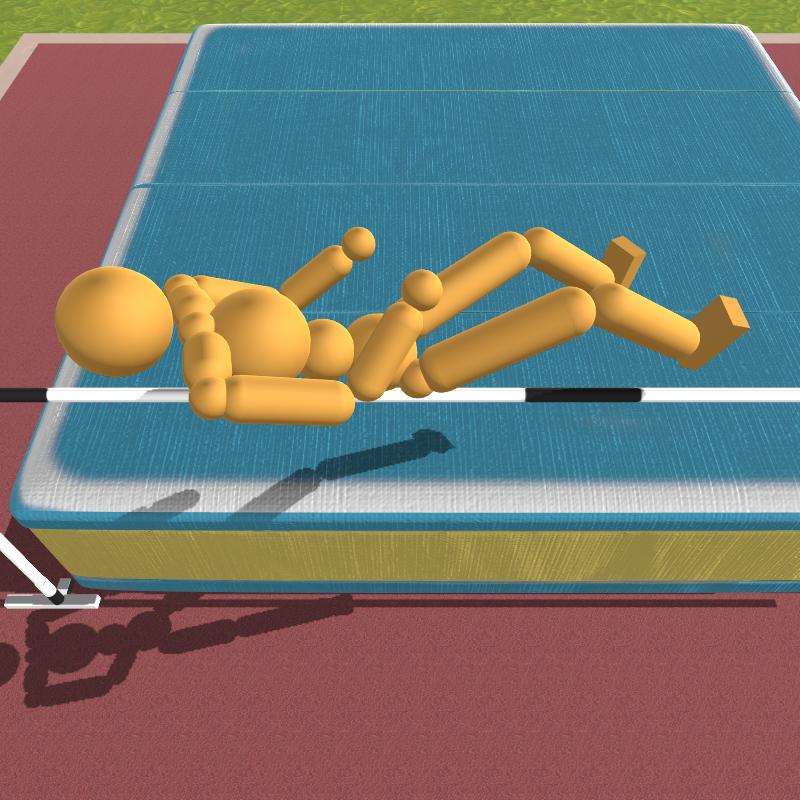}
    \caption{Scissor}
    \end{subfigure}
    \begin{subfigure}[b]{0.12\textwidth}
    \includegraphics[width=\linewidth]{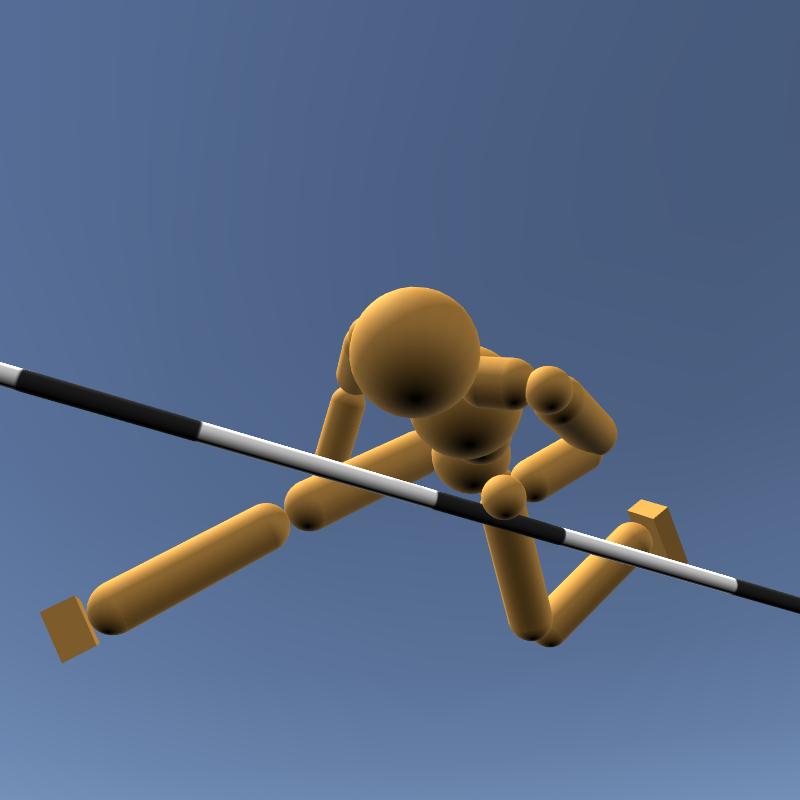}
    \includegraphics[width=\linewidth]{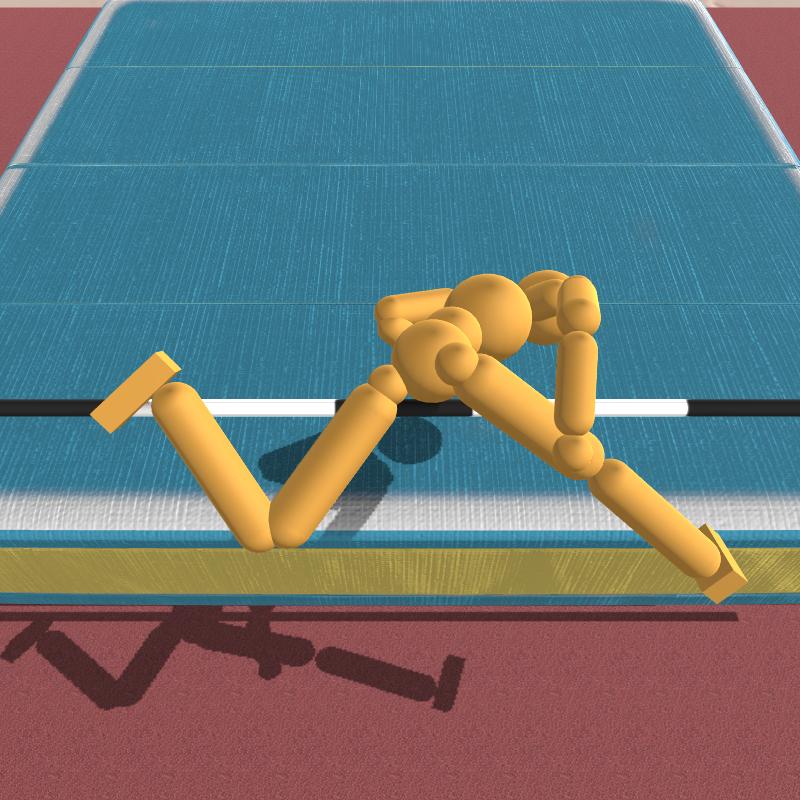}
    \caption{Side Dive}
    \end{subfigure}
    \begin{subfigure}[b]{0.12\textwidth}
    \includegraphics[width=\linewidth]{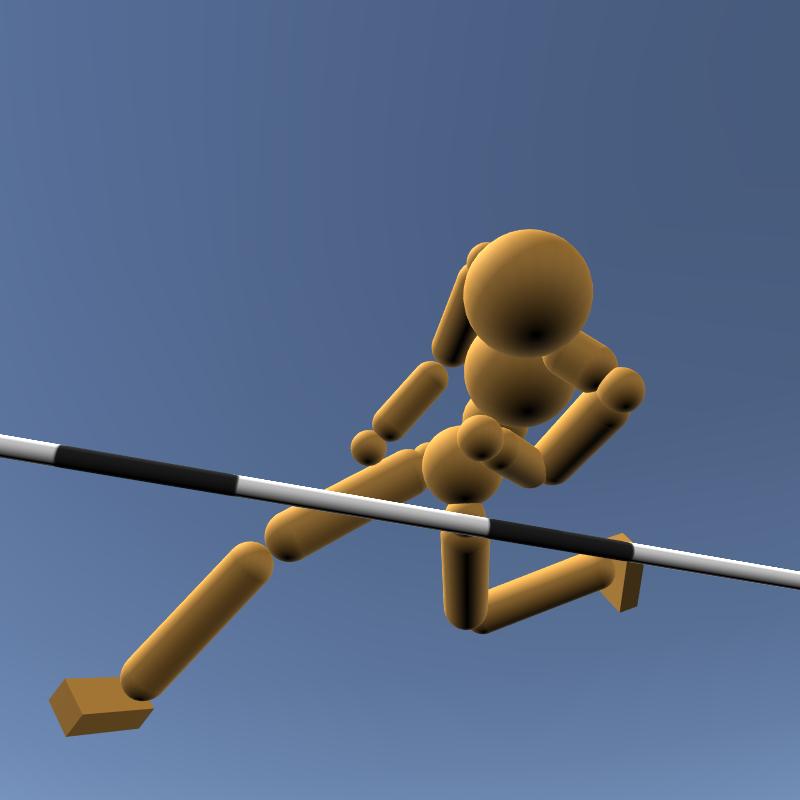}
    \includegraphics[width=\linewidth]{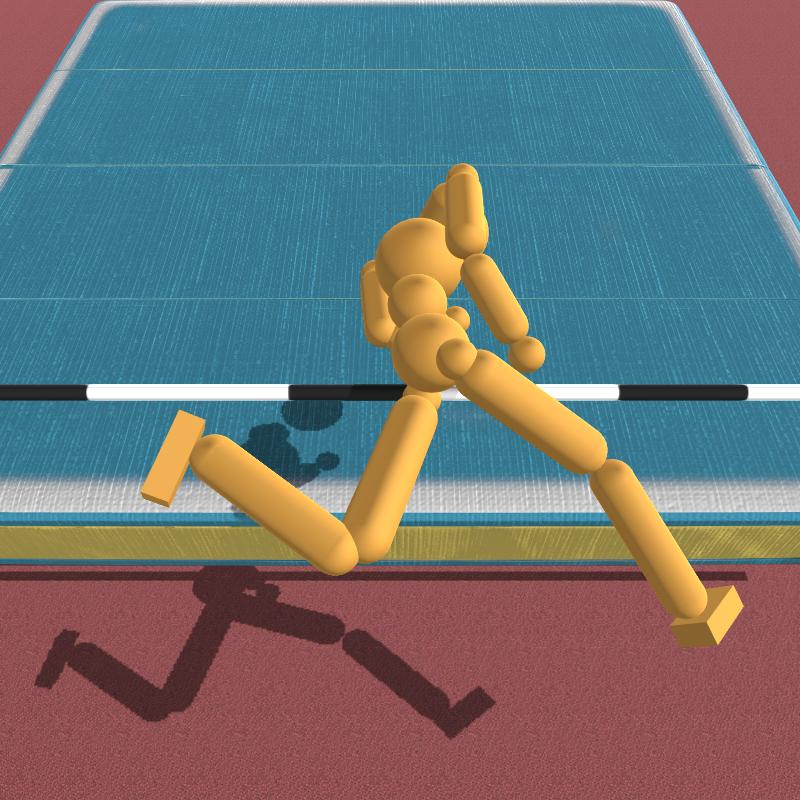}
    \caption{Side Jump}
    \end{subfigure}
    \caption{Peak poses of discovered high jump strategies, ordered by their maximum cleared height. First row: look-up views; Second row: look-down views.}
    \label{fig:High-jump-peak-poses}
\end{figure*}

\begin{figure*}
    \centering
    \begin{subfigure}[b]{\textwidth}
    \includegraphics[width=0.12\linewidth]{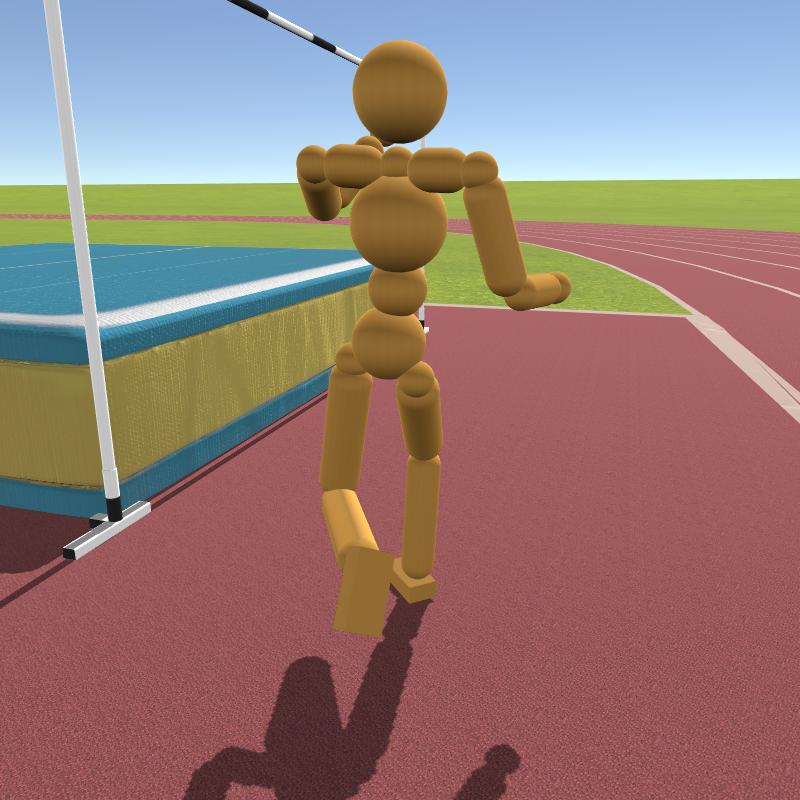}
    \includegraphics[width=0.12\linewidth]{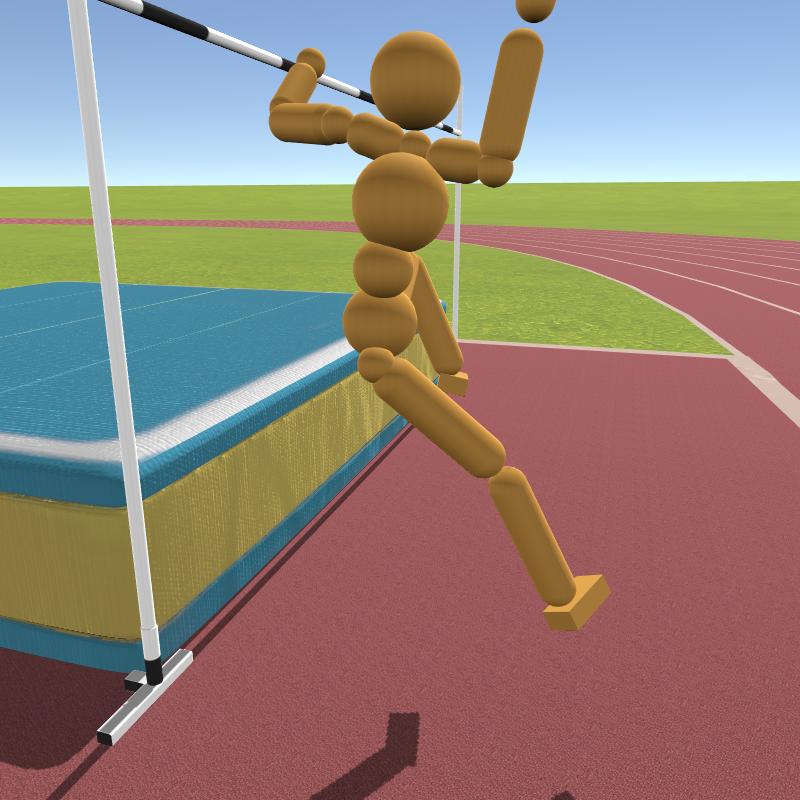}
    \includegraphics[width=0.12\linewidth]{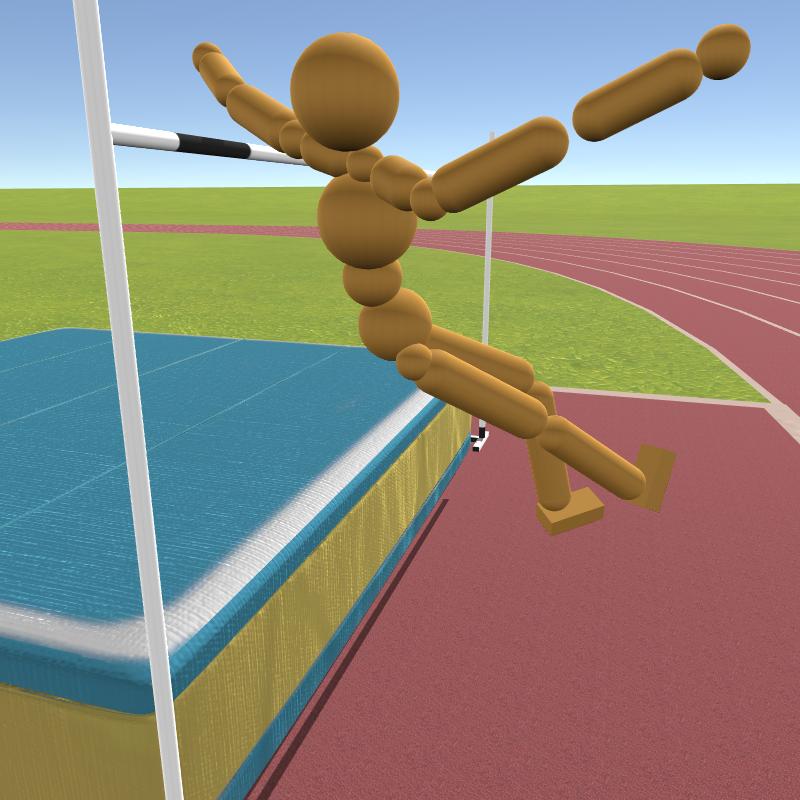}
    \includegraphics[width=0.12\linewidth]{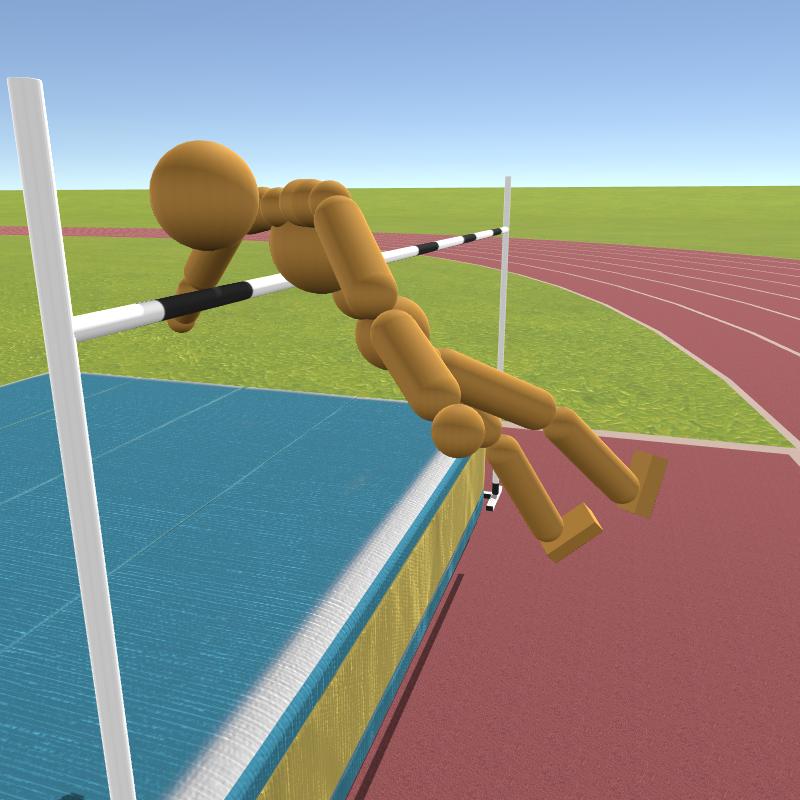}
    \includegraphics[width=0.12\linewidth]{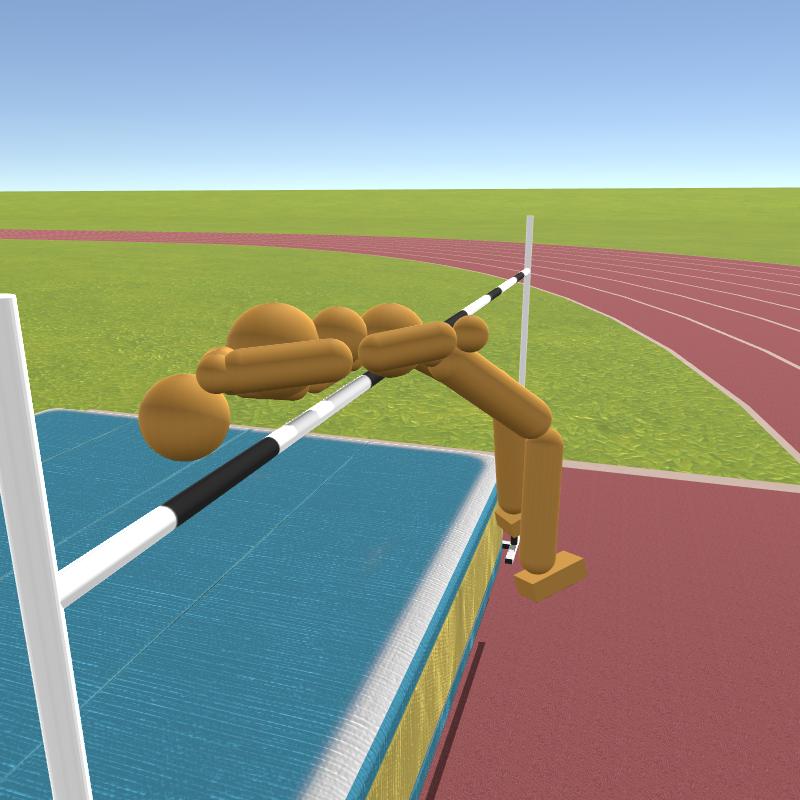}
    \includegraphics[width=0.12\linewidth]{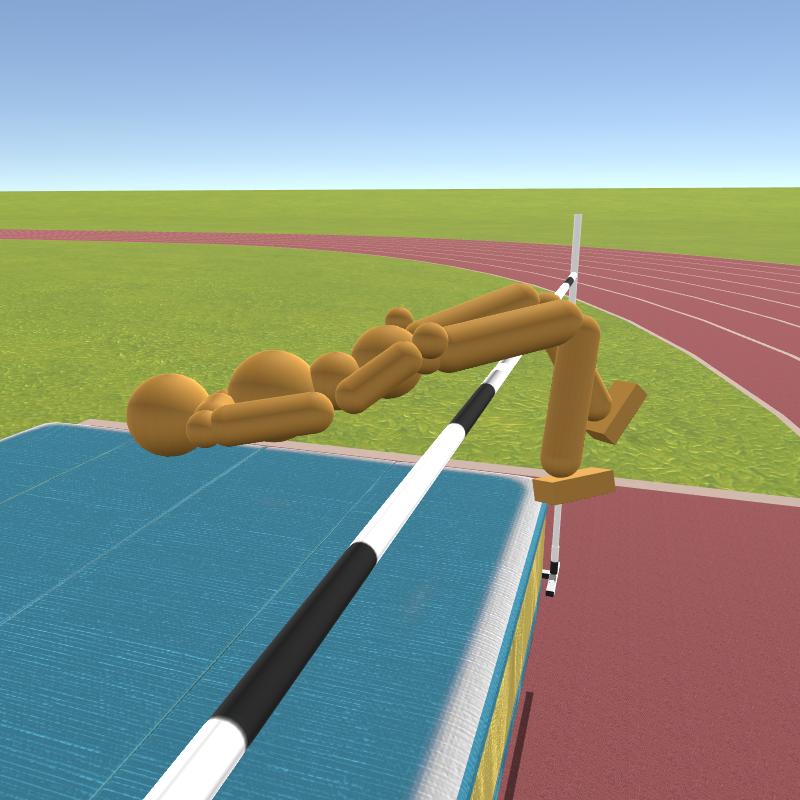}
    \includegraphics[width=0.12\linewidth]{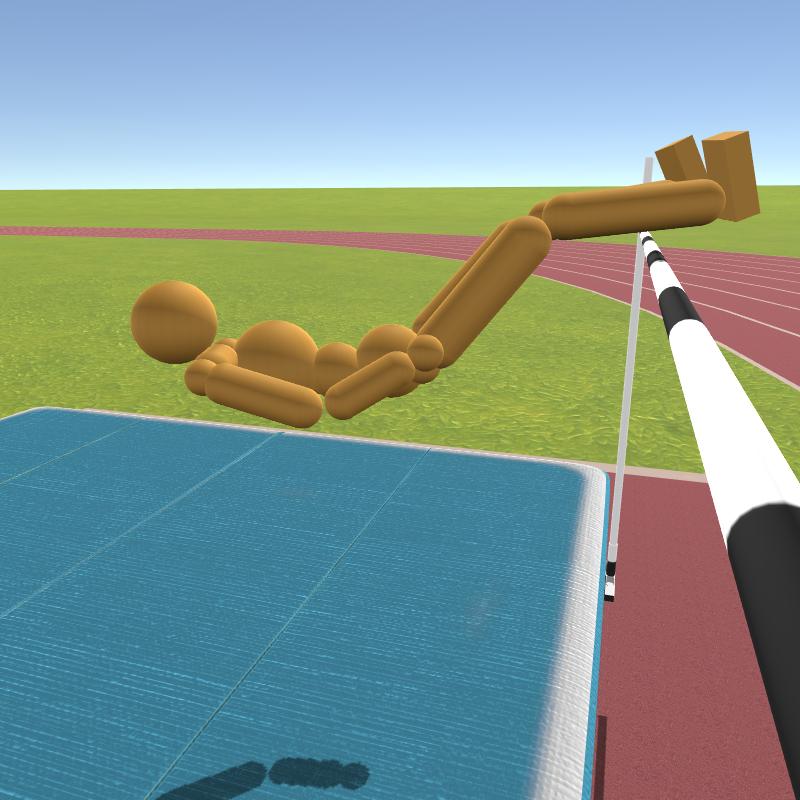}
    \includegraphics[width=0.12\linewidth]{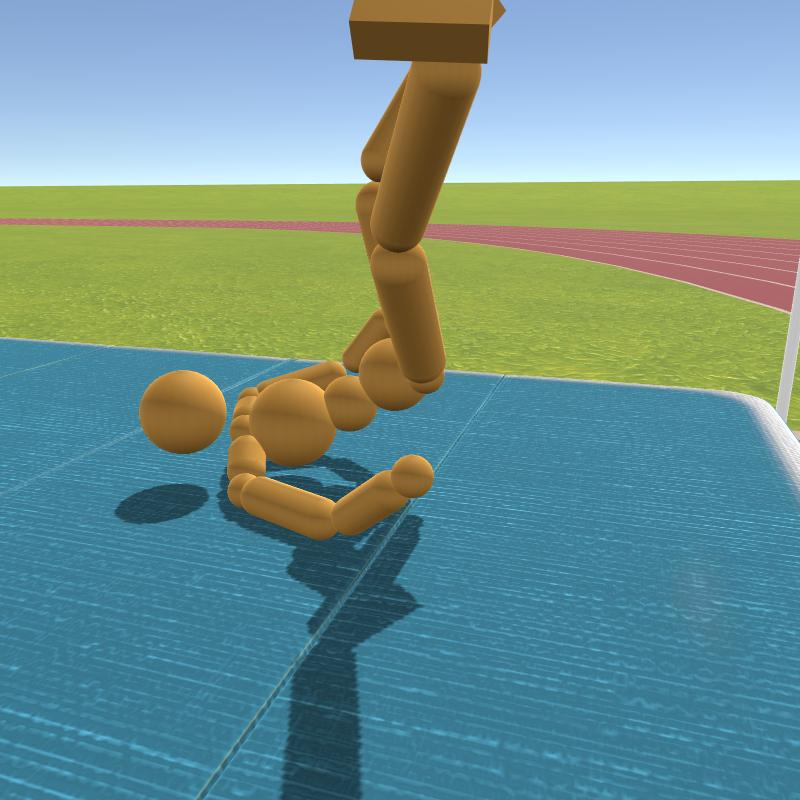} \\
    \includegraphics[width=0.12\linewidth]{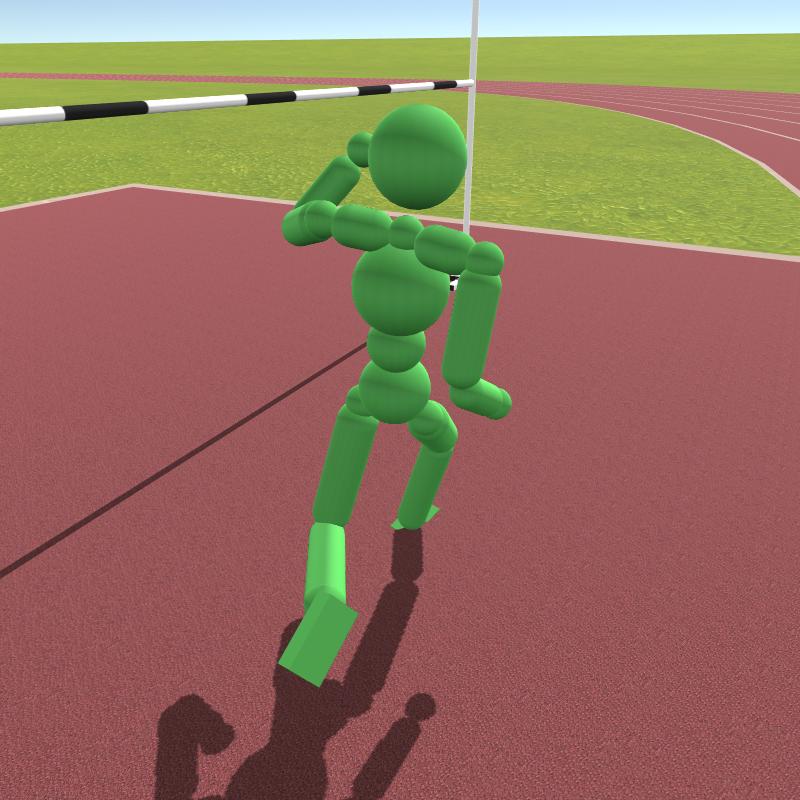}
    \includegraphics[width=0.12\linewidth]{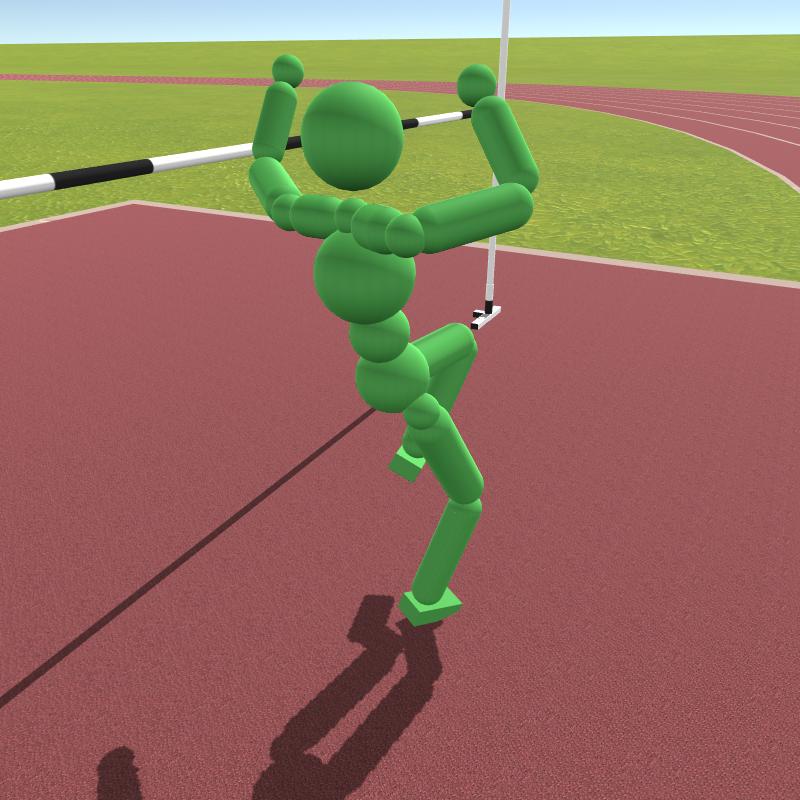}
    \includegraphics[width=0.12\linewidth]{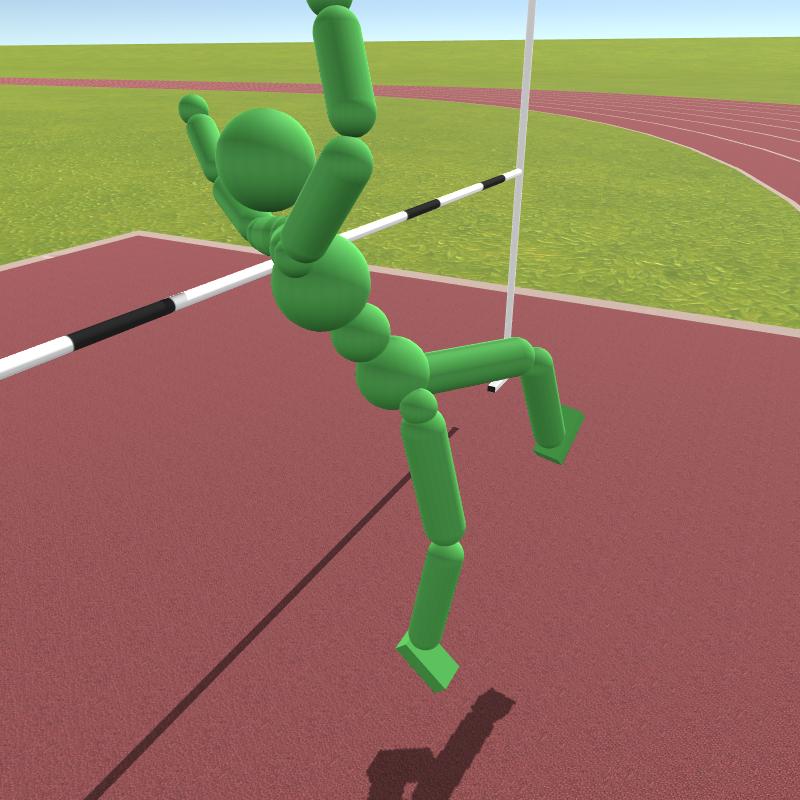}
    \includegraphics[width=0.12\linewidth]{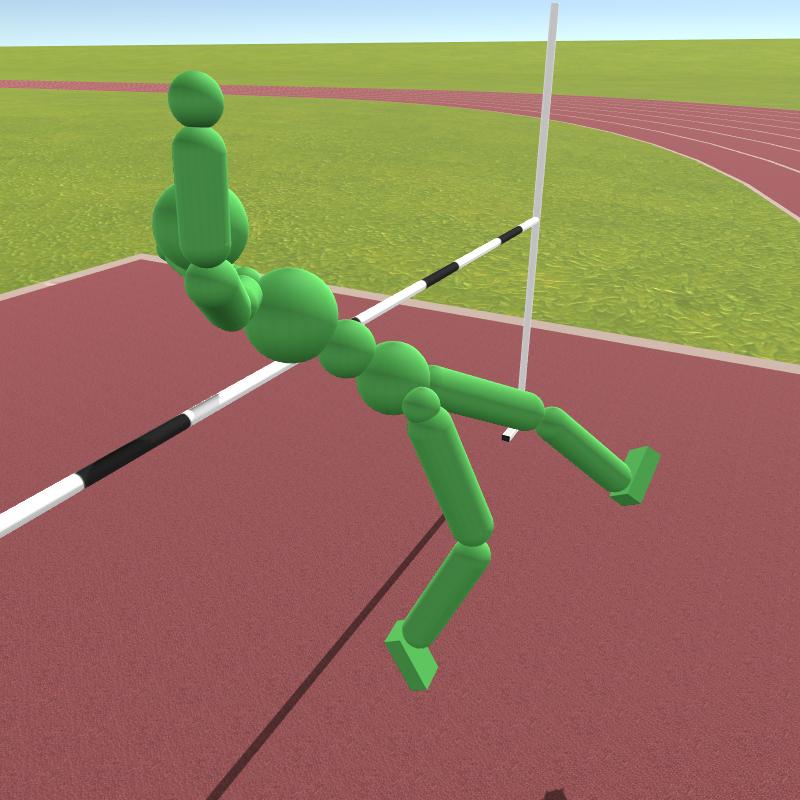}
    \includegraphics[width=0.12\linewidth]{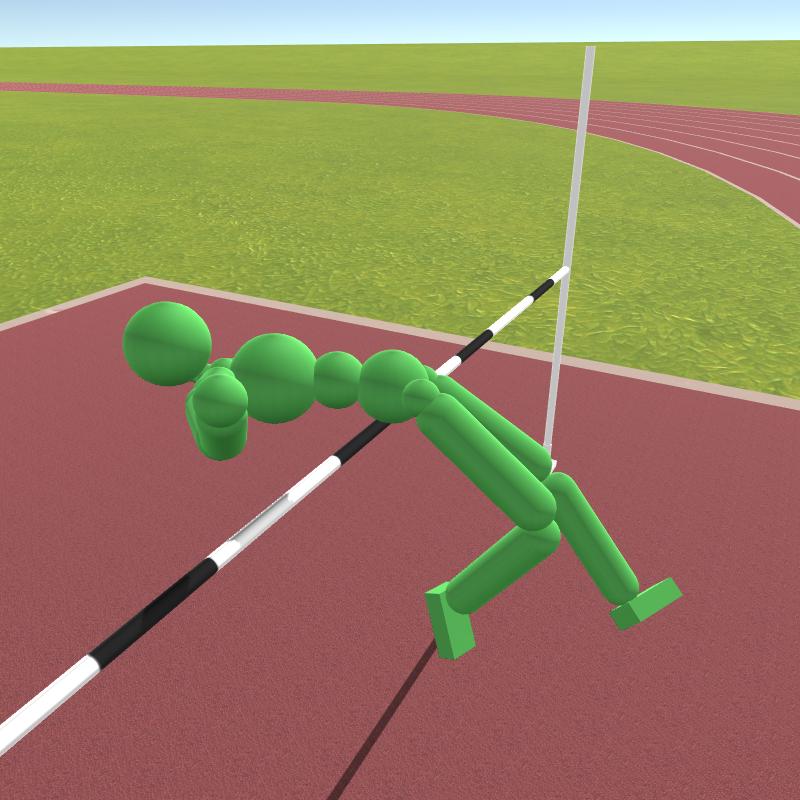}
    \includegraphics[width=0.12\linewidth]{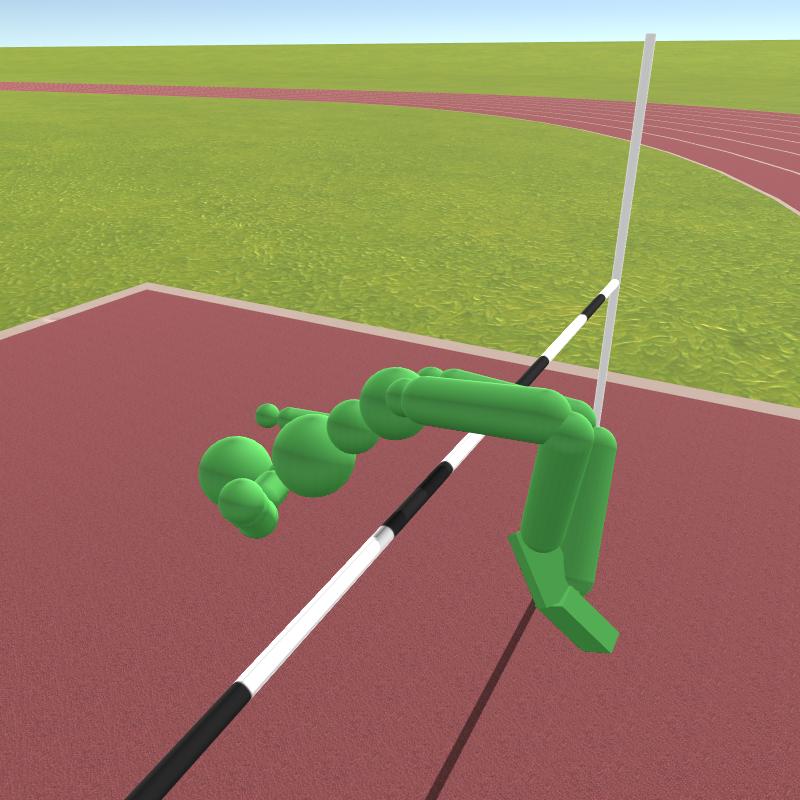}
    \includegraphics[width=0.12\linewidth]{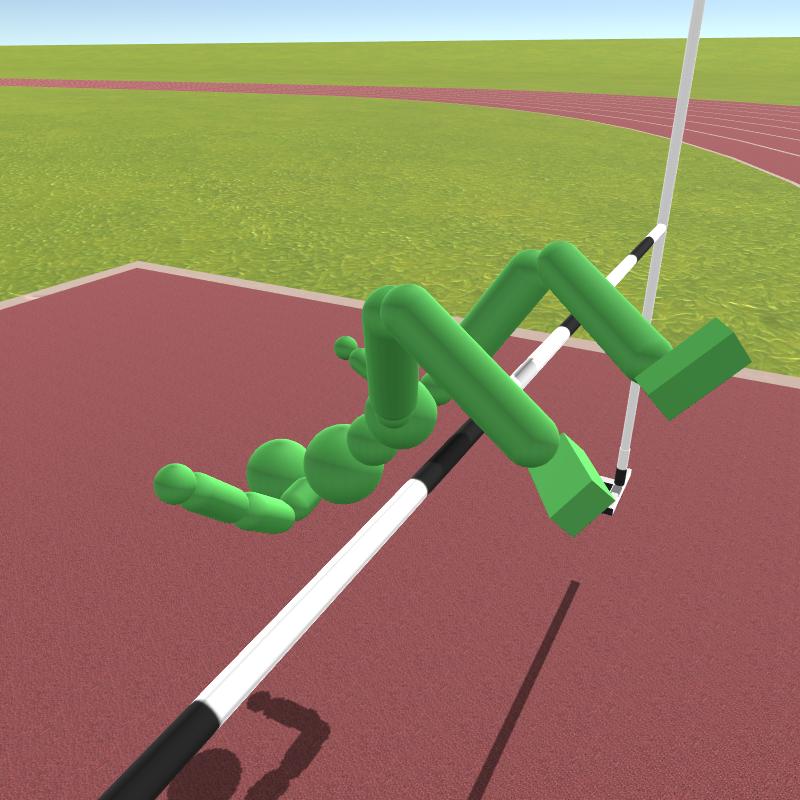}
    \includegraphics[width=0.12\linewidth]{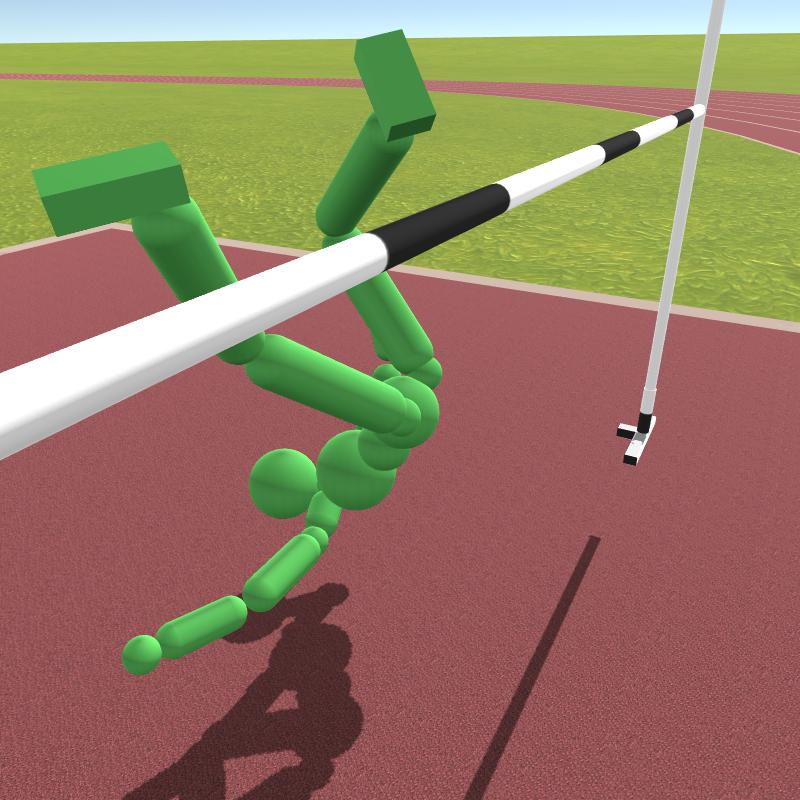}
    \caption{Fosbury Flop. First row: synthesized -- max height=$200cm$; Second row: motion capture -- capture height=$130cm$.}
    \end{subfigure}
    \begin{subfigure}[b]{\textwidth}
    \includegraphics[width=0.12\linewidth]{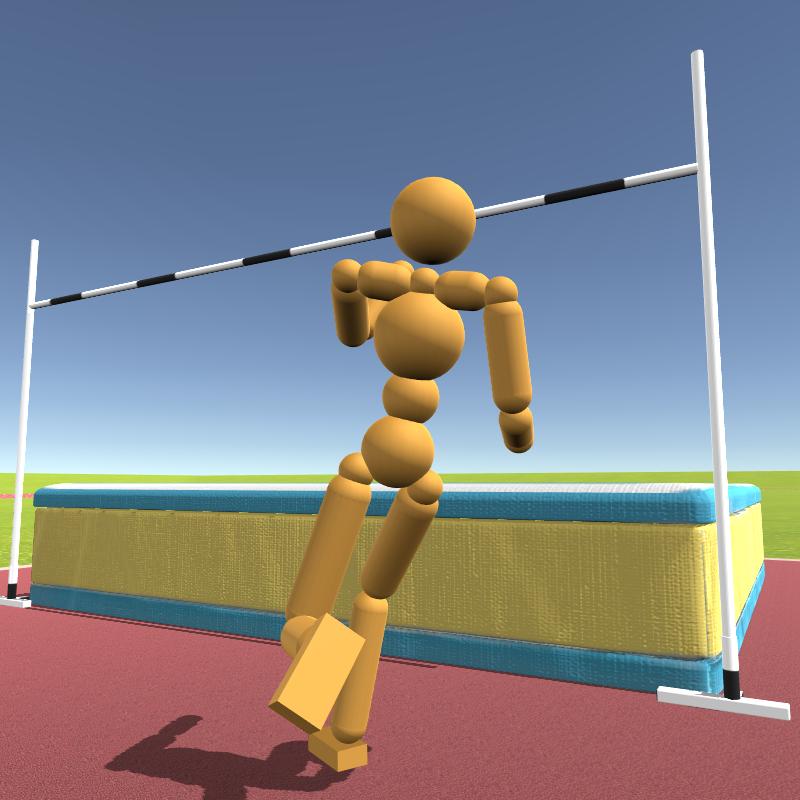}
    \includegraphics[width=0.12\linewidth]{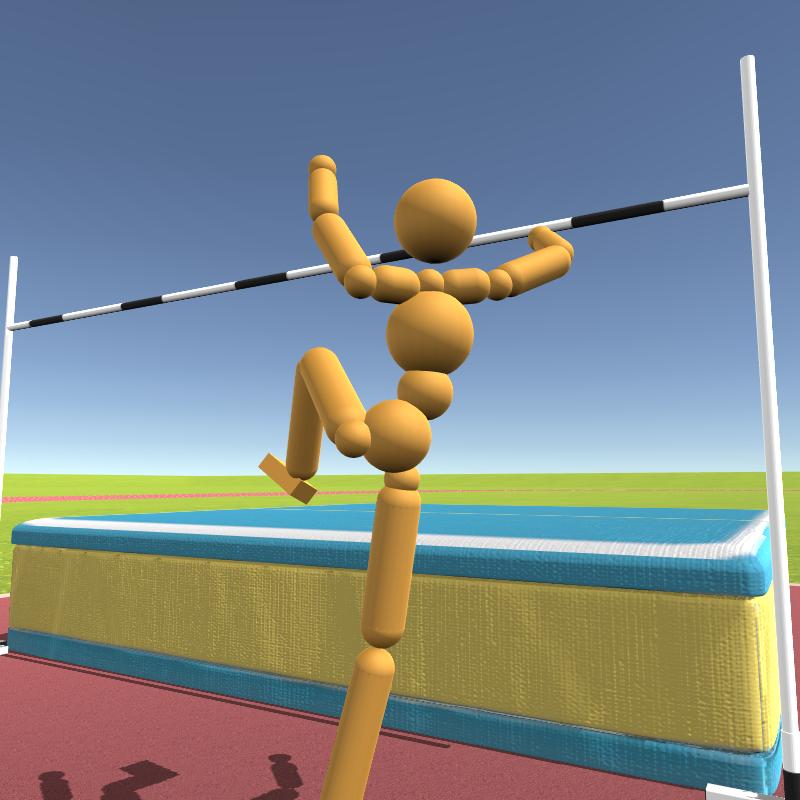}
    \includegraphics[width=0.12\linewidth]{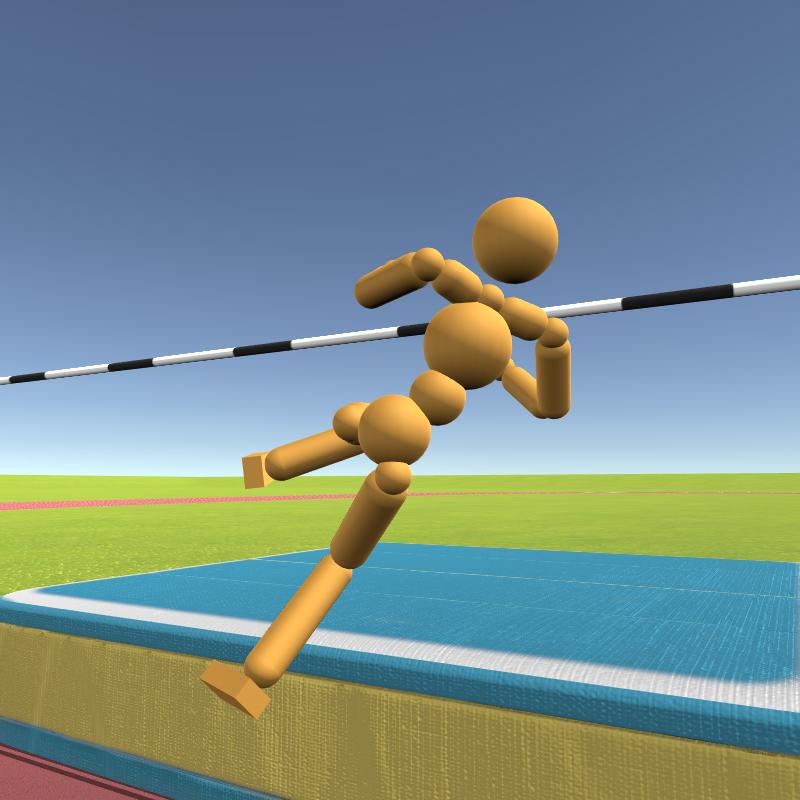}
    \includegraphics[width=0.12\linewidth]{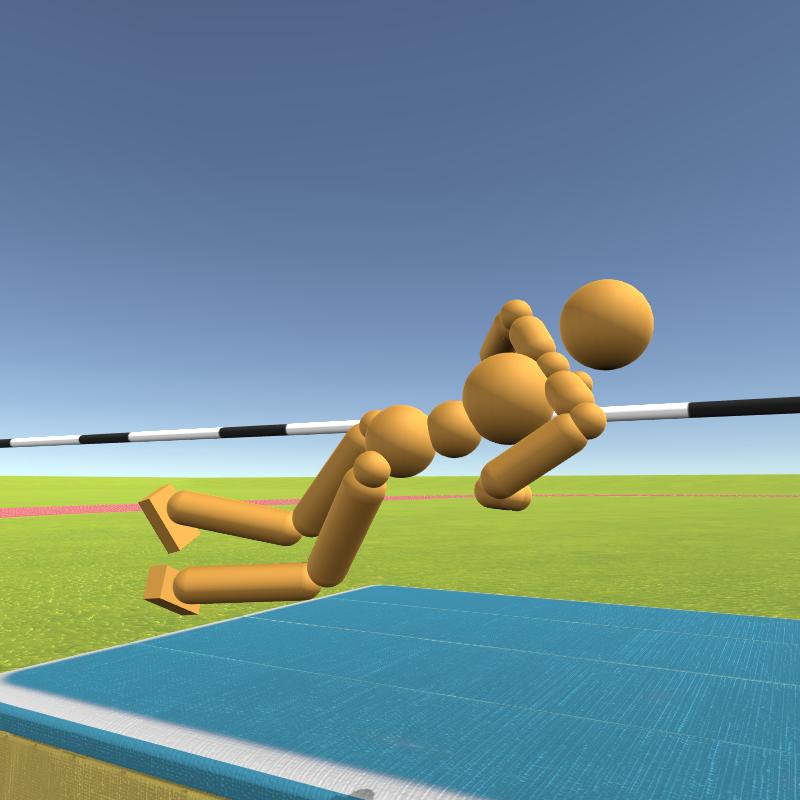}
    \includegraphics[width=0.12\linewidth]{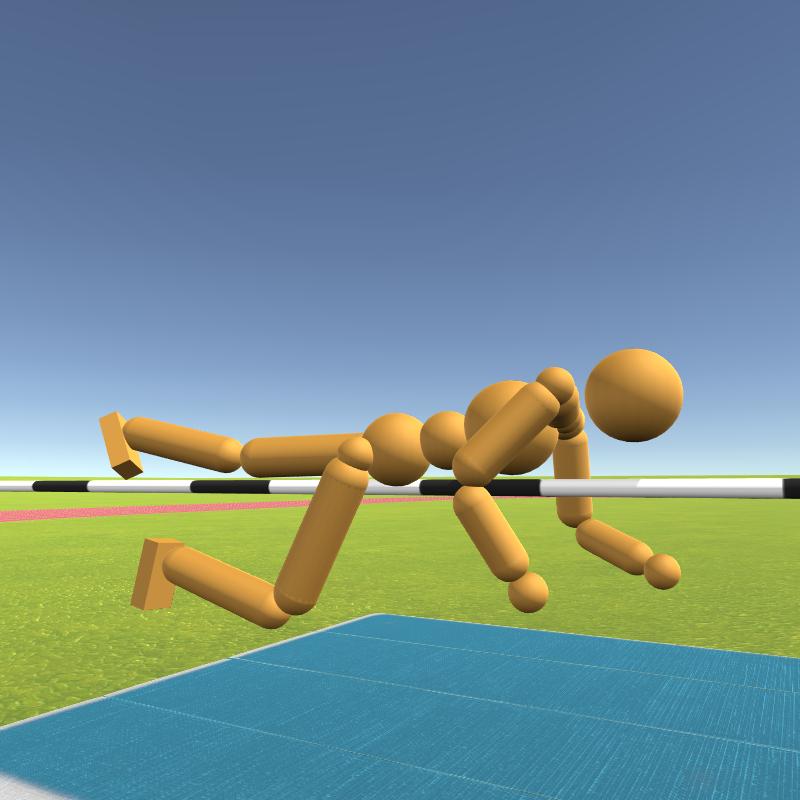}
    \includegraphics[width=0.12\linewidth]{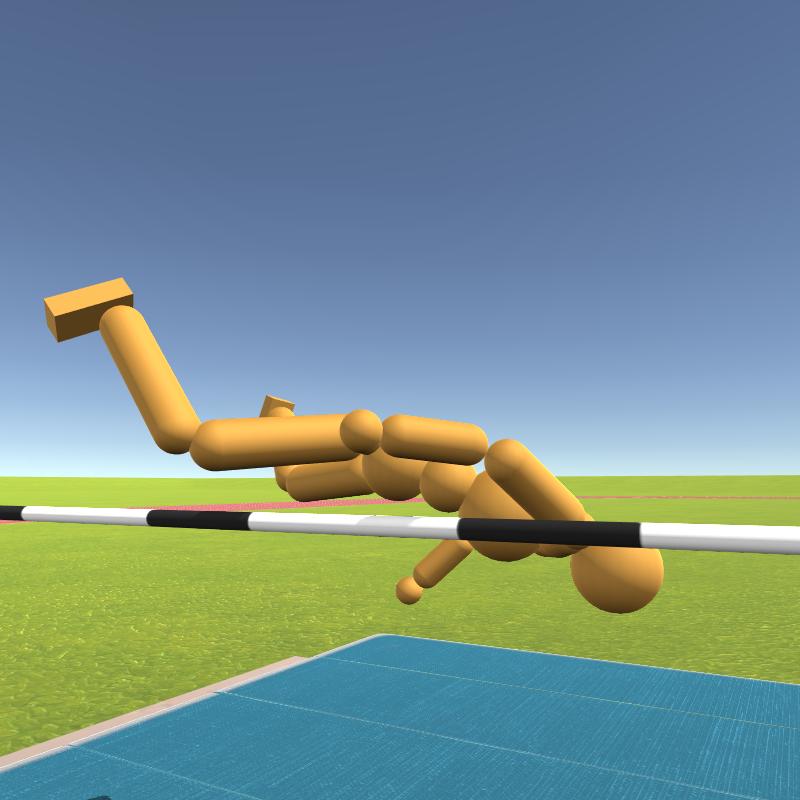}
    \includegraphics[width=0.12\linewidth]{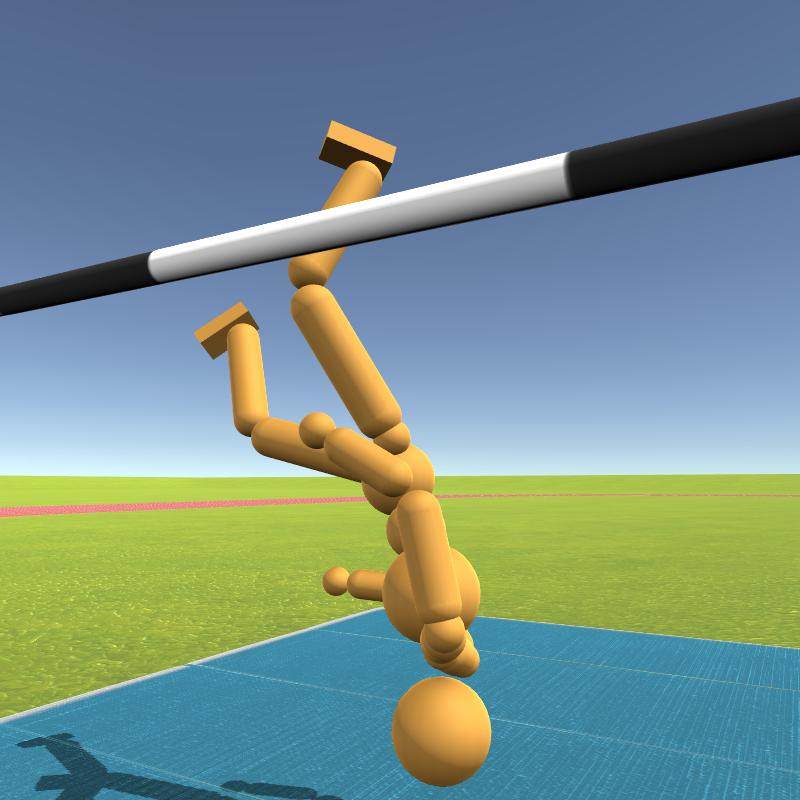}
    \includegraphics[width=0.12\linewidth]{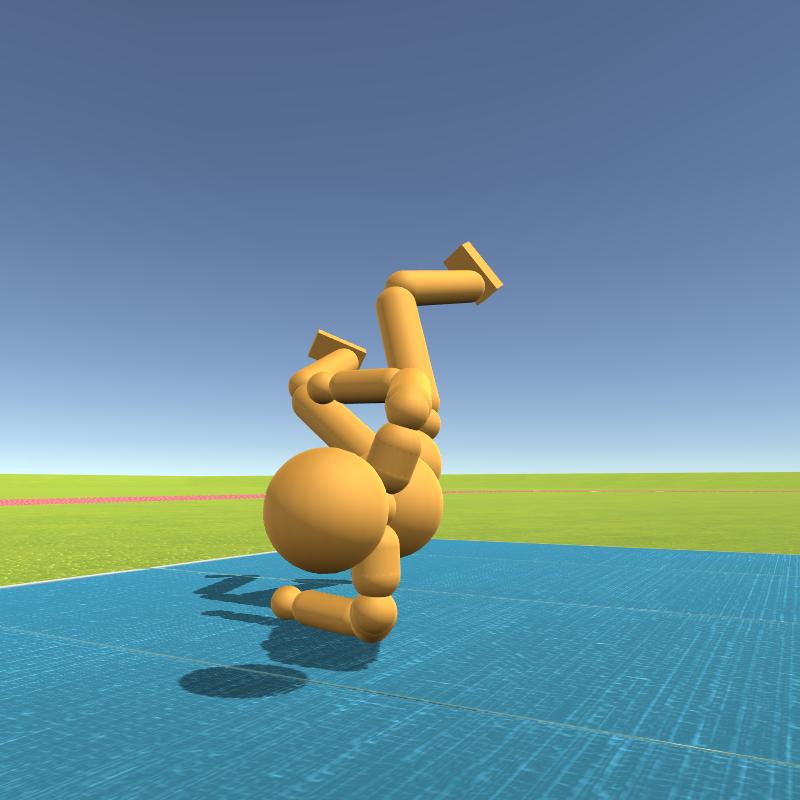} \\
    \includegraphics[width=0.12\linewidth]{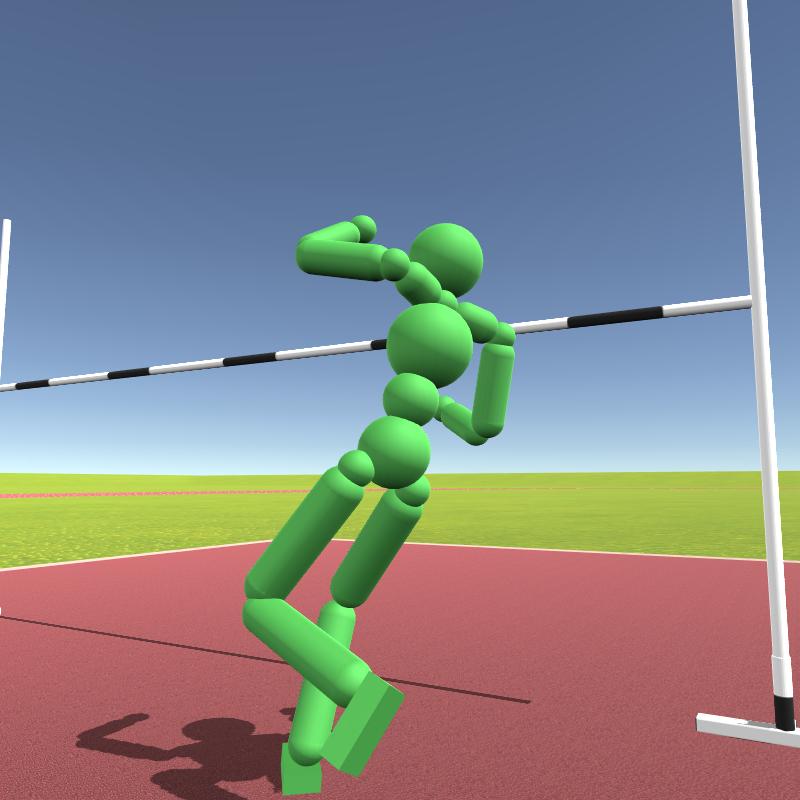}
    \includegraphics[width=0.12\linewidth]{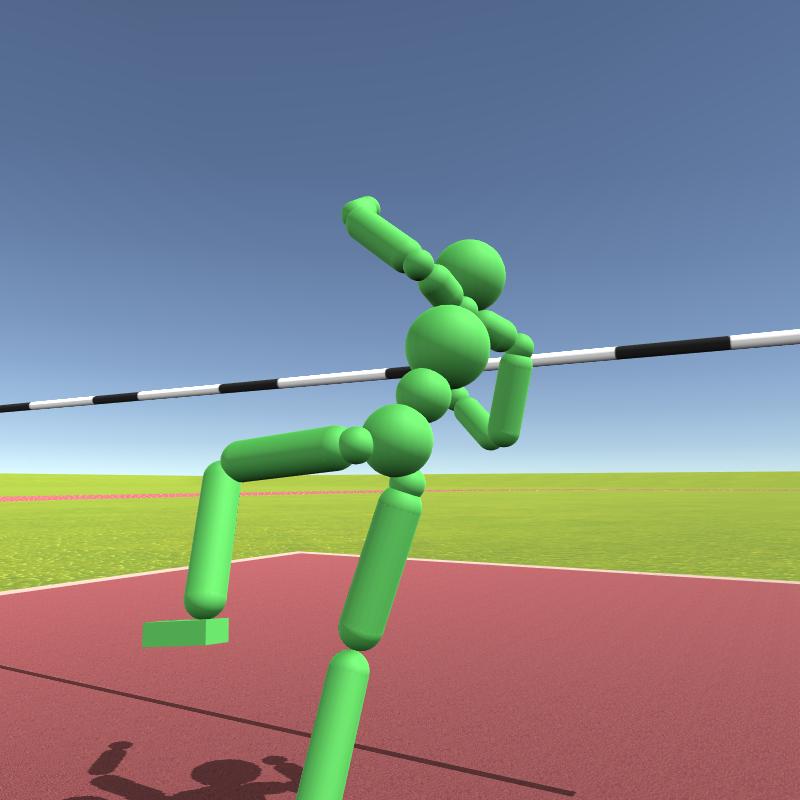}
    \includegraphics[width=0.12\linewidth]{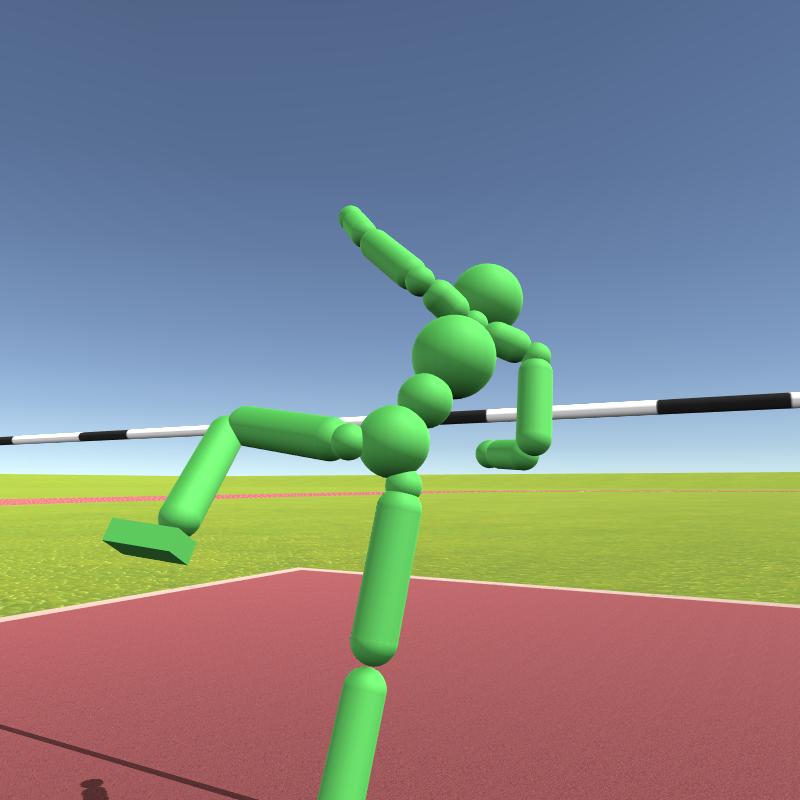}
    \includegraphics[width=0.12\linewidth]{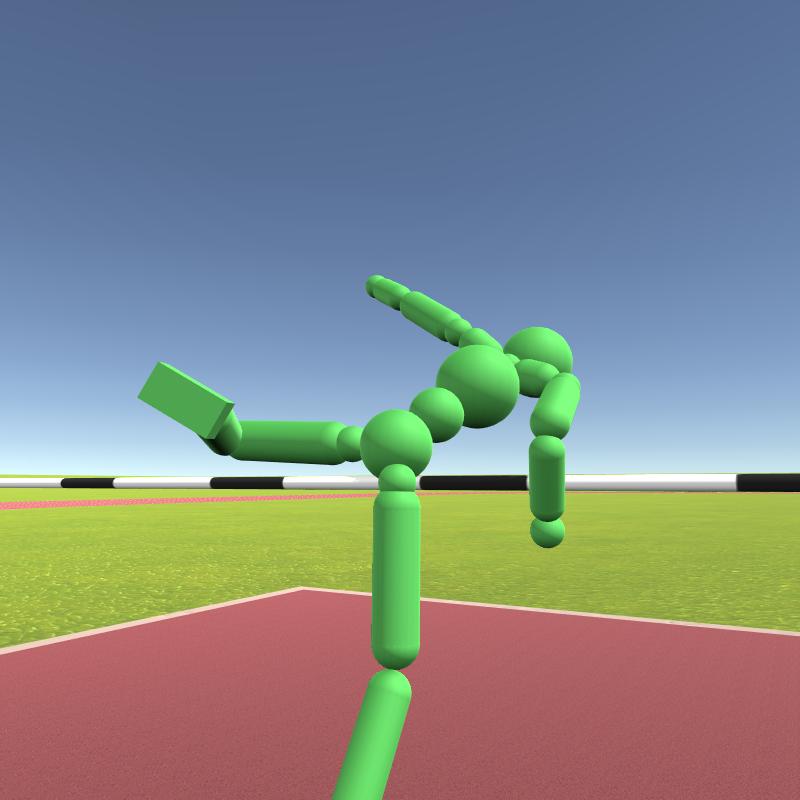}
    \includegraphics[width=0.12\linewidth]{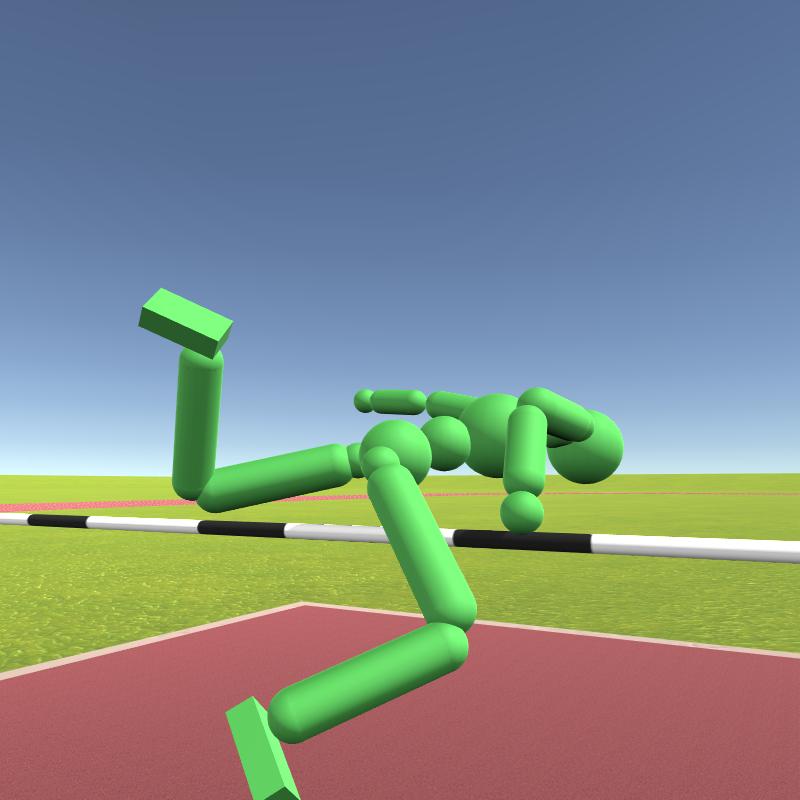}
    \includegraphics[width=0.12\linewidth]{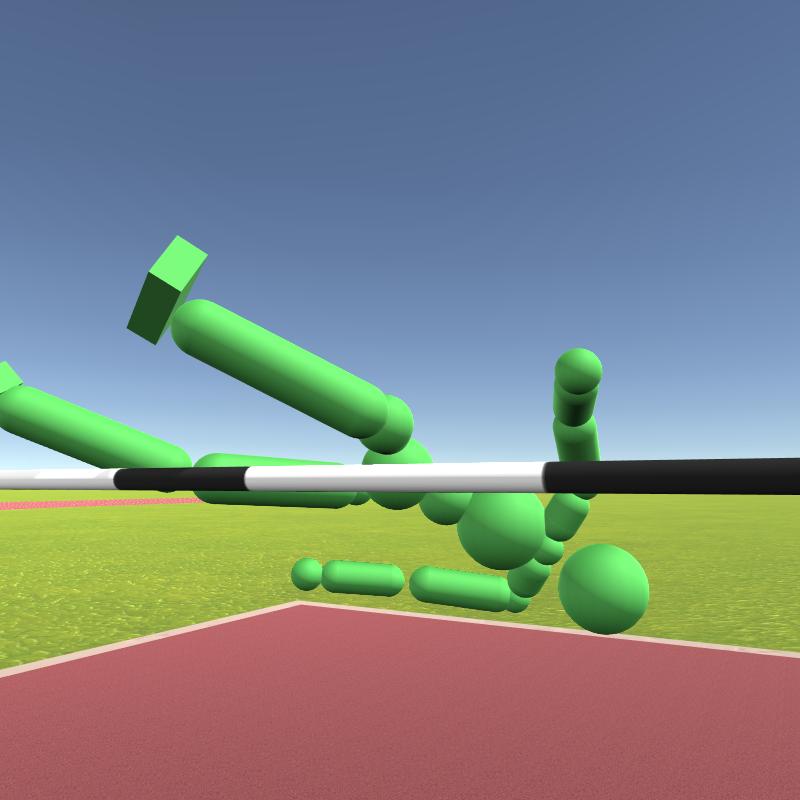}
    \includegraphics[width=0.12\linewidth]{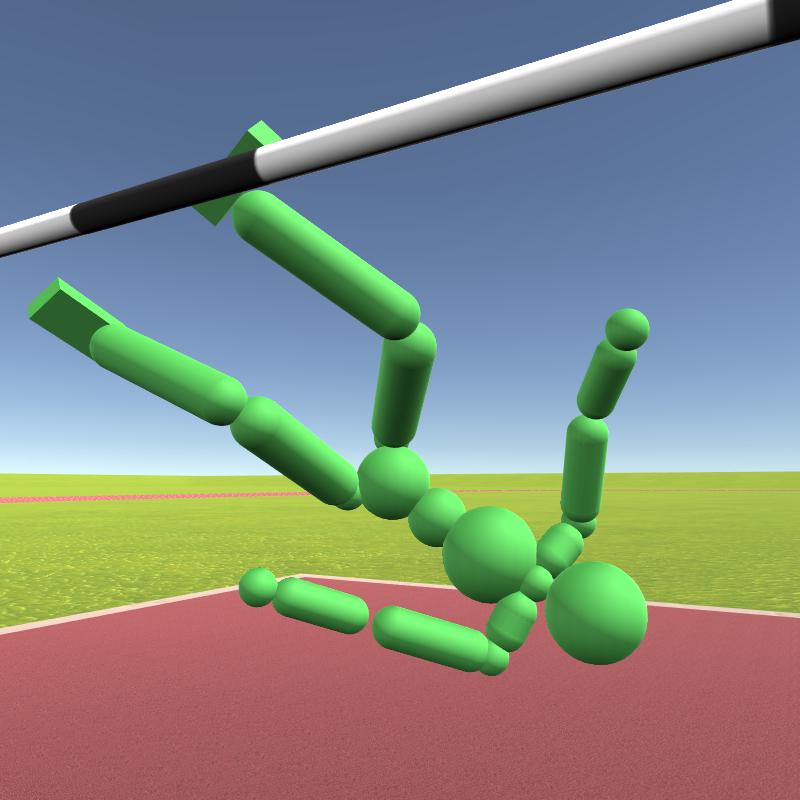}
    \includegraphics[width=0.12\linewidth]{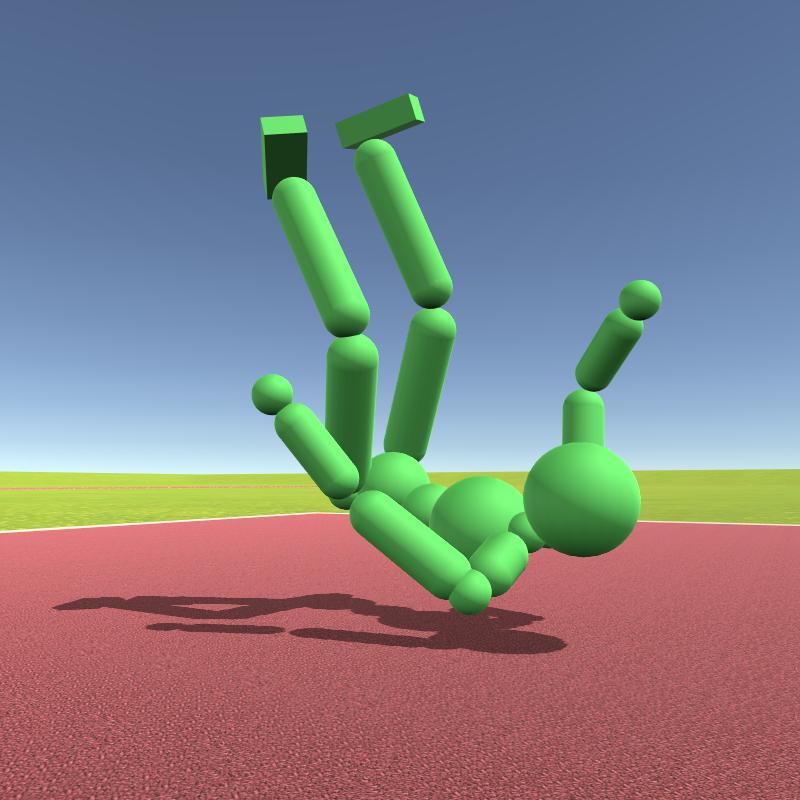}
    \caption{Straddle. First row: synthesized -- max height=$195cm$; Second row: motion capture -- capture height=$130cm$.}
    \end{subfigure}
    \caption{Comparison of our synthesized high jumps with those captured from a human athlete.}
    \label{fig:compare-mocap}
\end{figure*}

\begin{figure}
    \centering
    \begin{subfigure}[b]{0.99\linewidth}
        \includegraphics[width=0.99\linewidth]{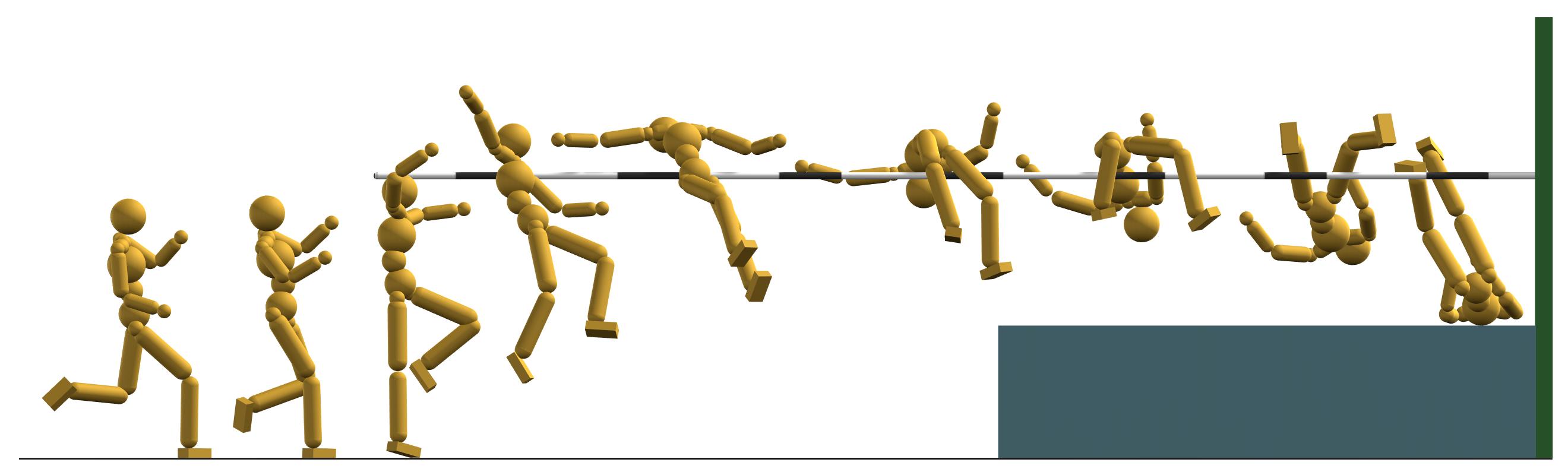}
        \caption{Fosbury Flop -- max height=$160cm$, performed by a character with a weaker take-off leg, whose take-off hip, knee and ankle torque limits are set to $60\%$ of the normal values.}
        \label{fig:variation-weakerLeg}
    \end{subfigure}
    \begin{subfigure}[b]{0.99\linewidth}
        \includegraphics[width=0.99\linewidth]{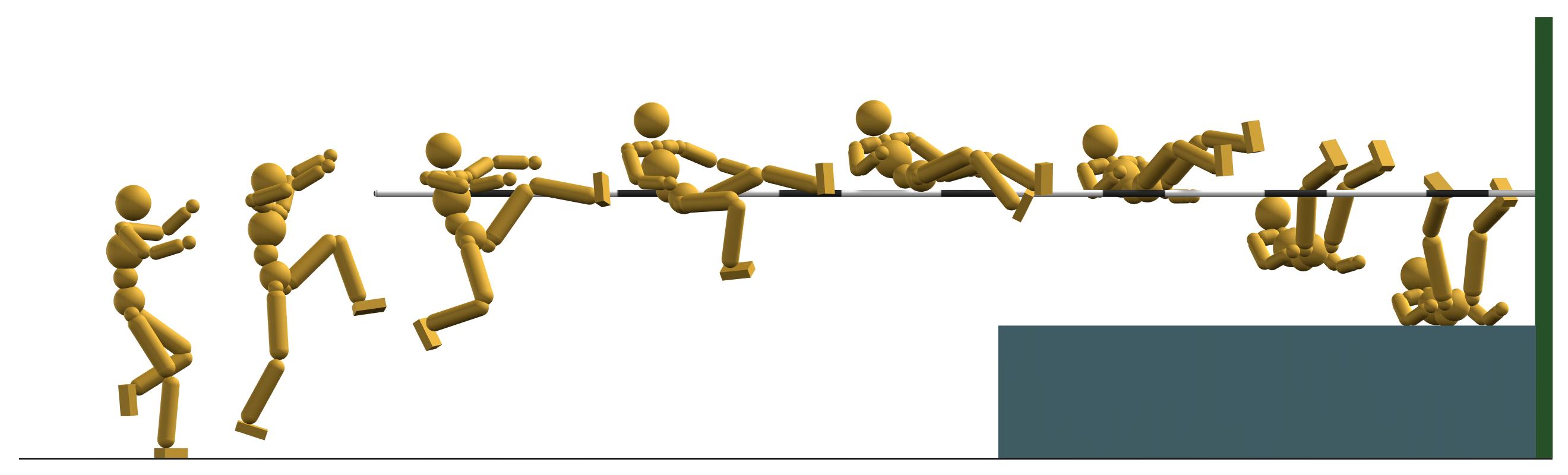}
        \caption{Fosbury Flop -- max height=$150cm$, performed by a character with an inflexible spine that does not permit arching backwards.}
        \label{fig:variation-inflexibleSpine}
    \end{subfigure}
    \begin{subfigure}[b]{0.99\linewidth}
        \includegraphics[width=0.99\linewidth]{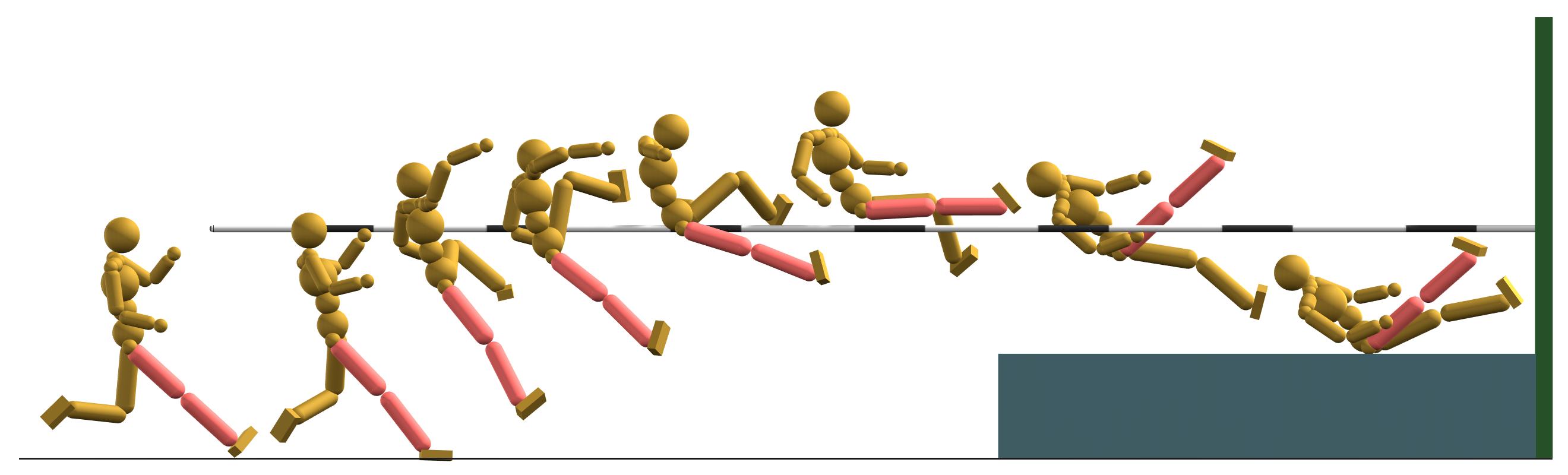}
        \caption{Scissor Kick -- max height=$130cm$, learned by a character with a cast on its take-off leg.}
        \label{fig:variation-fixedKnee}
    \end{subfigure}
    \begin{subfigure}[b]{0.99\linewidth}
    \includegraphics[width=0.99\linewidth]{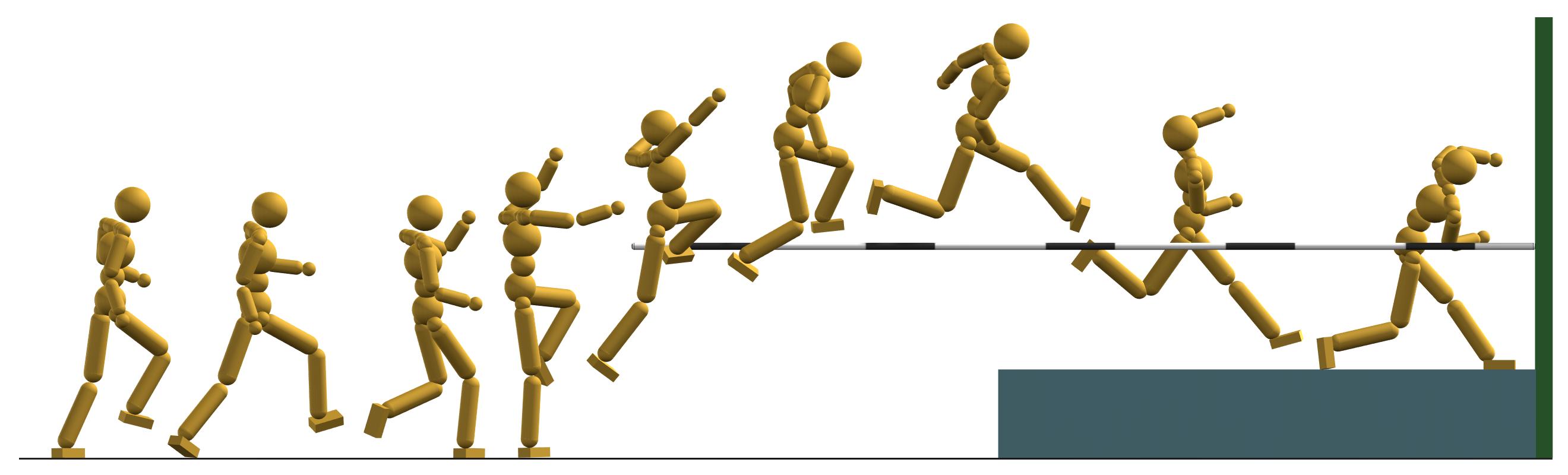}
    \caption{Front Kick -- max height=$120cm$, performed with an additional constraint requiring landing on feet.}
    \label{fig:variation-landonFoot}
    \end{subfigure}
    \caption{High jump variations. The first three policies are trained from the initial state of the Fosbury Flop discovered in Stage 1, and the last policy is trained from the initial state of the Front Kick discovered in Stage 1.}
    \label{fig:variations}
\end{figure}

\begin{figure*}
    \centering
    \begin{subfigure}[b]{0.99\linewidth}
        \includegraphics[width=0.495\linewidth]{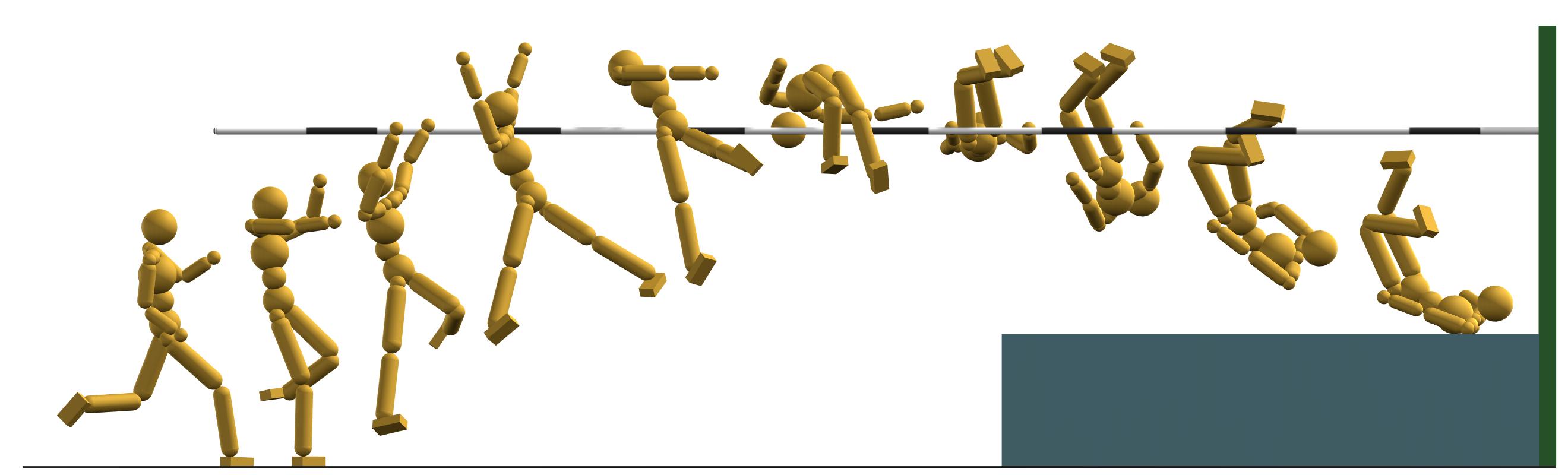}
        \includegraphics[width=0.495\linewidth]{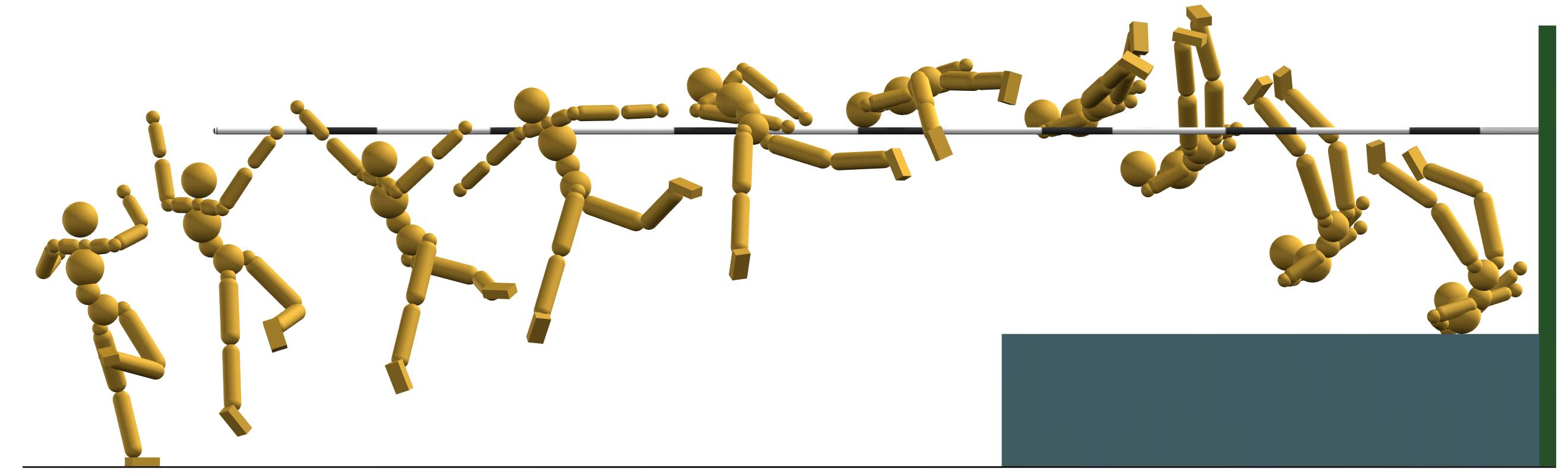}
        \caption{High jumps trained without P-VAE, given the initial state of Fosbury Flop and Straddle respectively. Please compare with Figure~\ref{fig:highjump-flop} and Figure~\ref{fig:highjump-straddle}.}
    \end{subfigure}
    \begin{subfigure}[b]{0.99\linewidth}
        \includegraphics[width=0.495\linewidth]{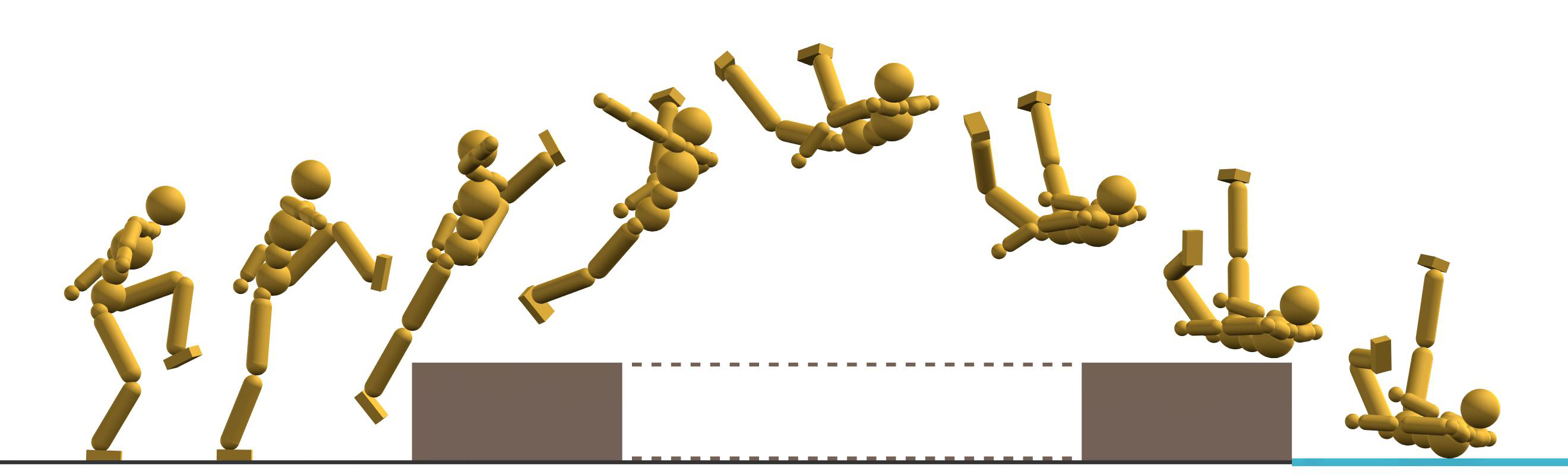}
        \includegraphics[width=0.495\linewidth]{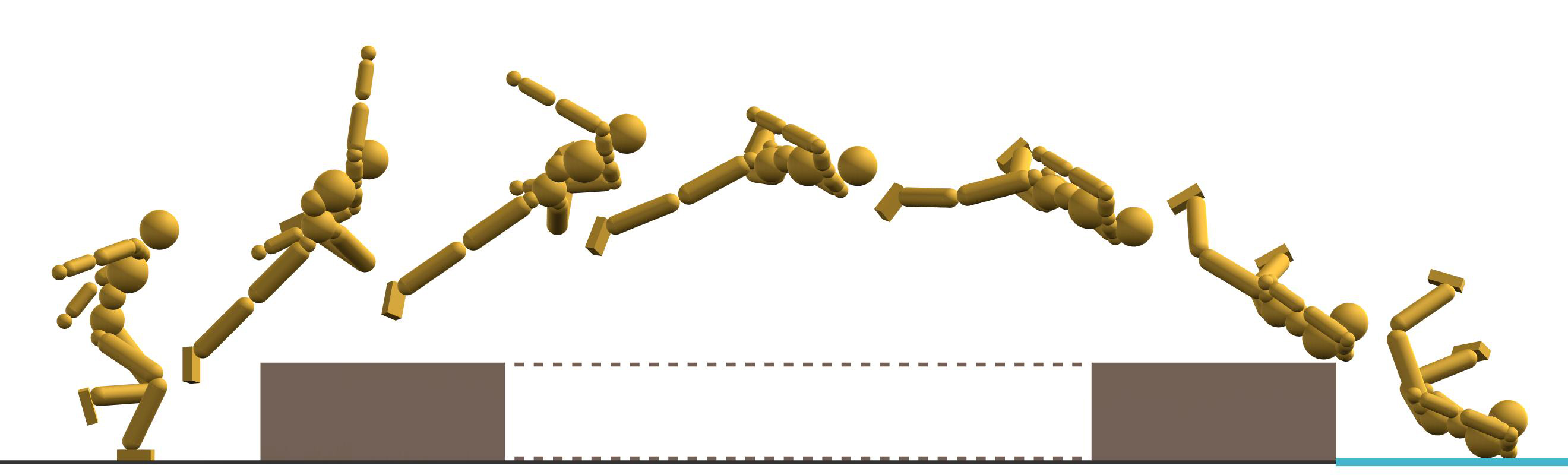}
        \caption{Obstacle jumps trained without P-VAE, given the initial state of Straddle and Twist Jump (cc) respectively. Please compare with Figure~\ref{fig:obstacle-straddle} and Figure~\ref{fig:obstacle-twistJumpCC}.}
    \end{subfigure}
    \caption{Jumping strategies learned without P-VAE. Although the character can still complete the tasks, the poses are less natural.}
    \label{fig:ablation}
\end{figure*}

\begin{figure*}
    \centering
    \includegraphics[width=0.12\linewidth]{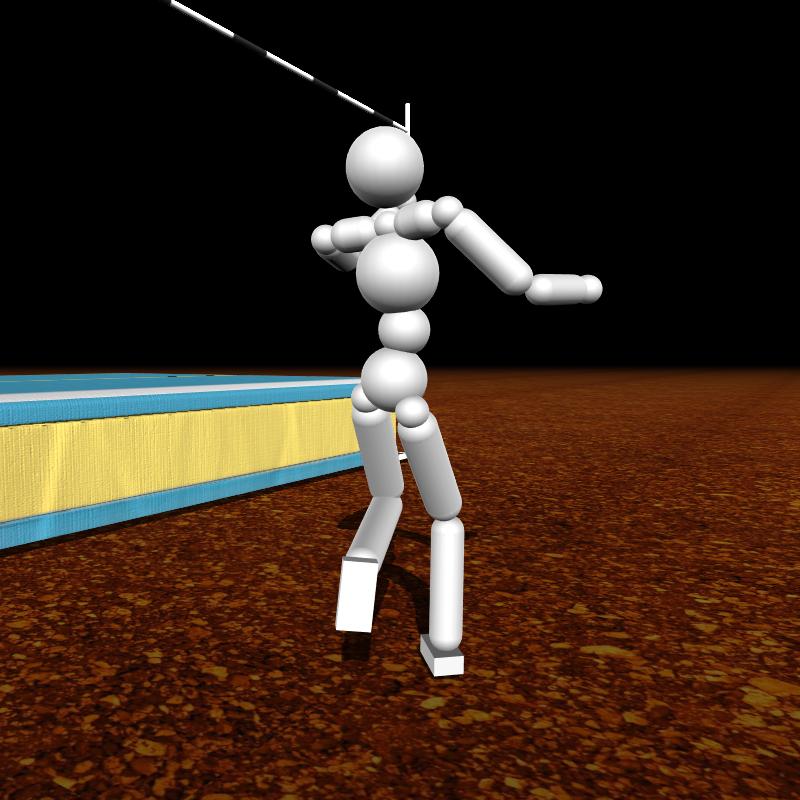}
    \includegraphics[width=0.12\linewidth]{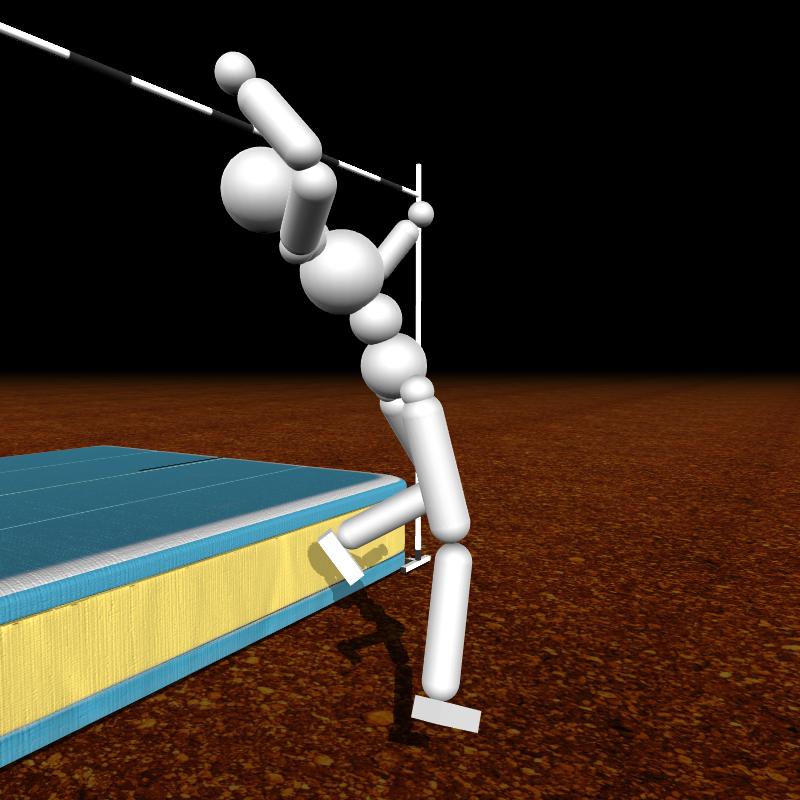}
    \includegraphics[width=0.12\linewidth]{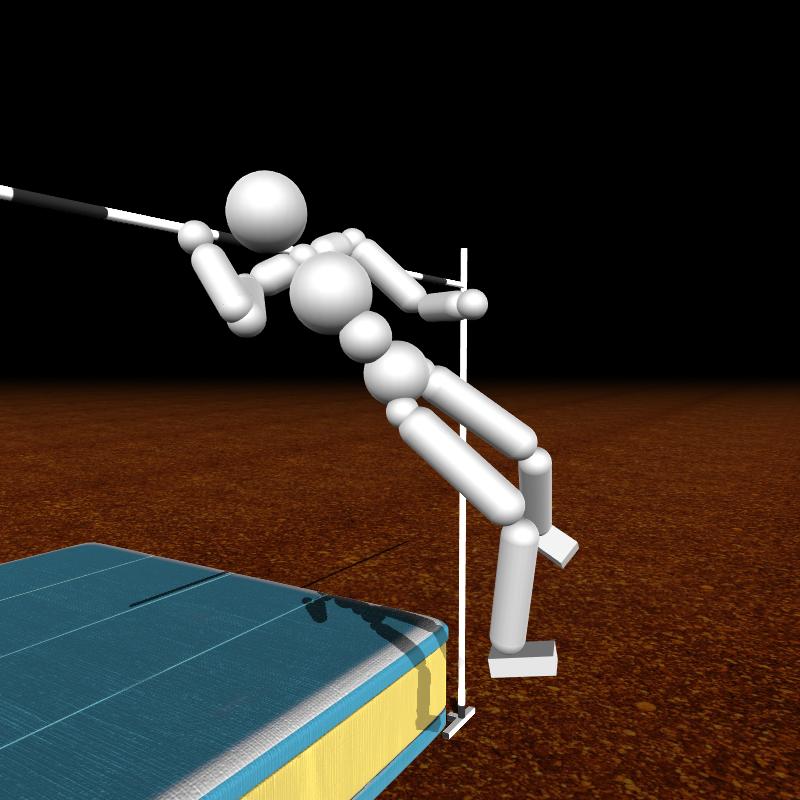}
    \includegraphics[width=0.12\linewidth]{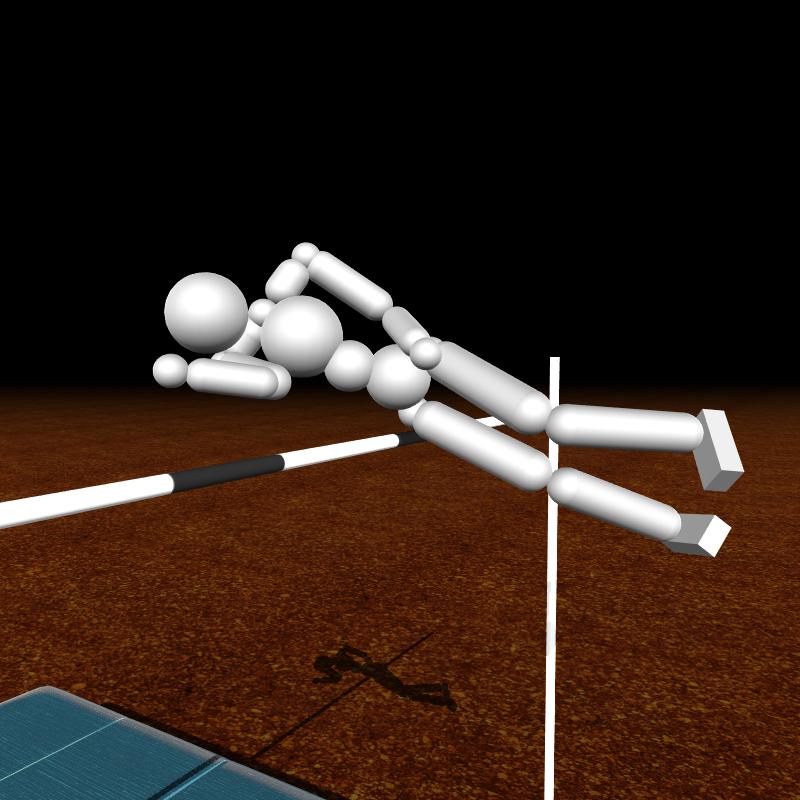}
    \includegraphics[width=0.12\linewidth]{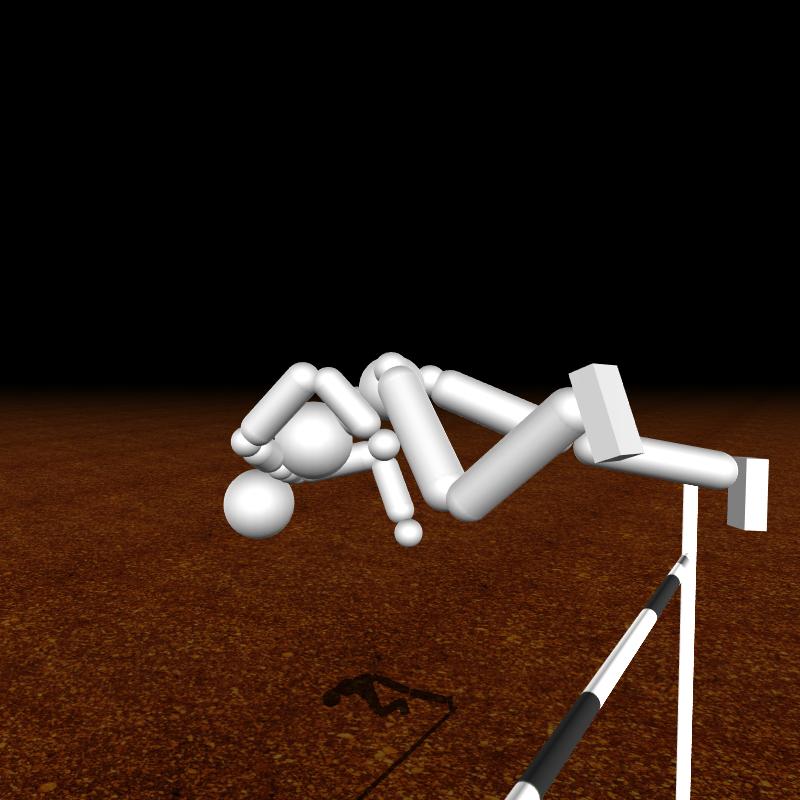}
    \includegraphics[width=0.12\linewidth]{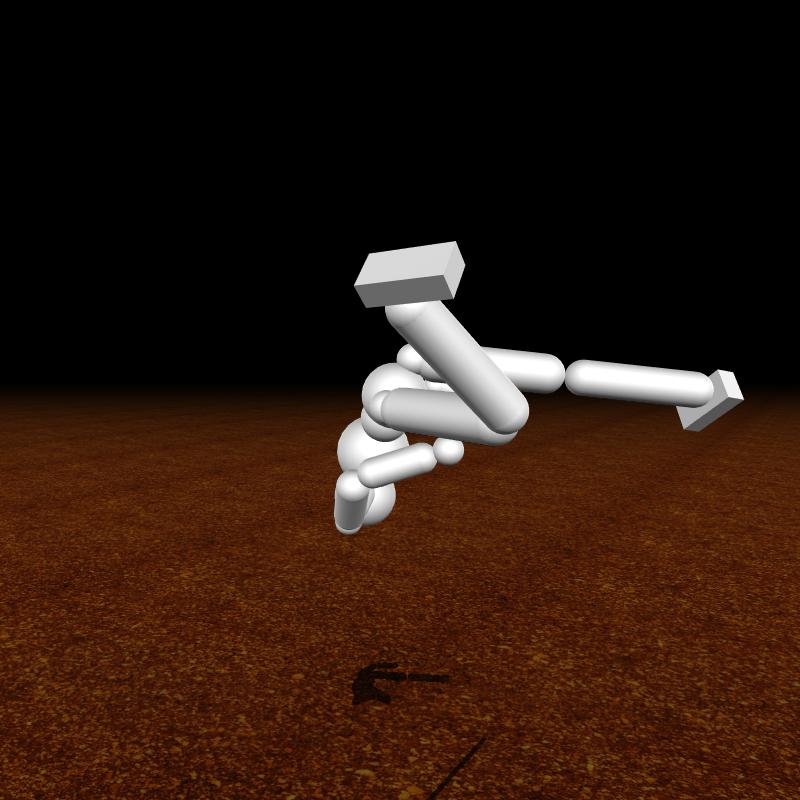}
    \includegraphics[width=0.12\linewidth]{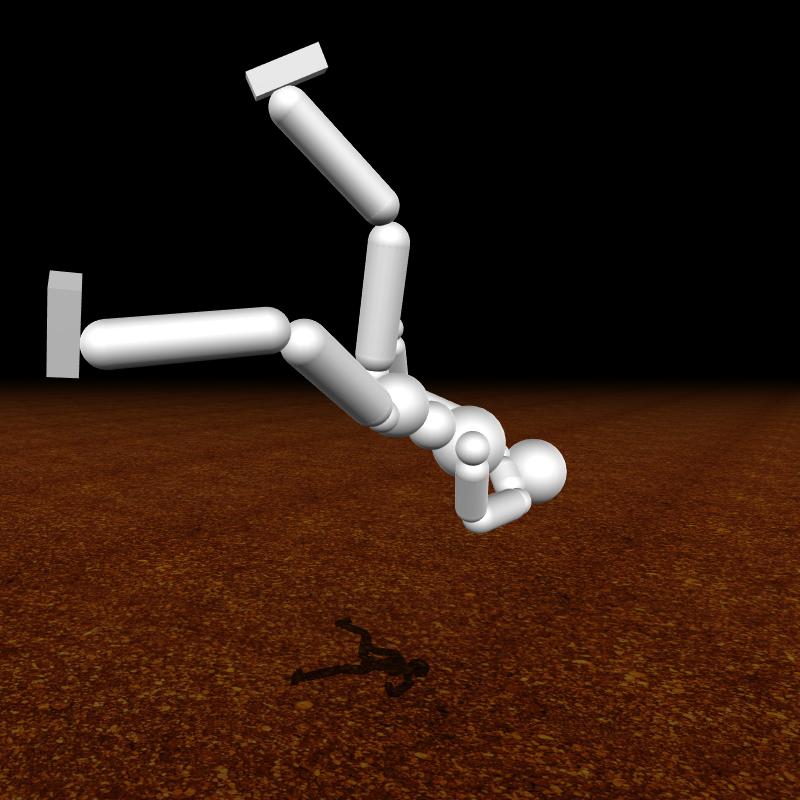}
    \includegraphics[width=0.12\linewidth]{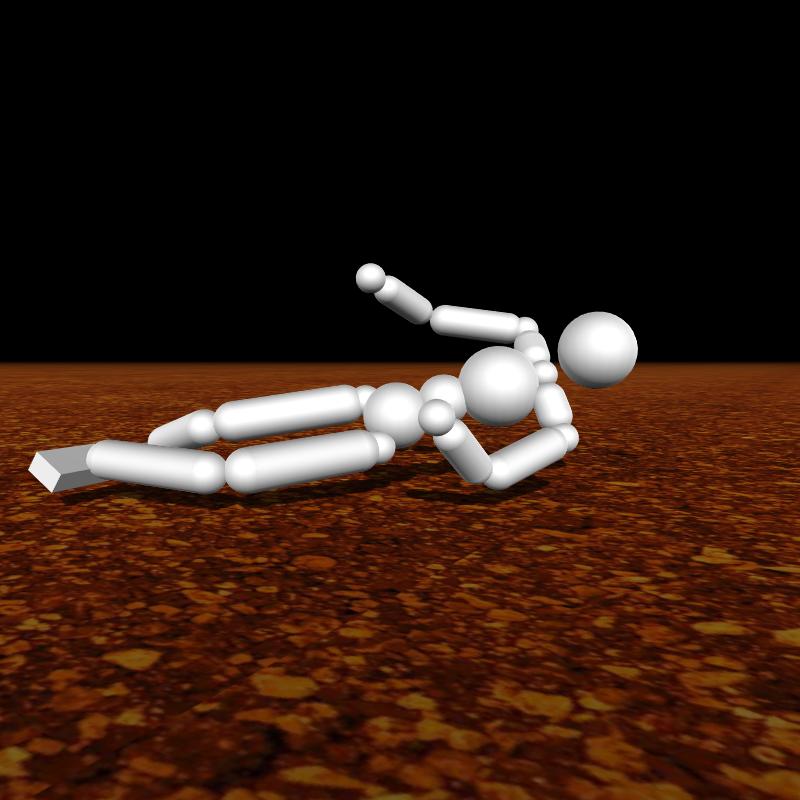}
    \caption{High jump policy trained on Mars with a lower gravity ($g=3.711 m/s^2$), given the initial state of the Fosbury Flop discovered on Earth.}
    \label{fig:High-jump-mars}
\end{figure*}

\section{Results}
\label{sec:results}

We demonstrate multiple strategies discovered through our framework for high jumping and obstacle jumping in Section~\ref{sec:Experiments-Diverse-Strategies}. We validate the effectiveness of BDS and P-VAE in Section~\ref{sec:Experiments-Comparison-and-Ablation-Study}. Comparison with motion capture examples, and interesting variations of learned policies are given in Section~\ref{sec:Results-variations}. All results are best seen in the supplementary videos in order to judge the quality of the synthesized motions.

\subsection{Diverse Strategies}
\label{sec:Experiments-Diverse-Strategies}

\subsubsection{High Jumps}\label{sec:results:high-jump}
In our experiments, six different high jump strategies are discovered during the Stage 1 initial state exploration within the first ten BDS samples: Fosbury Flop, Western Roll (facing up), Straddle, Front Kick, Side Jump, Side Dive. The first three are high jump techniques standard in the sports literature. The last three strategies are not commonly used in sporting events, but still physically valid so we name them according to their visual characteristics. The other four samples generated either repetitions or failures. Strategy repetitions are generally not avoidable due to model errors and large flat regions in the motion space. Since the evaluation of one BDS sample takes about six hours, the Stage 1 exploration takes about 60 hours in total. The discovered distinct strategies at $z_\text{freeze}=100cm$ are further optimized separately to reach their maximum difficulty level, which takes another 20 hours. Representative take-off state feature values of the discovered strategies can be found in Appendix~\ref{app:takeoffStates}. {We also show the DRL learning and curriculum scheduling curves for two strategies in Appendix~\ref{app:curves}.}

In Stage 2, we perform novel policy search for five DRL iterations from each good initial state of Stage 1. Training is warm started with the associated Stage 1 policy for efficient learning. The total time required for Stage 2 is roughly 60 hours. More strategies are discovered in Stage 2, but most are repetitions and only two of them are novel strategies not discovered in Stage 1: Western Roll (facing sideways) and Scissor Kick. Western Roll (sideways) shares the same initial state with Western Roll (up). Scissor Kick shares the same initial state with Front Kick. The strategies discovered in each stage are summarized in Figure~\ref{fig:strategies}. We visualize all eight distinct strategies in Figure~\ref{fig:teaser} and Figure~\ref{fig:highJumps}. We also visualize their peak poses in Figure~\ref{fig:High-jump-peak-poses}.

While the final learned control policies are stochastic in nature, the majority of the results shown in our supplementary video are the deterministic version of those policies, i.e., using the mean of the learned policy action distributions. In the video we further show multiple simulations from the final stochastic policies, to help give insight into the true final endpoint of the optimization. As one might expect for a difficult task such as a maximal-height high jump, these stochastic control policies will also fail for many of the runs, similar to a professional athlete.

\subsubsection{Obstacle Jumps}\label{sec:results:obstacle-jump}
Figure~\ref{fig:obstacleJumps1} shows the six different obstacle jump strategies discovered in Stage 1 within the first 17 BDS samples: Front Kick, Side Kick, Twist Jump (clockwise), Twist Jump (counterclockwise), Straddle and Dive Turn. More strategies are discovered in Stage 2, but only two of them are novel as shown in Figure~\ref{fig:obstacleJumps2}: Western Roll and Twist Turn. Western Roll shares the initial state with Twist Jump (clockwise). Twist Turn shares the initial state with Dive Turn. The two stages together take about 180 hours. We encourage readers to watch the supplementary video for better visual perception of the learned strategies.

Although our obstacle jump task is not an Olympic event, it is analogous to a long jump in that it seeks to jump a maximum-length jumped. 
Setting the obstacle height to zero yields a standard long jump task. The framework discovers several strategies, including one similar to the standard long jump adopted in competitive events, with the strong caveat that the distance achieved is limited by the speed of the run up. Please refer to the supplementary video for the long jump results.

\subsection{Validation and Ablation Study}
\label{sec:Experiments-Comparison-and-Ablation-Study}

\subsubsection{BDS vs. Random Search}
\label{sec:comparison-bds-random-search}
We validate the sample efficiency of BDS compared with a random search baseline. Within the first ten samples of initial states exploration in the high jump task, BDS discovered six distinct strategies as discussed in Section~\ref{sec:results:high-jump}. Given the same computational budget, random search only discovered three distinct strategies: Straddle, Side Jump, and one strategy similar to Scissor Kick. Most samples result in repetitions of these three strategies, due to the presence of large flat regions in the strategy space where different initial states lead to the same strategy. In contrast, BDS largely avoids sampling the flat regions thanks to the acquisition function for diversity optimization and guided exploration of the surrogate model.

\subsubsection{Motion Quality with/without P-VAE}
We justify the usage of P-VAE for improving motion naturalness with results shown in Figure~\ref{fig:ablation}. Without P-VAE, the character can still learn physically valid skills to complete the tasks successfully, but the resulting motions usually exhibit unnatural behavior. In the absence of a natural action space constrained by the P-VAE, the character can freely explore any arbitrary pose during the course of the motion to complete the task, which is unlikely to be within the natural pose manifold all the time.

\subsection{Comparison and Variations}
\label{sec:Results-variations}

\subsubsection{Synthesized High Jumps vs. Motion Capture}
\label{sec:synthesized-mocap}
We capture motion capture examples from a university athlete in a commercial motion capture studio for three well-known high jump strategies: Scissor Kick, Straddle, and Fosbury Flop. We retarget the mocap examples onto our virtual athlete, which is shorter than the real athlete as shown in Table~\ref{tb:modelParams}. We visualize keyframes sampled from our simulated jumps and the retargeted mocap jumps in Figure~\ref{fig:compare-mocap}. Note that the bar heights are set to the maximum heights achievable by our trained policies, while the bar heights for the mocap examples are just the bar heights used at the actual capture session. We did not set the mocap bar heights at the athlete's personal record height, as we wanted to ensure his safety and comfort while jumping in a tight mocap suit with a lot of markers on.

\subsubsection{High Jump Variations}
\label{sec:synthesized-variations}

In addition to discovering multiple motion strategies, our framework can easily support physically valid motion variations. We show four high jump variations generated from our framework in Figure~\ref{fig:variations}. We generate the first three variations by taking the initial state of the Fosbury Flop strategy discovered in Stage 1, and retrain the jumping policy with additional constraints starting from  a random initial policy. Figure~\ref{fig:variation-weakerLeg} shows a jump with a weaker take-off leg, where the torque limits are reduced to $60\%$ of its original values. Figure~\ref{fig:variation-inflexibleSpine} shows a character jumping with a spine that does not permit backward arching. Figure~\ref{fig:variation-fixedKnee} shows a character jumping with a fixed knee joint. All these variations clear lower maximum heights, and are visually different from the original Fosbury Flop in Figure~\ref{fig:highjump-flop}. For the jump in Figure~\ref{fig:variation-landonFoot}, we take the initial state of the Front Kick, and train with an additional constraint that requires landing on feet. In Figure~\ref{fig:High-jump-mars} we also show a high jump trained on Mars, where the gravity $g=3.711m/s^2$ is lower, from the initial state of the Fosbury flop discovered on Earth.

\section{Conclusion and Discussion}

{We have presented a framework for discovering and learning multiple natural and distinct strategies for highly challenging athletic jumping motions. 
A key insight is to explore the take-off state, which is a strong determinant of the jump strategy that follows once airborne. In a second phase, we additionally use explicit rewards for novel motions. Another crucial aspect is to constrain the action space inside the natural human pose manifold. With the proposed two-stage framework and the pose variational autoencoder, natural and physically-nuanced jumping strategies emerge automatically without any reference to human demonstrations. Within the proposed framework, the take-off state exploration is specific to jumping tasks, while the diversity search algorithms in both stages and the P-VAE are task independent. We leave further adaptation of the proposed framework to additional motor skills as future work. We believe this work demonstrates a significant advance by being able to learn a highly-technical skill such as high-jumping.}

{We note that the current world record for men's high jump as of year 2021 is $245cm$, set in year 1993 by an athlete of $193cm$ in height. Our high jump record is $200cm$ for a virtual athlete $170cm$ tall. The performance and realism of our simulated jumps are bounded by many simplifications in our modeling and simulation.} We simplify the athlete's feet and specialized high jump shoes as rigid rectangular boxes, which reduces the maximum heights the virtual athlete can clear. We model the high jump crossbar as a wall at training time and as a rigid bar at run time, while real bars are made from more elastic materials such as fiberglass. We use a rigid box as the landing surface, while real-world landing cushions protect the athlete from breaking his neck and back, and also help him roll and get up in a fluid fashion.

The run-up phase of both jump tasks imitates reference motions, one single curved run for all the high jumps and one single straight run for all the obstacle jumps. The quality of the two reference runs affect the quality of not only the run-up controllers, but also the learned jump controllers. This is because the jump controllers couple with the run-up controllers through the take-off states, for which we only explore a low-dimensional feature space. The remaining dimensions of the take-off states stay the same as in the original reference run. As a result, the run-up controllers for our obstacle jumps remain in medium speed, and the swing leg has to kick backward sometimes in order for the body to dive forward. If we were to use a faster sprint with more forward leaning as the reference run, the discovered jumps could potentially be more natural and more capable to clear wider obstacles. Similarly, we did not find the Hurdle strategy for high jumping, likely due to the reference run being curved rather than straight. In both reference runs, there is a low in-place jump after the last running step. We found this jumping intention successfully embedded into the take-off states, which helped both jump controllers to jump up naturally. Using reference runs that anticipate the intended skills is definitely recommended, although retraining the run-up controller and the jump controller together in a post-processing stage may be helpful as well.

{We were able to discover most well-known high-jump strategies, and some lesser-known variations. There remains a rich space of further parameters to consider for optimization, with our current choices being a good fit for our available computational budget. It would be exciting to discover a better strategy than the Fosbury flop, but a better strategy may not exist. We note that Stage 1 can discover most of the strategies shown in Figure~\ref{fig:strategies}. Stage 2 is only used to search for additional unique strategies and not to fine tune the strategies already learned in Stage 1. We also experimented with simply running Stage 1 longer with three times more samples for the BDS. However, this could not discover any new strategies that can be discovered by Stage 2. In summary, Stage 2 is not absolutely necessary for our framework to work, but it complements Stage 1 in terms of discovering additional visually distinctive strategies. We also note that our Stage 2 search for novel policies is similar in spirit to the algorithm proposed in \cite{zhang2019novel-policies}. An advantage of our approach is its simplicity and the demonstration of its scalability to the discovery of visually distinct strategies for athletic skills.} 

There are many exciting directions for further investigations. First, we have only focused on strategy discovery for the take-off and airborne parts of jumping tasks. For landing, we only required not to land on head first. We did not model get-ups at all. How to seamlessly incorporate landing and get-ups into our framework is a worthy problem for future studies. Second, there is still room to further improve the quality of our synthesized motions. The P-VAE only constrains naturalness at a pose level, while ideally we need a mechanism to guarantee naturalness on a motion level. This is especially helpful for under-constrained motor tasks such as crawling, where feasible regions of the tasks are large and system dynamics cannot help prune a large portion of the state space as for the jumping tasks. Lastly, our strategy discovery is computationally expensive. We are only able to explore initial states in a four dimensional space, limited by our computational resources. If more dimensions could be explored, more strategies might be discovered. Parallel implementation is trivial for Stage 2 since searches for novel policies for different initial states are independent. For Stage 1, batched BDS where multiple points are queried together, similar to the idea of \cite{azimi2010batch}, may be worth exploring. The key challenge of such an approach is how to find a set of good candidates to query simultaneously.

\begin{acks}
We would like to thank the anonymous reviewers for their constructive feedback. We thank Beyond Capture for helping us with motion capture various high jumps. We also thank Toptal for providing us the jumping scene. Yin is partially supported by NSERC Discovery Grants Program RGPIN-06797 and RGPAS-522723. Van de Panne is partially supported by NSERC RGPIN-2020-05929.
\end{acks}

\bibliographystyle{ACM-Reference-Format}
\bibliography{jumping}

\appendix

\begin{table}[t]
\centering
\caption{Representative take-off state features for discovered high jumps.}
\begin{tabular}{|c|c|c|c|c|c|} 
\hline
Strategy & $v_z$ & $\omega_x$ & $\omega_z$ & $\alpha$ \\ 
\hline
Fosbury Flop & -2.40 & -3.00 & 1.00 & -0.05 \\
\hline
Western Roll (up) & -0.50 & 1.00 & -1.00 & 2.09 \\
\hline
Straddle & -2.21 & 1.00 & 0.88 & 1.65 \\
\hline
Front Kick & -0.52 & 1.00 & -0.26 & 0.45 \\
\hline
Side Dive & -1.83 & -2.78 & -0.32 & 1.18 \\
\hline
Side Jump & -1.99 & -1.44 & 0.44 & 0.70 \\
\hline
\end{tabular}
\label{tb:take-off-states}
\end{table}


\begin{table}[t]
\centering
\caption{Representative take-off state features for discovered obstacle jumps.}
\begin{tabular}{|c|c|c|c|} 
\hline
Strategy & $\omega_x$ & $\omega_y$ & $\omega_z$ \\ 
\hline
Front Kick & 1.15 & -1.11 & 3.89 \\
\hline
Side Kick & 3.00 & 3.00 & -2.00 \\
\hline
Twist Jump (c) & -1.50 & 1.50 & -2.00 \\
\hline
Straddle & 0.00 & 0.00 & 1.00  \\
\hline
Twist Jump (cc) & -2.67 & 0.00 & -1.44 \\
\hline
Dive Turn & -0.74 & -2.15 & -0.41 \\
\hline
\end{tabular}
\label{tb:take-off-states-box}
\end{table}

\begin{figure}[t]
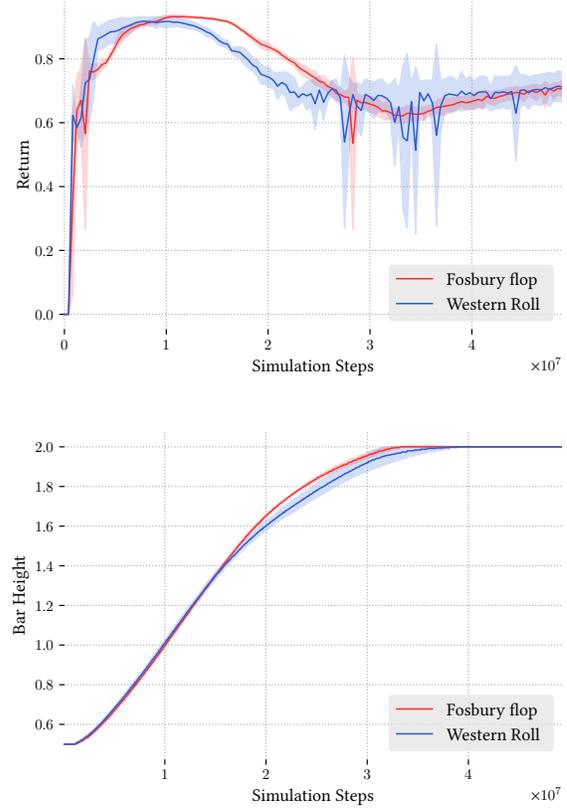

    \centering
    \begin{subfigure}[b]{\linewidth}
        \centering
        \scalebox{0.56}{\input{images/learning-curves/rwd.pgf}}
    \end{subfigure}
    \begin{subfigure}[b]{\linewidth}
        \centering
        \scalebox{0.56}{\input{images/learning-curves/curriculum.pgf}}
    \end{subfigure}
    \caption{{Stage 1 DRL learning and curriculum scheduling curves for two high jump strategies. As DRL learning is stochastic, the curves shown are the average of five training runs. The shaded regions indicates the standard deviation.}}
    \label{fig:learning-curves}
\end{figure}

\section{Representative Take-off State Features}
\label{app:takeoffStates}
We list representative take-off state features discovered through BDS in Table~\ref{tb:take-off-states} for high jumps and Table~\ref{tb:take-off-states-box} for obstacle jumps. The approach angle $\alpha$ for high jumps is defined as the wall orientation in a facing-direction invariant frame. The orientation of the wall is given by the line $x\text{sin}\alpha - z\text{cos}\alpha = 0$. 

\section{Learning Curves}
\label{app:curves}
{We plot Stage 1 DRL learning and curriculum scheduling curves for two high jump strategies in Figure~\ref{fig:learning-curves}. An initial solution for the starting bar height $0.5m$ can be learned relatively quickly. After a certain bar height has been reached (around $1.4m$), the return starts to drop  because larger action offsets are needed to jump higher, which decreases the $r_{naturalness}$ in Equation~\ref{eq:r-pvae} and therefore the overall return in Equation~\ref{eq:stage1-reward}. Subjectively speaking, the learned motions remain as natural for high crossbars, as the lower return is due to the penalty on action offsets.}

\end{document}